\pdfoutput=1

\documentclass[11pt,a4paper]{article}
\usepackage[hyperref]{emnlp2020}
\usepackage{times}
\usepackage{latexsym}

\usepackage[whole]{bxcjkjatype}

\usepackage{tikz}

\usepackage{graphicx}
\usepackage{booktabs,multirow}
\usepackage{subcaption}

\usepackage{amsmath,amssymb,mathtools}
\newcommand{\mathboldface}[1]{\boldsymbol{#1}}
\newcommand{\VEC}[1]{\mathboldface{#1}}

\newcommand{\argmax}{\mathop {\rm argmax}\limits}

\usepackage{zi4}

\aclfinalcopy %
\usepackage{enumitem}
\setlist{%
itemsep=0pt, parsep=0pt, topsep=0pt, partopsep=0pt, %
leftmargin=*, labelsep=0.3em %
}

\AtBeginDocument{ %
\setlength\abovedisplayskip{4pt plus 2pt minus 2pt}
\setlength\belowdisplayskip{4pt plus 2pt minus 2pt}
\setlength\abovedisplayshortskip{\abovedisplayskip}
\setlength\belowdisplayshortskip{\belowdisplayskip}
}

\usepackage[hang,flushmargin]{footmisc} %

\newcommand{\bgf}[1]{\tikz[baseline=(X.base)]{\node(X)[rectangle, fill=blue!10, rounded corners, text height=1.4ex,text depth=-0.5ex,draw=white]{#1};}}
\newcommand{\bgfg}[1]{\tikz[baseline=(X.base)]{\node(X)[rectangle, fill=gray!15, rounded corners, text height=1.4ex,text depth=-0.5ex,draw=white]{#1};}}
\newcommand{\bgfr}[1]{\tikz[baseline=(X.base)]{\node(X)[rectangle, fill=purple!9, rounded corners, text height=1.4ex,text depth=-0.5ex]{#1};}}

\newcommand{\malpha}{$\alpha$}
\newcommand{\mfx}{$f(\VEC{x})$}

\newcommand{\msfx}{$\lVert f(\VEC{x})\rVert$}
\newcommand{\msafx}{$\lVert\alpha f(\VEC{x})\rVert$}

\title{Attention is Not Only a Weight: \\Analyzing Transformers with Vector Norms}

\author{
Goro\,Kobayashi$^{1}$\hspace{1em}
Tatsuki\,Kuribayashi$^{1,2}$\hspace{1em}
Sho\,Yokoi$^{1,3}$\hspace{1em}
Kentaro\,Inui$^{1,3}$\\[2pt]
$^{1}$ Tohoku University\hspace{1em}
$^{2}$ Langsmith Inc.\hspace{1em}
$^{3}$ RIKEN\hspace{1em}
\\
\texttt{\{goro.koba, kuribayashi, yokoi, inui\}@ecei.tohoku.ac.jp}}
\vspace{-4ex}
\date{\ttfamily
\{goro.koba, kuribayashi, yokoi, inui\}@ecei.tohoku.ac.jp
}

\date{}

\begin{document}
\maketitle
\begin{abstract}
\label{sec:abstract}

Attention is a key component of %
Transformers, which have recently achieved considerable success in natural language processing.
Hence, attention is being extensively studied to investigate various linguistic capabilities of Transformers, focusing on analyzing the parallels between \emph{attention weights} and specific linguistic phenomena.
This paper shows that attention weights alone are only one of the two factors that determine the output of attention
and proposes a norm-based analysis that incorporates the second factor, the norm of the transformed input vectors.
The findings of our norm-based analyses of BERT and a Transformer-based neural machine translation system include the following:
(i) contrary to previous studies, BERT pays poor attention to special tokens, 
and (ii) reasonable word alignment can be extracted from attention mechanisms of Transformer.
These findings provide insights into the inner workings of %
Transformers.
\end{abstract}

\section{Introduction}
\label{sec:introduction}
Transformers~\citep{vaswani17,devlin2018bert,yang19xlnet,liu19,lan2020albert} have improved the state-of-the-art in a wide range of natural language processing tasks.
The success of the models has not yet been sufficiently explained; hence, substantial research has focused on assessing the linguistic capabilities of these models~\citep{bertology,clark19}.

One of the main features of Transformers is that they utilize an attention mechanism without the use of recurrent or convolutional layers. 
The attention mechanism computes an output vector by accumulating relevant information from a sequence of input vectors.
Specifically, it assigns attention weights (i.e., relevance) to each input, and sums up input vectors based on their weights.
The analysis of correlations between attention weights and various linguistic phenomena (i.e., \textit{weight-based analysis}) is a prominent research area~\citep{clark19,kovaleva19,coenen19,lin19,marecek19balustrades,Htut2019,raganato2018,tang_transformer_attention_analysis}.

This paper first shows that weight-based analysis is insufficient to analyze the attention mechanism.
Weight-based analysis is a common approach to analyze the attention mechanism by simply tracking attention weights.
The attention mechanism can be expressed as a \textit{weighted} sum of \textit{linearly transformed vectors} (Section~\ref{subsec:summation_mechanism}); however, the effect of transformed vectors in weight-based analysis is ignored.
We propose a \textit{norm-based analysis} that considers the previously ignored factors (Section~\ref{sec:proposal}).
In this analysis, we measure the norms (lengths) of the vectors that were summed to compute the output vector of the attention mechanism.

Using the norm-based analysis of BERT (Section~\ref{sec:experiment}), we 
interpreted the internal workings of the model in more detail than when weight-based analysis was used.
For example, the weight-based analysis~\cite{clark19, kovaleva19} reports that specific tokens, such as periods, commas, and special tokens (e.g., separator token; \texttt{[SEP]}), tend to have high attention weights.
However, our norm-based analysis found that the information collected from vectors corresponding to special tokens was considerably lesser than that reported in the weight-based analysis, and the large attention weights of these vectors were canceled by other factors.
Additionally, we found that BERT controlled the levels of contribution from frequent, less informative words by controlling the norms of their vectors. 

In the analysis of a Transformer-based NMT system (Section~\ref{sec:nmt}), 
we reinvestigated how accurate word alignment can be extracted from the source-target attention.
The weight-based results of~\citet{li19-word_align},~\citet{ding19-saliency}, and~\citet{zenkel_aer_giza} have empirically shown that word alignments induced by the source-target attention of the Transformer-based NMT systems are noisy.
Our experiments show that more accurate alignments can be extracted by focusing on the vector norms.

The contributions of this study are as follows:
\begin{itemize}
\item 
We propose a novel method of analyzing an attention mechanism based on vector norms (norm-based analysis). 
The method considers attention weights and previously ignored factors, i.e., the norm of the transformed vector.

\item Our norm-based analysis of BERT reveals that
(i) the attention mechanisms pay considerably lesser attention to special tokens than to observations that are solely based on attention weights (weight-based analysis),
and (ii) the attention mechanisms tend to %
discount 
frequent words.

\item Our norm-based analysis of a Transformer-based NMT system
reveals that reasonable word alignment can be extracted from source-target attention, in contrast to the previous results of the weight-based analysis.
\end{itemize}

\noindent
The codes of our experiments are publicly available.\footnote{
    \url{https://github.com/gorokoba560/norm-analysis-of-transformer}
}

\section{Background}
\label{sec:background}
\subsection{Attention mechanism}%
\label{subsec:mechanism_overview}

\begin{figure}[t]
    \centering
    \includegraphics[width=\hsize]{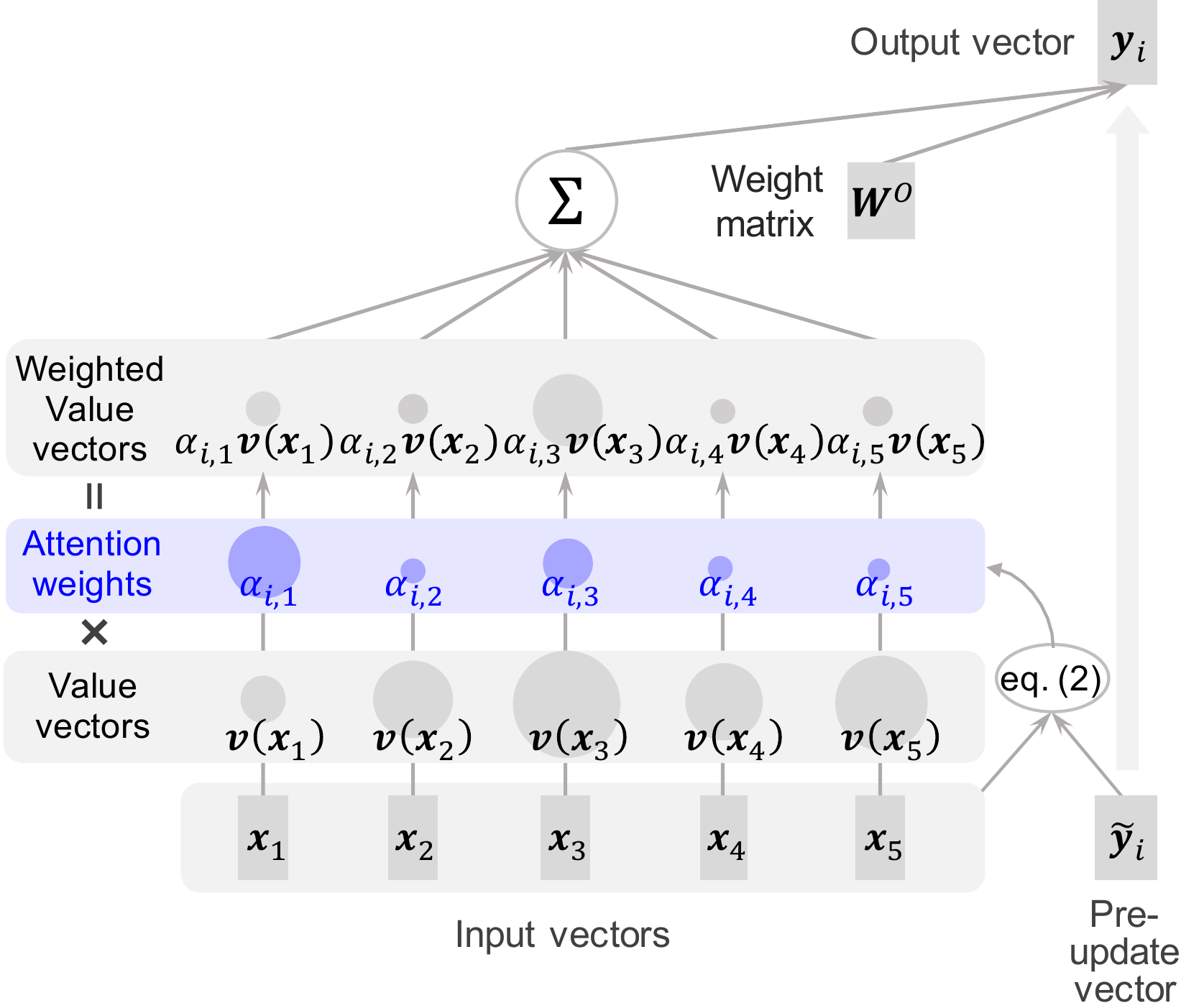}
    \caption{
    Overview of attention mechanism in Transformers.
    Sizes of the colored circles illustrate the value of the scalar or the norm of the corresponding vector.
    }
    \label{fig:pre-transformed_module}
\end{figure}

Attention is a core component of Transformers, which consist of several layers, each containing multiple attentions (``heads'').
We focused on analyzing the inner workings of these heads.
As illustrated in Figure~\ref{fig:pre-transformed_module}, 
each attention head gathers relevant information from the input vectors.
A vector is updated by vector transformations, attention weights, and a summation of vectors.
Mathematically, attention computes each output vector $\VEC{y}_i\in\mathbb{R}^d$ from the corresponding pre-update vector $\widetilde{\VEC{y}}_i\in\mathbb{R}^d$ and a sequence of input vectors $\mathcal X = \{\VEC{x}_1,\dots,\VEC{x}_n\} \subseteq \mathbb R^d$:
\begingroup
\allowdisplaybreaks %
\begin{align}
    &\VEC{y}_i=\biggl(\sum^n_{j=1}\alpha_{i,j}\VEC{v}(\VEC{x}_j)\biggr)\VEC{W}^O
    \label{eq1:self_attention}
    \\
    &\alpha_{i,j} := \mathop{\operatorname{softmax}}_{\VEC{x}_j \in \mathcal X}\left(\frac{\VEC{q}(\widetilde{\VEC{y}}_i) \VEC{k}(\VEC{x}_j)^\top}{\sqrt{d'}}\right)\in\mathbb{R}
    \label{eq2:attention_weight}
    \text{,}
\end{align}
\endgroup
where $\alpha_{i,j}$ is the attention weight assigned to the token $x_j$ for computing $y_i$, and
$\VEC{q}(\cdot)$, $\VEC{k}(\cdot)$, and $\VEC{v}(\cdot)$ are the query, key, and value transformations, respectively.

\small
\begin{align}
    \nonumber
    &\VEC{q}(\widetilde{\VEC{y}}_i):=\widetilde{\VEC{y}}_i \VEC{W}^Q+\VEC{b}^Q
    \quad\left(\VEC{W}^{Q}\in\mathbb{R}^{d\times d'},\, \VEC{b}^{Q}\in\mathbb{R}^{d'}\right)
    \\
    \nonumber
    &\VEC{k}(\VEC{x}_j):=\VEC{x}_j \VEC{W}^K+\VEC{b}^K
    \quad\left(\VEC{W}^{K}\in\mathbb{R}^{d\times d'},\, \VEC{b}^{K}\in\mathbb{R}^{d'}\right)
    \\
    \nonumber
    &\VEC{v}(\VEC{x}_j):=\VEC{x}_j \VEC{W}^V+\VEC{b}^V
    \quad\left(\VEC{W}^{V}\in\mathbb{R}^{d\times d'},\, \VEC{b}^{V}\in\mathbb{R}^{d'}\right)\!
    \text{.}
\end{align}

\normalsize
\noindent
Attention gathers value vectors $\VEC{v}(\VEC{x}_j)$ based on attention weights and then, applies matrix multiplication $\VEC{W}^O\in\mathbb{R}^{d'\times d}$ (Figure~\ref{fig:pre-transformed_module}).~\footnote{
    Whether bias $\VEC{b}$ is added to calculate query, key, and value vectors depends on the implementation. 
    $\VEC{W}^O\in\mathbb{R}^{d'\times d}$ in Equation~\ref{eq1:self_attention} corresponds to the part of $\VEC{W}^O\in\mathbb{R}^{hd' \times d}$ that was introduced in~\citet{vaswani17} which is applied to each head; where $h$ is the number of heads, and $hd'=d$ holds.
}
Boldface letters such as $\VEC{x}$ denote row (not column) vectors, following the notations in~\citet{vaswani17}. 

In self-attention, the input vectors $\mathcal X$ and the pre-update vector $\widetilde{\VEC{y}}_i$ are previous layer's output representations.
In source-target attention, %
$\mathcal X$ corresponds to the representations of the encoder,
and vector $\widetilde{\VEC{y}}_i$ (and updated vector $\VEC{y}_i$) corresponds to the vector of the $i$-th input token of the decoder.

\subsection{Attention is a weighted sum of vectors}
\label{subsec:summation_mechanism}
\begin{figure}[t]
    \centering
    \includegraphics[width=\hsize]{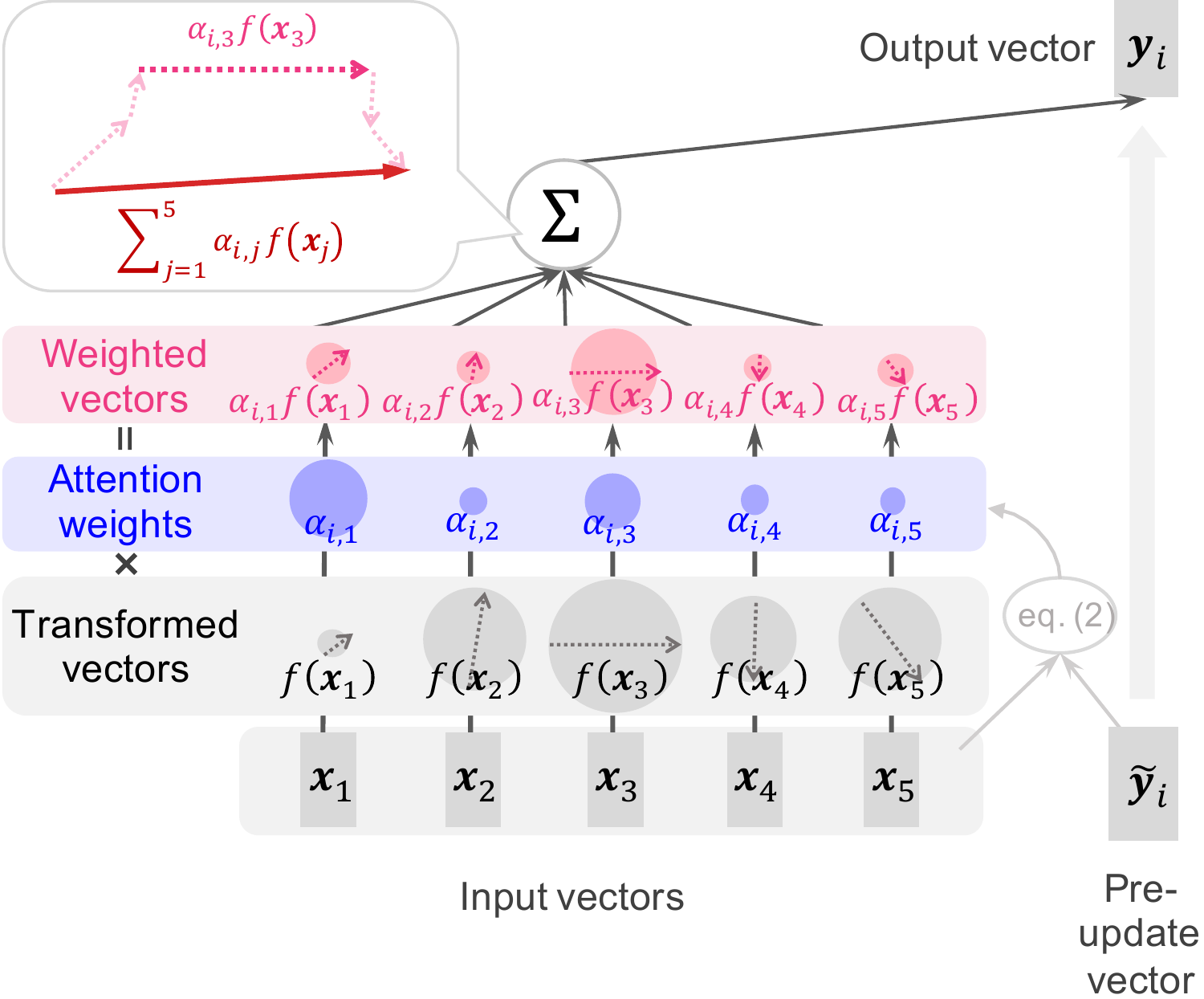}
    \caption{
    Overview of attention mechanism based on Equation~\ref{eq7:transformed_self_attention}.
    It computes the output vector by summing the weighted vectors; vectors with larger norms have higher contributions.
    Sizes of the colored circles illustrate the value of the scalar or the norm of the corresponding vector.
    }
    \label{fig:transformed_attention}
\end{figure}

With a simple reformulation, one can observe that the attention mechanism computes the weighted sum of the transformed input vectors.
Because of the linearity of the matrix product, we can rewrite Equation~\ref{eq1:self_attention} as
\begingroup
\allowdisplaybreaks %
\begin{align}
    \VEC{y}_i &=
    \sum^n_{j=1}{\tikz[baseline=(X.base)]{\node(X)[rectangle, fill=purple!9, rounded corners, text height=2.2ex,text depth=1ex]{\bgf{\textcolor{blue}{$\alpha_{i,j}$}}\bgfg{\textcolor{black}{$f(\VEC{x}_j)$}}}}}
    \label{eq7:transformed_self_attention} \\
    f(\VEC{x}) &:= \left(\VEC{x}\VEC{W}^V + \VEC{b}^V\right)\VEC{W}^O
    \text{.}
    \label{eq8:transform_T}
\end{align}
\endgroup
Equation~\ref{eq7:transformed_self_attention} shows that
the attention mechanism first transforms each input vector $\VEC{x}$ to generate \bgfg{$f(\VEC{x})$}; computes attention weights \bgf{\textcolor{blue}{$\alpha$}}; and then compute the \emph{sum} \bgfr{\textcolor{magenta!50!purple}{$\alpha f(\VEC{x})$}} (see Figure~\ref{fig:transformed_attention}).

\subsection{Problems encountered in weight-based analysis}
\label{subsec:problem_previous_analysis}

The attention mechanism has been designed to update representations by gathering relevant information from the input vectors.
Prior studies have analyzed attention, %
focusing on attention weights, to ascertain which input vectors contribute  (weight-based analysis)~\citep{clark19,kovaleva19,coenen19,lin19,marecek19balustrades,Htut2019,raganato2018,tang_transformer_attention_analysis}.

Analyses solely based on attention weight are based on the assumption that the larger the attention weight of an input vector, the higher its contribution to the output.
However, this assumption disregards the magnitudes of the transformed vectors.
The problem encountered when neglecting the effect of $f(\VEC{x}_j)$ is illustrated in Figure~\ref{fig:transformed_attention}.
The transformed vector $f(\VEC{x}_1)$ for input $\VEC{x}_1$ is assumed to be very small ($\lVert f(\VEC{x}_1 ) \rVert \approx 0$), while its attention weight $\alpha_{i,1}$ is considerably large.
Note that the small $\alpha_{i,1} f(\VEC x_1)$ contributes a little to the output vector $\VEC y_i$ because $\VEC y_i$ is the sum of $\alpha f(\VEC x)$, where a larger vector contributes more to the output.
Conversely, the large $\alpha_{i,3} f(\VEC x_3)$ dominates the output $\VEC y_i$.
Therefore, in this case, 
only considering the attention weight may lead to a wrong interpretation of the high contribution of input vector $\VEC{x}_1$ to output $\VEC{y}_i$.
Nevertheless, $\VEC x_1$ hardly has any effect on $\VEC y_i$.

Analyses based on attention weights have not provided clear results in some cases.
For example, \citet{clark19} reported that input vectors for separator tokens \texttt{[SEP]} tend to receive remarkably large attention weights in BERT, while changing the magnitudes of these weights does not affect the masked-token prediction of BERT.
Such results can be attributed to the aforementioned issue of focusing only on attention weights.

\section{Proposal: norm as a degree of attention}
\label{sec:proposal}
As described in Section~\ref{subsec:problem_previous_analysis},
analyzing the attention mechanism with only attention weights neglects the effect of the transformed vector $f(\VEC{x}_j)$, which has a significant impact as we discussed later. %
Herein, we propose the measurement of the \emph{norm of the weighted transformed vector} \bgfr{\textcolor{magenta!50!purple}{$\lVert\alpha f(\VEC{x})\rVert$}}, given by Equation~\ref{eq7:transformed_self_attention}, to analyze the attention mechanism behavior.\footnote{We use the standard Euclidean norm.}
Unlike in previous studies, we analyzed the behaviors of the norms, $\lVert\alpha f(\VEC{x})\rVert$ and $\lVert f(\VEC{x})\rVert$, and $\alpha$ to gain more in-depth insights into the functioning of attention.
The proposed method of analyzing the attention mechanism is called \textbf{norm-based analysis} and the method that solely analyzes the attention weights is called \textbf{weight-based analysis}.

In Sections~\ref{sec:experiment} and~\ref{sec:nmt},
we provide insights into the working of Transformers using norm-based analysis.
Appendix~\ref{ap:re-multi} explains that our norm-based analysis can also be effectively applied to an entire multi-head attention mechanism.

\section{Experiments: BERT}
\label{sec:experiment}
First, we show that the previously ignored transformed-vector norm affects the analysis of attention in BERT (Section~\ref{subsec:ignored_effect}).
Applying our norm-based analysis, we re-examine the previous reports on BERT obtained by weight-based analysis (Section~\ref{subsec:re-examin}).
Next, we demonstrate the previously overlooked properties of BERT %
(Section~\ref{sec:freq_and_fx}).

\paragraph{General settings:}
Following the previous studies~\citep{clark19,kovaleva19,coenen19,lin19,Htut2019}, we used the pre-trained BERT-base\footnote{
    We used PyTorch implementation of BERT-base (uncased) released at \url{https://github.com/huggingface/transformers}.
},
with 12 layers, each containing 12 attention heads.
We used the data provided by~\citet{clark19} for the analysis.\footnote{\url{https://github.com/clarkkev/attention-analysis}}
The data contains 992 sequences extracted from Wikipedia, where each sequence consists of two consecutive paragraphs, in the form of: \texttt{[CLS] paragraph1 [SEP] paragraph2 [SEP]}.
Each sequence consists of up to 128 tokens, with an average of 122 tokens.

\subsection{Does \mfx{} have an impact?}
\label{subsec:ignored_effect}
\begin{table}[t]
\centering
\setlength{\tabcolsep}{3pt}  %
\renewcommand{\arraystretch}{0.8}
    {\small
\begin{tabular}{cccccc}
\toprule
Head & $\mu$ & $\sigma$ & CV & Max & Min \\
\cmidrule(r){1-1} \cmidrule(lr){2-2} \cmidrule(lr){3-3} \cmidrule(lr){4-4} \cmidrule(lr){5-5} \cmidrule(l){6-6} 
Layer 2--Head 4 (max CV) & 4.26 & 1.59 & \textbf{0.37} & 12.66 & 0.96 \\
Layer 2--Head 7 (min CV) & 4.00 & 0.50 & \textbf{0.12} & 6.15 & 1.35 \\
Average & 5.15 & 1.17 & \textbf{0.22} & - & - \\
\bottomrule
\end{tabular}
}
\caption{Mean ($\mu$), standard deviation ($\sigma$), coefficient of variance (CV), and maximum and minimum values of \msfx. In the last row, the former three are averaged over all the heads.
}
\label{table:fx_variance}
\end{table}

We analyzed the coefficient of variation (CV)%
\footnote{
    \emph{Coefficient of variation} (CV) is a standardized (scale-invariant) measure of dispersion, which is defined by the ratio of the standard deviation $\sigma$ to the mean $\mu$; $\text{CV} := \sigma/\mu$.
}
of previously ignored effect---\msfx---to first demonstrate the degree to which \msafx{} differs from weight \malpha.
We computed the CV of \msfx{} of all the example data for each head.
Table~\ref{table:fx_variance} shows that the average CV is 0.22.
Typically, the value of the norm $\lVert f(\VEC{x}) \rVert$ varies from 0.78 to 1.22 times the average value of the $\lVert f(\VEC{x}) \rVert$. %
Thus, there is a difference between the weight \malpha{} and \msafx{} due to the dispersion of \msfx, which motivated us to consider \msfx\ in the attention analysis.
Appendix~\ref{ap:x_and_f} presents the detailed results.

\subsection{Re-examining previous observation}%
\label{subsec:re-examin}
In this section, with the application of our norm-based analysis, we reinvestigate the previous observation of \citet{clark19}; they analyzed BERT using the weight-based analysis.

\begin{figure}[t]
    \centering
    \begin{minipage}[b]{\hsize}
        \centering
        \includegraphics[height=3.5cm]{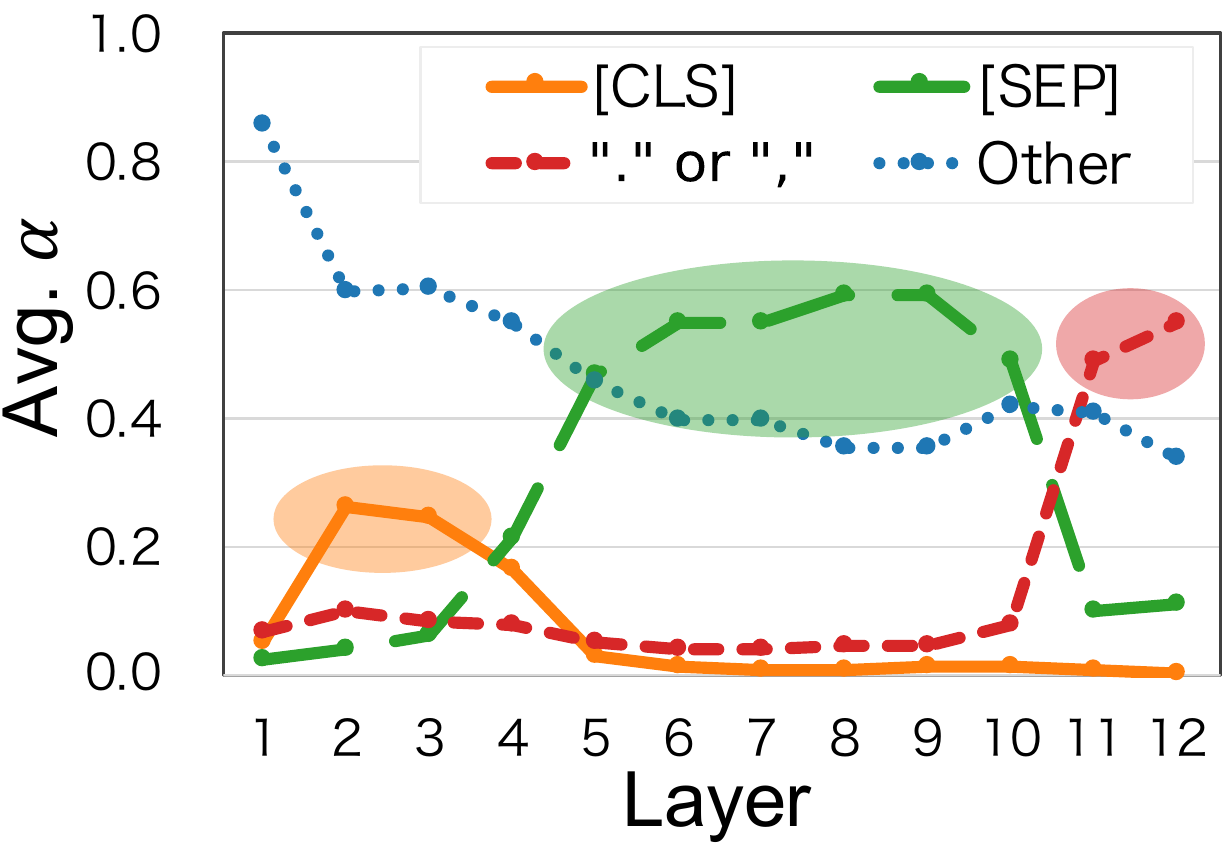}
        \subcaption{
        Weight-based analysis. %
        }
        \label{fig:clark_a}
    \end{minipage}
    \begin{minipage}[b]{\hsize}
        \centering
        \includegraphics[height=3.5cm]{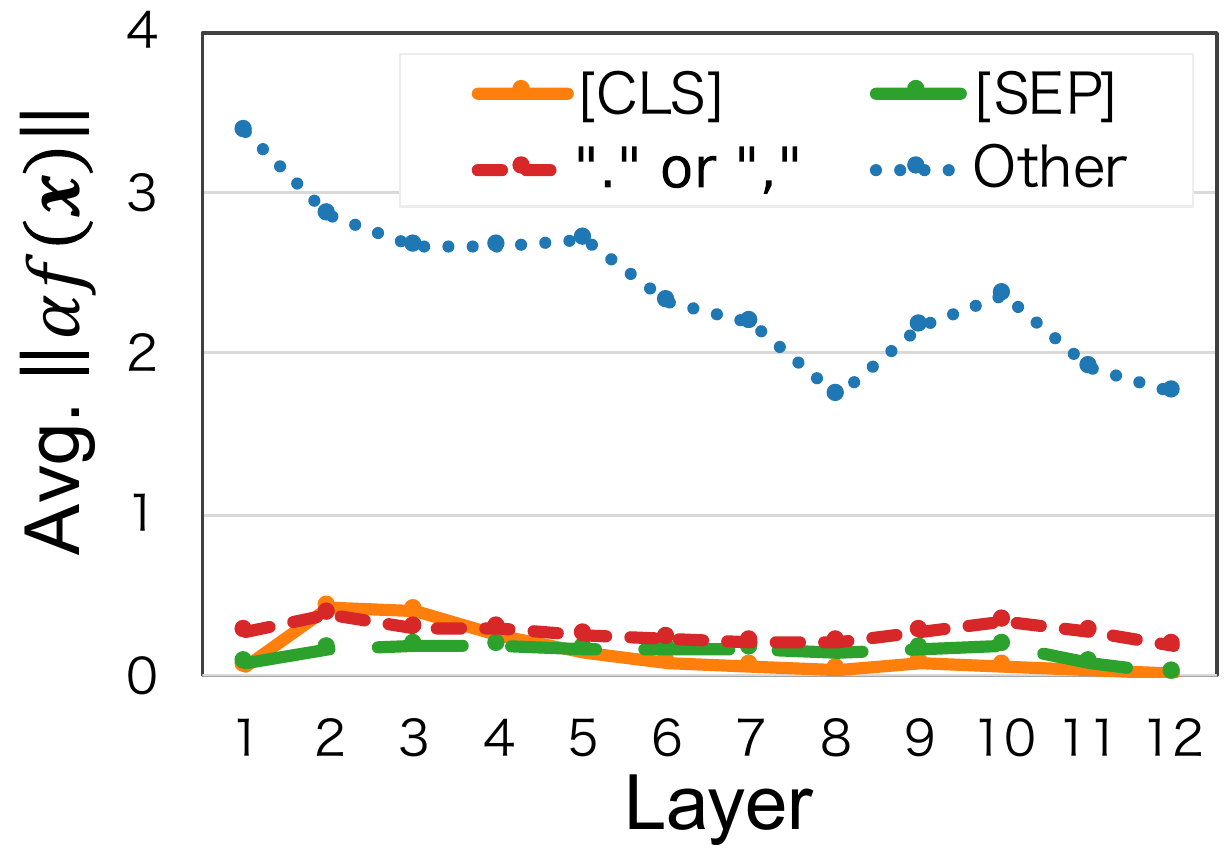}
        \subcaption{Norm-based analysis. %
        }
        \label{fig:clark_afx}
    \end{minipage}
    \caption{
    Each point corresponds to averaged \malpha{} or \msafx{} on a %
    word category in a given layer.
    Note that, in each layer, the sum of \malpha{} among all the categories is 1.
    The $x$-axis denotes the index of the layers. %
    }
    \label{fig:alpha_at_comparison}
\end{figure}
\paragraph{Settings:}
First, all the data were fed into BERT.
Then, the weight \malpha\ and \msafx\ were collected from each head. %
Following~\citet{clark19}, we report the results of the following categories: (i) \texttt{[CLS]}, (ii) \texttt{[SEP]}, (iii) periods and commas, and (iv) the other tokens.
More specific descriptions of the experiments are provided in Appendix~\ref{ap:re-examin}.

\begin{table}[t]
\centering
\renewcommand{\arraystretch}{0.8}
    {\small
\begin{tabular}{@{}crc@{}}
\toprule
Token category & Number of vectors & Spearman's $\rho$ \\
\cmidrule(r){1-1} \cmidrule(lr){2-2} \cmidrule(l){3-3} 
{[}CLS{]}    & 17,443,296   & -0.34 \\
{[}SEP{]}    & 34,886,592   & -0.69 \\
comma \& period     & 182,838,528  & -0.25 \\
Others               & 1,944,928,224 & -0.06 \\ \bottomrule
\end{tabular}
}
\caption{
Spearman rank correlation coefficient between \malpha\ and \msfx\ in each token category.}
\label{table:rank_corr}
\end{table}

\paragraph{Results:}
The weight-based and norm-based analyses exhibited entirely different trends (Figure~\ref{fig:alpha_at_comparison}).
The vectors for specific tokens---\texttt{[CLS]}, \texttt{[SEP]}, and punctuations---have remarkably large attention weights, which is consistent with the report of \citet{clark19}.
In contrast, our norm-based analysis demonstrated that the contributions of vectors corresponding to these tokens were generally small (Figure~\ref{fig:clark_afx}).
The result demonstrates that %
the size of the transformed vector \mfx{} plays a considerable role in controlling the amount of information obtained from the specific tokens.

\citet{clark19} hypothesized that if the necessary information is not present in the input vectors, BERT assigns large weights to \texttt{[SEP]}, which appears in every input sequence, to avoid the incorporation of any additional information via attention.\footnote{
    Note that the attention mechanism has the constraint that the sum of the attention weights becomes $1.0$ (see Equation~\ref{eq2:attention_weight}).}
\citet{clark19} called this operation no-operation (no-op).
However, it is unclear whether assigning large attention weights to \texttt{[SEP]} realizes the operation of collecting little information from the input sequence.

Our norm-based analysis demonstrates that the amount of information from the vectors corresponding to \texttt{[SEP]} is small (Figure~\ref{fig:clark_afx}).
This result supports the interpretation that BERT conducts ``no-op,'' in which attention to \texttt{[SEP]} is considered a signal that does not collect anything.
Additionally, we hope that our norm-based analysis can provide a better interpretation of other existing findings.

\paragraph{Analysis---The relationship between \malpha\ and \msfx{}:}
It remains unclear how attention \emph{collects only a little information} while assigning a high attention weight to a specific token, \texttt{[SEP]}.
Here, we demonstrate an interesting trend of \malpha\ and \msfx\ cancelling each other out on the tokens.\footnote{
    Note that for any positive scalar $\lambda \in \mathbb R$ and vector $\VEC x\in \mathbb R^d$, $\lVert \lambda \VEC x\rVert = \lambda \lVert\VEC x\rVert$.
}
Table~\ref{table:rank_corr} shows the Spearman rank correlation coefficient between \malpha\ and \msfx{},
corresponding to the vectors in each category.
The weight \malpha\ and the norm \msfx\ have %
a negative correlation in terms of \texttt{[CLS]}, \texttt{[SEP]}, periods, and commas.
This cancellation manages to \emph{collect a little information} even with large weights.

\begin{figure}[t]
    \centering
    \begin{minipage}[t]{.47\hsize}
        \centering
        \includegraphics[width=\hsize]{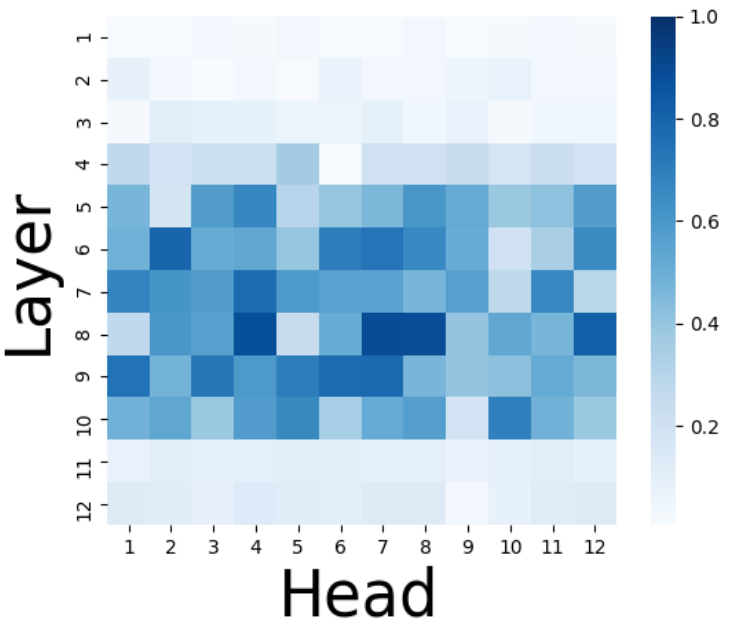}
        \subcaption{
        \malpha{}.
        }
        \label{fig:sep_weight}
    \end{minipage}
    \;
    \begin{minipage}[t]{.47\hsize}
        \centering
        \includegraphics[width=\hsize]{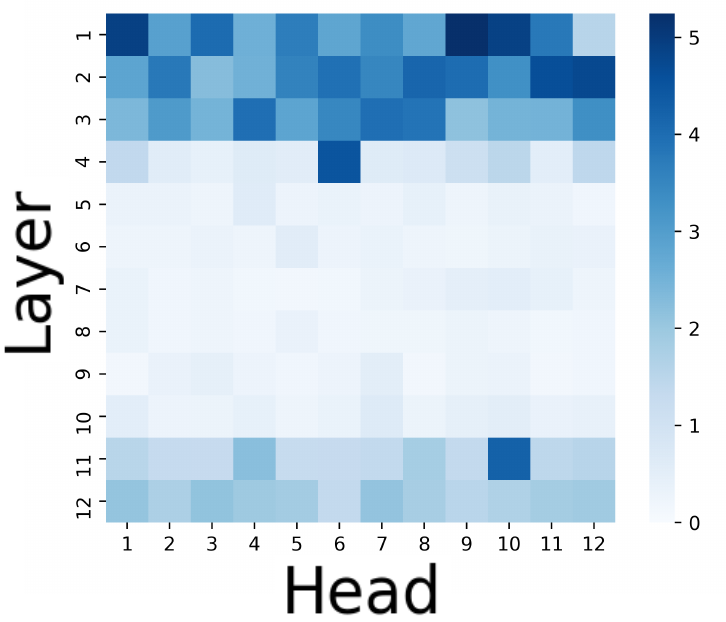}
        \subcaption{
        \msfx{}.
        }
        \label{fig:sep_xt}
    \end{minipage}
    \caption{
    The higher value of averaged $\alpha$ or \msfx{} for \texttt{[SEP]} tokens in a given head, the darker its cell.
    }
    \label{fig:sep_detail_analysis}
\end{figure}

Figure~\ref{fig:sep_detail_analysis} illustrates the contrast between \malpha{} and \msfx{}
corresponding to \texttt{[SEP]} in each head.
For most of the heads, \malpha\ and \msfx\ clearly negate the magnitudes of each other.
A similar trend was observed in %
\texttt{[CLS]}, periods, and commas.
Conversely, 
no significant trend was observed in the other tokens (see Appendix~\ref{ap:detail:a-fx}).
Figure~\ref{fig:relation_alpha-fx} shows 1\% randomly selected pairs of \malpha\ and \msfx\ in each word category.
Even when the same weight \malpha\ is assigned, %
\msfx\ can vary, suggesting that \malpha\ and \msfx\ play a different roles in attention.

\begin{figure}[t]
    \center
    \includegraphics[width=6.0cm]{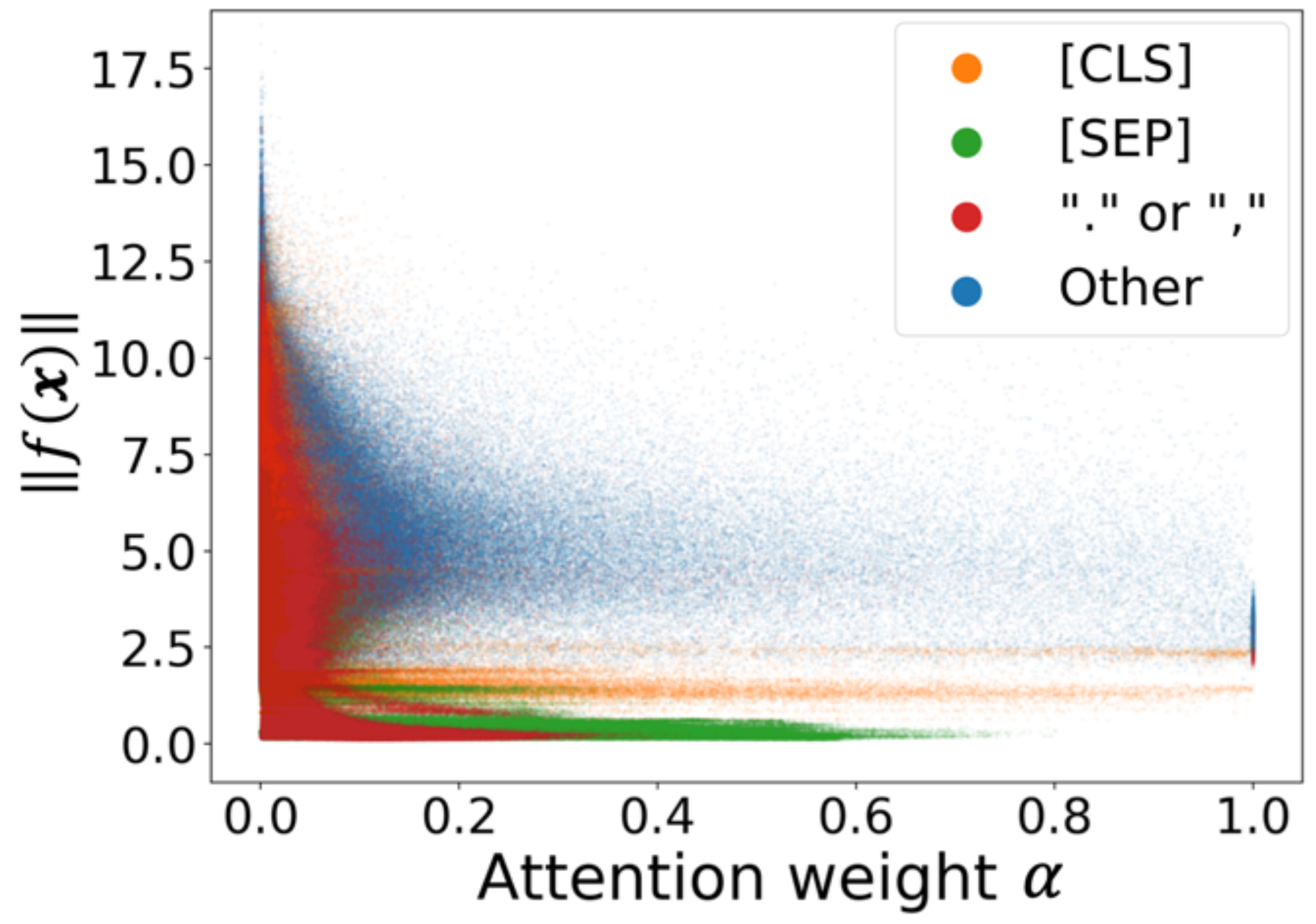}
    \caption{
    Relationship between \malpha\ and \msfx. 
    Each plot corresponds to a pair of $\alpha_{i,j}$ and $\lVert f(\VEC{x}_j)\rVert$ in one of the attention heads.
    Each plot is colored by the word category corresponding to $\VEC{x}_j$.
    Visualizations by category are shown in Appendix~\ref{ap:detail:a-fx}.
    }
    \label{fig:relation_alpha-fx}
\end{figure}

\subsection{Relation between frequency and \msfx{}}
\label{sec:freq_and_fx}

In the previous section, we demonstrated that \msfx\ corresponding to the specific tokens (e.g., \texttt{[SEP]}) is small.
Based on the high frequencies\footnote{
    The frequency ranks of the words \texttt{[CLS]}, \texttt{[SEP]}, period, and comma, out of approximately 30,000 words, are 50, 28, 2, and 3, respectively.
    }
of these word types\footnote{
    We call word type as ``word.'' Each instance of a word is called ``token.''},
we hypothesized that BERT controlled contributions of highly frequent, less informative words by adjusting the norm of \mfx{}.

\paragraph{Settings:}
First, all the data were fed into the model. 
Then, for each input token $t$, we collected the weight $\alpha$ and $\lVert f(\VEC{x})\rVert$.
We averaged $\alpha$ and $\lVert f(\VEC{x})\rVert$ for all the heads for each $t$ to analyze the trend of the entire model.
Let $r(\cdot)$ be a function that returns the frequency rank of a given word.%
\footnote{
    We counted the frequency for each word type by reproducing the training data of BERT.  
}
We analyzed the relationship of $r(t)$ with $\alpha$ and $\lVert f(\VEC{x})\rVert$.

\paragraph{Results:}
The Spearman rank correlation coefficient between the frequency rank $r(t)$ and $\lVert f(\VEC{x})\rVert$ was $0.75$, indicating a strong positive correlation.
In contrast, the Spearman rank correlation coefficient %
did not show any correlation ($\rho=0.06$) between $r(t)$ and $\alpha$.\footnote{
The Spearman rank correlation coefficient without special tokens, periods, and commas was $0.28$ for the attention weights and $0.69$ for the norms.
}
The visualizations of their relationships are shown in Appendix~\ref{ap:freq}.

These results demonstrate that the self-attentions %
in BERT reduce the information from highly frequent words by adjusting $\lVert f(\VEC{x}) \rVert$ and not $\alpha$.
This frequency-based effect is consistent with the intuition that highly frequent words, such as stop words, are unlikely to play an important role in solving the pre-training tasks (masked-token prediction and next-sentence prediction).
\section{Experiments: Transformer for NMT}
\label{sec:nmt}
Additionally, we analyzed the source-target attention in a Transformer-based NMT system.
One major research topic in the NMT field is whether NMT systems internally capture word alignment between source and target texts, and if so, how word alignment can be extracted from black-box NMT systems.
\citet{li19-word_align},~\citet{ding19-saliency}, and~\citet{zenkel_aer_giza} empirically showed, using the weight-based method, that word alignment induced by the attention %
of the Transformer is noisy.
In this section, we show the analysis of source-target attention using vector norms \msafx{} and %
demonstrate that clean alignments can be extracted from the source-target attention.
Word alignment can be used to provide %
rich information for the users of NMT systems~\citep{ding19-saliency}.

\paragraph{Experimental procedure:}
Following~\citet{zenkel_aer_giza} and~\citet{ding19-saliency}, we trained a Transformer-based NMT system for German-to-English translation on the Europarl v7 corpus\footnote{\url{http://www.statmt.org/europarl/v7}}.
Next, we extracted word alignments from \malpha{} and \msafx{} under the force decoding setup. %
Finally, we evaluated the derived alignment using the alignment error rate (AER)~\citep{och&ney00-aer}.
A low AER score indicates that the extracted word alignments are close to the reference.
We used the gold alignment dataset provided by~\citet{vilar2006_rwth}\footnote{\url{https://www-i6.informatik.rwth-aachen.de/goldAlignment/}}.
Experiments were performed on five random seeds, and the average AER scores were reported.
The experimental settings are detailed in Appendix~\ref{ap:nmt}.

\subsection{Alignment extraction from attention}

\paragraph{Weights or norms:}
A typical alignment extraction method uses attention weights~\citep{li19-word_align,ding19-saliency,zenkel_aer_giza}.
Specifically, given a source-target sentence pair, $\{s^{}_1,\dots, s^{}_{J}\}$ and $\{t^{}_1,\dots, t^{}_{I}\}$,
word alignment is estimated
by calculating a source word $s_j$ that has the highest weight when generating a target word $t_i$.
We call this method the \textit{weight-based alignment extraction}.
In contrast, we propose a \textit{norm-based alignment extraction} method that extracts word alignments based on \msafx{} instead of \malpha.
Formally, in these methods, the source word $s_j$ with the highest attention weight or norm during the generating of target word $t^{}_i$ is extracted as the word that is aligned with $t^{}_i$:
\begin{align}
    \argmax_{s^{}_j} \alpha^{}_{i,j}
    \quad\text{or}\quad
    \argmax_{s^{}_j} \, \lVert\alpha^{}_{i,j} f(\VEC x^{}_j)\rVert
    \text{.}
    \label{eq:align_afx_head_next}
\end{align}

In Section~\ref{subsec:nmt:comp}, %
following~\citet{li19-word_align}, we %
analyze the word alignments that we obtained from each layer by integrating $H$ heads within the same layer:
\begin{align}
    \nonumber
    \argmax_{s^{}_j} \sum_{h=1}^H \alpha_{i,j}^{h}
    \quad \text{or}\quad
    \argmax_{s^{}_j} \lVert \sum_{h=1}^H \alpha_{i,j}^{h} f^{h}(\VEC{x}_j) \rVert
    \text{,}
\end{align}

\noindent
where $f^{h}(\VEC{x}_j)$ and $\alpha_{i,j}^{h}$ are the transformed vector and the attention weight at the $h$-th head, respectively.

\begin{figure}[t]
    \center
    \includegraphics[width=\hsize]{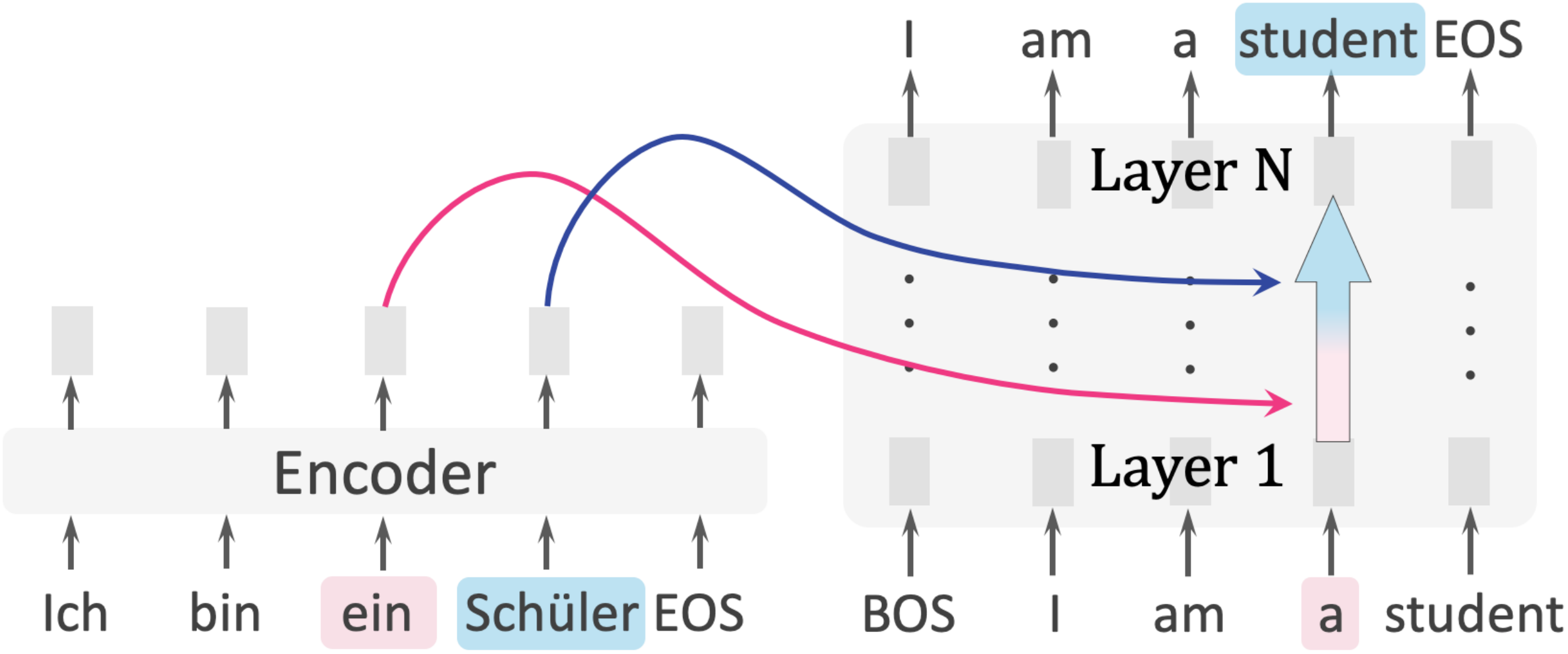}
    \caption{
    An example of behavior of the source-target attentions in an NMT system (German-to-English).
    Attentions in the earlier layers focus the source word ``ein'' aligned with the input word ``a,'' while those in the latter layers focus the source word ``Schüler'' aligned with the output word ``student.''}
    \label{fig:enc-dec}
\end{figure}

\paragraph{Alignment with input or output word:}
In our preliminary experiments (Appendix~\ref{subsec:nmt:layer-wise}), we observed that the behavior of the source-target attention of the decoder differs between the earlier and later layers.
As shown in Figure~\ref{fig:enc-dec}, at the time decoding the word $t_{i+1}$ with the input $t_i$, attention heads in the earlier layers assign large weights or norms to $s_j$ corresponding to the input $t_i$ ``a,'' whereas those in the latter layers assign 
large values to $s_j$ corresponding to the output word $t_{i+1}$ ``student.''

Based on this observation, we explored two settings for investigating alignment extraction methods:
\textit{alignment with output} (AWO) and \textit{alignment with input} (AWI).
The AWO setting refers to the approach introduced in Equation~\ref{eq:align_afx_head_next}.
Specifically, alignments $(s_j, t_i)$ were extracted by considering a source word $s_j$ that gained the highest weight (norm) when \emph{outputting} a particular target word $t_i$.

In the AWI setting, alignments $(s_j, t_i)$ were extracted by considering a source word $s_j$ that gained the highest weight (norm) when \emph{inputting} the word $t_i$ (i.e., predicting a word $t_{i+1}$).
Formally, alignment with the AWI setting is calculated as follows:
\begin{align}
    \argmax_{s^{}_j} \alpha^{}_{i+1,j}
    \quad\text{or}\quad
    \argmax_{s^{}_j} \, \lVert\alpha^{}_{i+1,j} f(\VEC x^{}_j)\rVert
    \text{.}
    \label{eq:align_afx_head_current}
\end{align}
\subsection{Comparative experiments}
\label{subsec:nmt:comp}
We compared the quality of the alignments that were obtained by the following %
six methods:
\vspace{0.1cm}
\begin{itemize}
    \item norm-based extraction with the AWO/AWI settings
    \item weight-based extraction with the AWO/AWI settings~\cite{li19-word_align,zenkel_aer_giza,ding19-saliency}
    \item gradient-based extraction~\cite{ding19-saliency}
    \item existing word aligners~\cite{och&ney_giza,dyer13-fastalign}
\end{itemize}
\vspace{0.1cm}
\noindent
We report the best and averaged AER scores across the layers.
In addition, we report on the AER score at the head and the layer with the highest average \msafx\ in the norm-based extraction.\footnote{
    The average \msafx\ of the layer was determined by the sum of the average \msafx\ at each head in the layer.
}
The settings are detailed in Appendix~\ref{ap:nmt_ext}.

\begin{table}[t]
\centering
\setlength{\tabcolsep}{4pt}  %
\renewcommand{\arraystretch}{0.85}
{\small
\begin{tabular}{lcc}
\toprule
\multicolumn{1}{c}{\textbf{Methods}} & AER & ±SD \\
\cmidrule(r){1-1} \cmidrule{2-2} \cmidrule(l){3-3}
\textbf{Transformer -- Attention-based Approach} \\
\multicolumn{1}{c}{--- \textit{Alignment with output} setting ---} \\
Weight-based \\
\hspace{15pt} layer mean & 68.4 & 1.0 \\ 
\hspace{15pt} best layer (layer 4 or 5) & 47.7 & 1.7 \\ 
Norm-based (ours) \\
\hspace{15pt} layer mean & 62.9 & 0.7 \\ 
\hspace{15pt} best layer (layer 5) & 41.4 & 1.4 \\
\hspace{15pt} layer with the highest average \msafx\ & 83.0 & 1.1 \\
\hspace{15pt} head with the highest average \msafx\ & 87.1 & 2.3 \\
\\
\multicolumn{1}{c}{--- \textit{Alignment with input} setting ---} \\
Weight-based \\
\hspace{15pt} layer mean & 68.5 & 1.9 \\ 
\hspace{15pt} best layer (layer 2) & 29.8 & 3.7 \\ 
Norm-based (ours) \\
\hspace{15pt} layer mean & 60.4 & 1.3 \\ 
\hspace{15pt} best layer (layer 2) & 25.0 & 1.5 \\
\hspace{15pt} layer with the highest average \msafx\ & 25.0 & 1.5 \\
\hspace{15pt} head with the highest average \msafx\  & 35.5 & 21.0 \\
\cmidrule(r){1-1} \cmidrule{2-2} \cmidrule(l){3-3}
\textbf{Transformer -- Gradient-based Approach} \\
SmoothGrad from \citet{ding19-saliency} & 36.4 & - \\
\cmidrule(r){1-1} \cmidrule{2-2} \cmidrule(l){3-3}
\textbf{Word Aligner} \\
fast\_align from \citet{zenkel_aer_giza} & 28.4 & - \\
GIZA++ from \citet{zenkel_aer_giza} & 21.0 & - \\ \bottomrule
\end{tabular}
}
\caption{
AER scores with different %
methods for German-to-English translation.
The closer the extracted word alignment is to the reference, the lower the AER score.
The ``layer mean'' denotes the average of AER scores across all layers.
Each value is the average of five random seeds.
}
\label{table:aer_comparison}
\end{table}

\begin{figure*}[t]
    \centering
    \begin{minipage}[t]{.3\hsize}
        \centering
        \includegraphics[height=4.5cm]{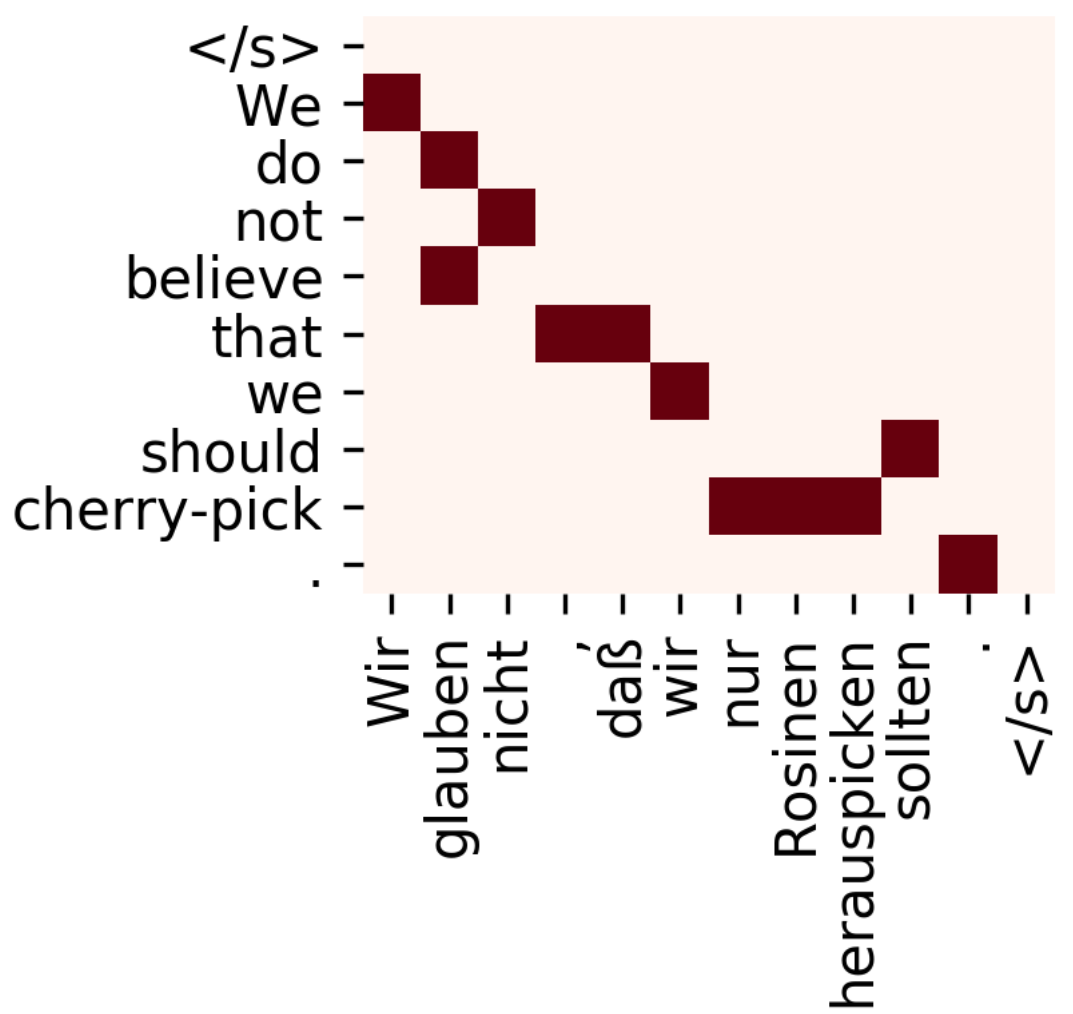}
        \subcaption{
        Reference.
        }
        \label{fig:aer_ex_gold}
    \end{minipage}
    \;
    \begin{minipage}[t]{.3\hsize}
        \centering
        \includegraphics[height=4.5cm]{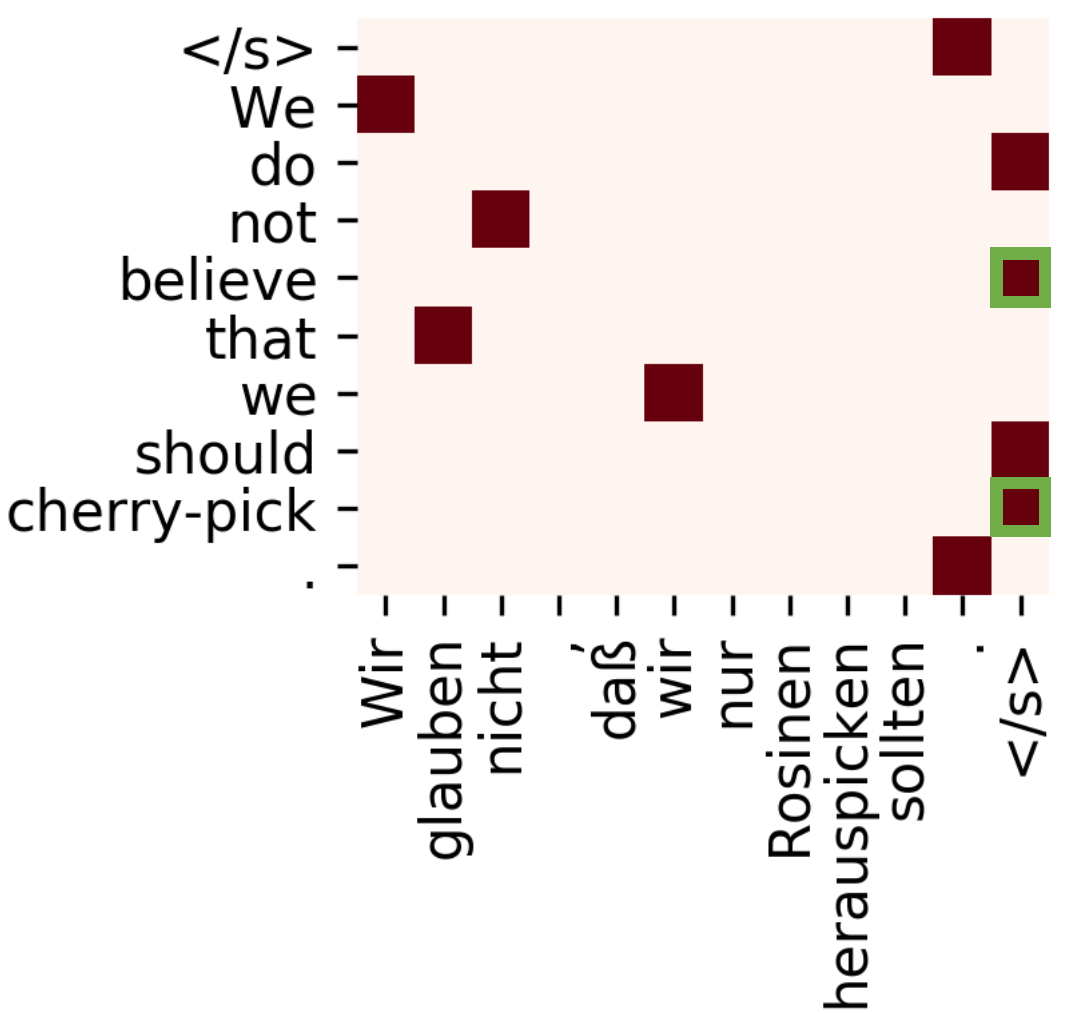}
        \subcaption{
        \malpha. %
        }
        \label{fig:aer_ex_a}
    \end{minipage}
    \;
    \begin{minipage}[t]{.3\hsize}
        \centering
        \includegraphics[height=4.5cm]{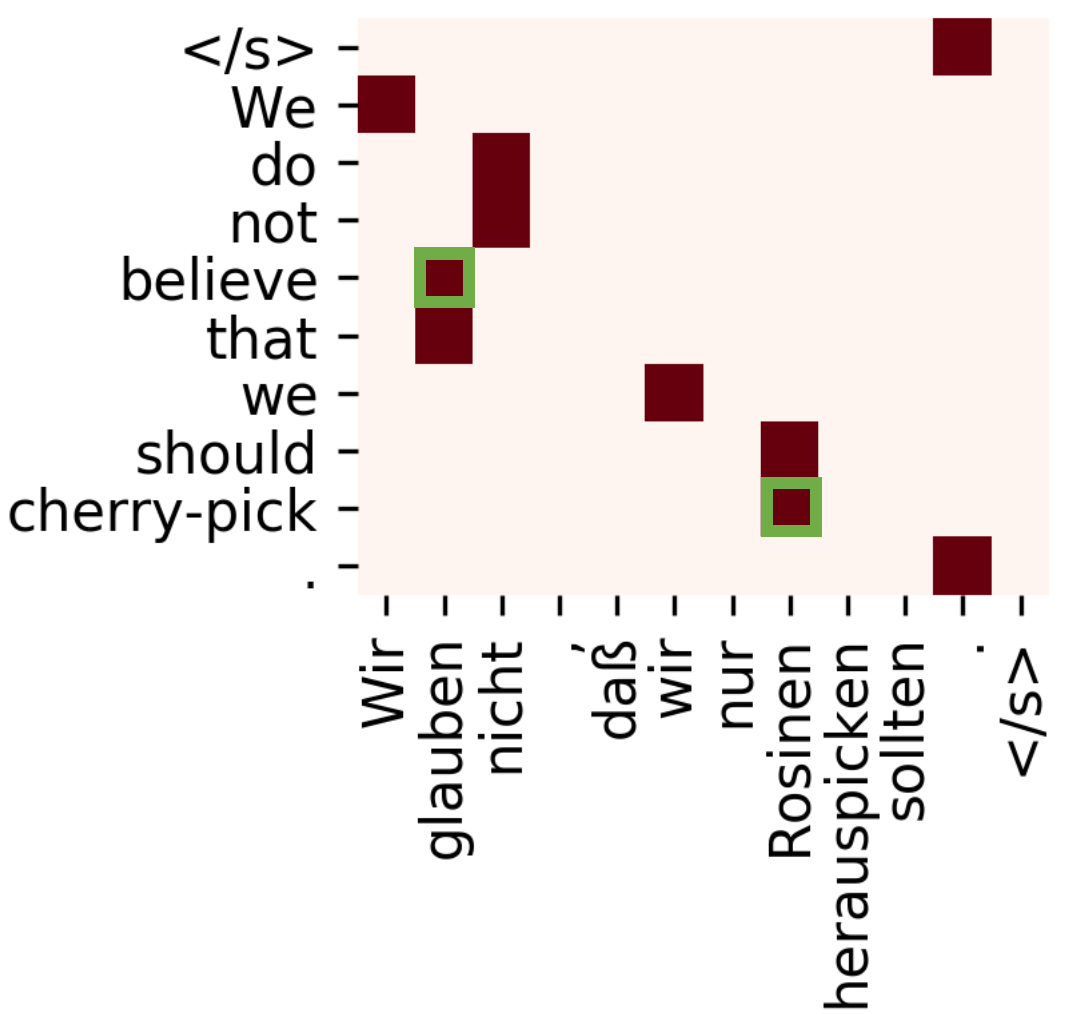}
        \subcaption{
        \msafx. %
        }
        \label{fig:aer_ex_summed_afx}
    \end{minipage}
    \caption{
    Examples of the reference and extracted alignments using each method in layer 2 (best layer) in the AWI setting on one out of five seeds.
    Two misalignments in the weight-based extraction were resolved in the norm-based analysis---alignments with the green frame.
    Examples of the extracted alignments in all the layers are shown in Appendix~\ref{ap:nmt_other_lyaer}.
    }
    \label{fig:aer_ex}
\end{figure*}

The AER scores of each method are listed in Table~\ref{table:aer_comparison}.
The results show that word alignments extracted using the proposed norm-based approach are more reasonable than those extracted using the weight-based approach.
Additionally, better word alignments were extracted in the AWI setting than in the AWO setting.
The alignment extracted using the layer with the highest average \msafx\ in the AWI setting is better than the gradient-based method, and competitive with one of the existing word aligners---fast\_align.\footnote{
Even at the head with the highest average \msafx. Although the average score of five seeds in the AWI setting was $35.5$, four seeds out of them achieved great score range from $23.6$-to $25.7$. The score was $77.5$ for a remaining seed.
}
These results show that much clearer word alignments can be extracted from a Transformer-based NMT system than the results reported by existing research.

The primary reason behind the differences between the results of the weight- and norm-based methods was analogous to the finding discussed in Section~\ref{subsec:re-examin},
while some specific tokens, such as \texttt{$\langle$/s$\rangle$}, the special token for the end of the sentence, tended to obtain heavy attention weights; their transformed vectors were adjusted to be smaller, as shown in Figure~\ref{fig:aer_ex}.

\subsection{Relationship between norms and alignment quality}
\label{subsec:nmt:aer_norm}
\begin{figure*}[t]
    \centering
    \begin{minipage}[t]{.3\hsize}
        \centering
        \includegraphics[height=5cm]{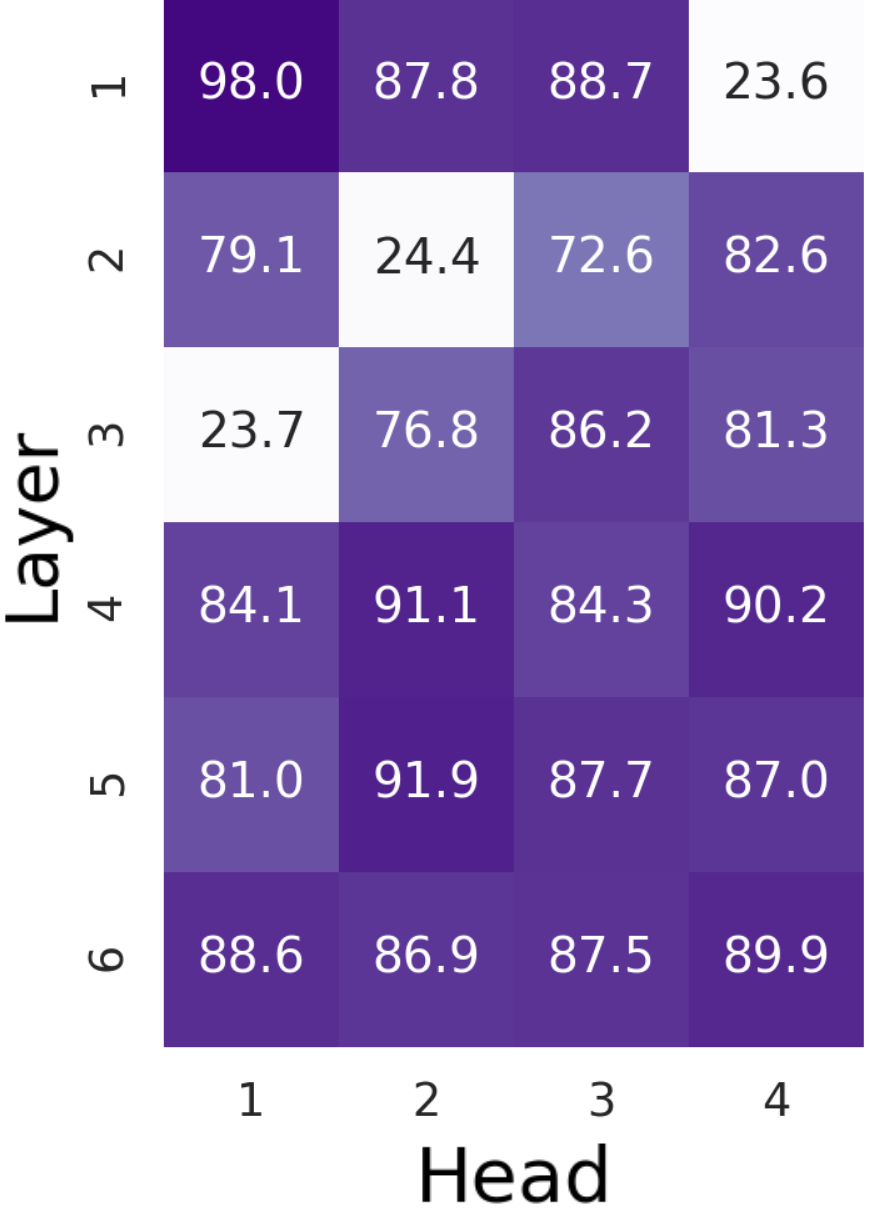}
        \subcaption{
        AER in the AWI setting.
        }
    \label{fig:head_aer_current}
    \end{minipage}
    \begin{minipage}[t]{.3\hsize}
        \centering
        \includegraphics[height=5cm]{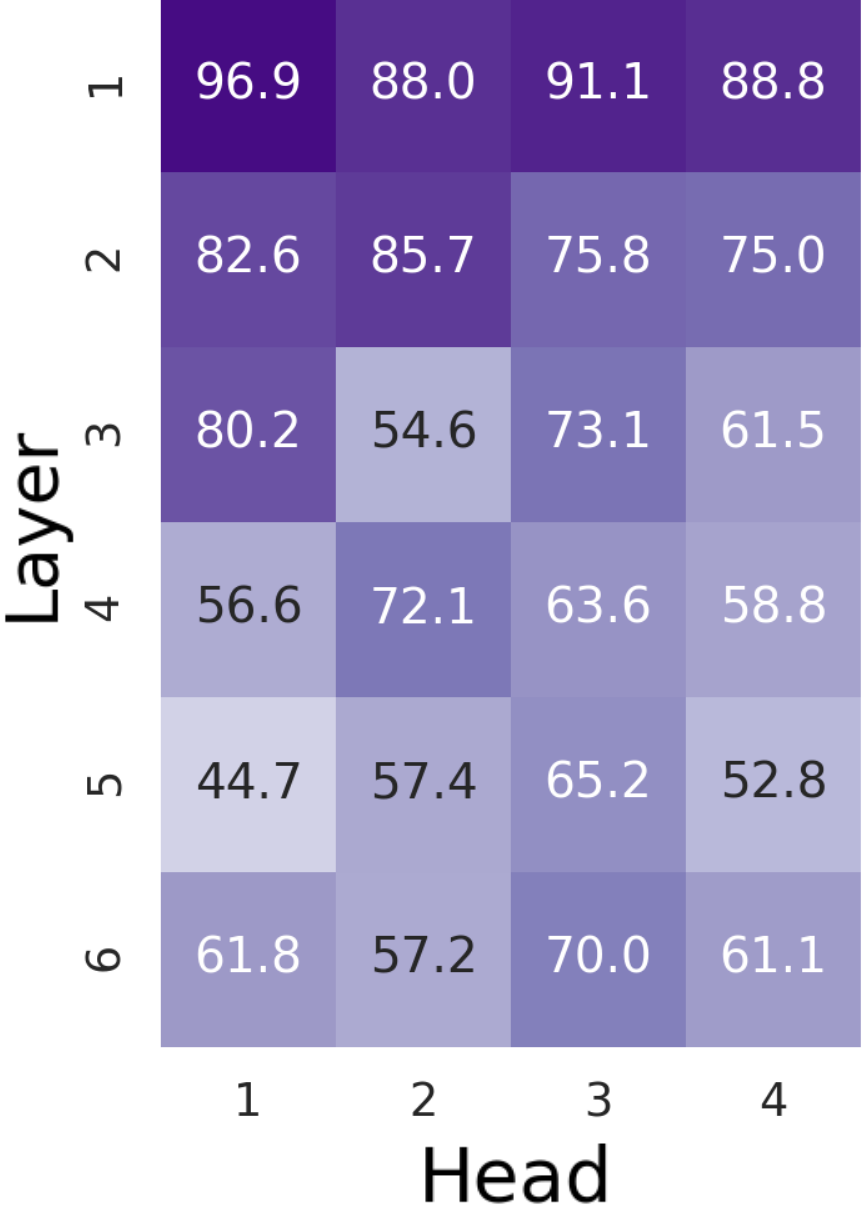}
        \subcaption{
        AER in the AWO setting.
        }
        \label{fig:head_aer_next}
    \end{minipage}
    \;
    \begin{minipage}[t]{.3\hsize}
        \centering
        \includegraphics[height=5cm]{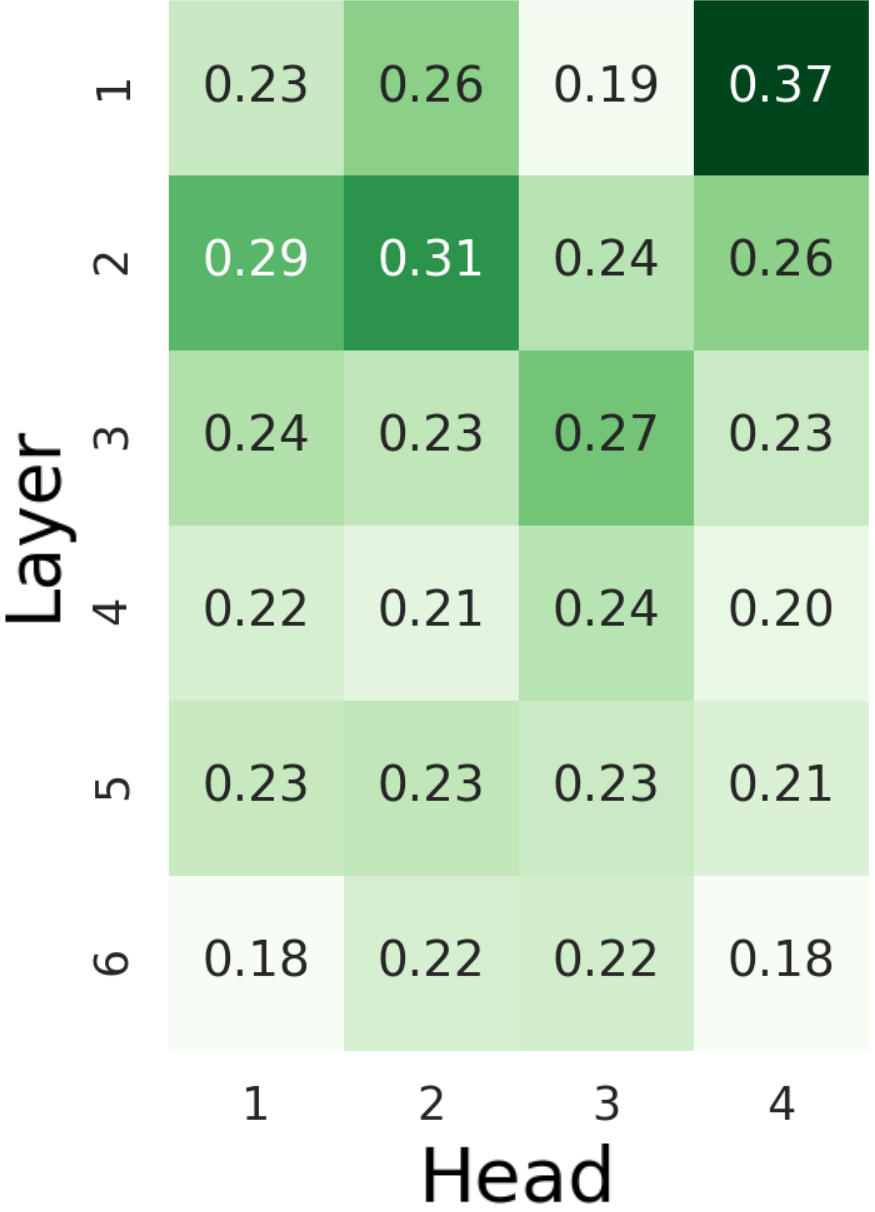}
        \subcaption{
        Averaged \msafx.
        }
    \label{fig:head_job}
    \end{minipage}
    \caption{
    AER scores %
    and averaged \msafx{} in each head on one out of five seeds.
    The closer the extracted word alignment is to the reference, the lower the AER score---the lighter the color. 
    The larger the averaged \msafx{}, the darker the color.
    }
    \label{fig:aer_job}
\end{figure*}

We further analyze the relationship between \msafx\ and AER scores in the head-level.
Figures~\ref{fig:head_aer_current} and \ref{fig:head_aer_next} show the AER scores of the alignments obtained by the norm based extraction at each head in the AWO and AWI settings.
Figure~\ref{fig:head_job} shows the average of \msafx\ at each head.
The small \msafx\ implies that %
\malpha\ and 
\msfx\ tend to cancel out in the head.

Comparing Figures~\ref{fig:head_aer_current} and~\ref{fig:head_job}, the average \msafx\ and AER scores in the AWI setting are inversely correlated (the Spearman rank and Pearson correlation coefficients are $-0.44$ and $-0.52$, respectively).
This result is consistent with Table~\ref{table:aer_comparison}, where the head or the layer with the highest average \msafx\ provides clean alignments in the AWI setting.
This result suggests that Transformer-based NMT systems may rely on specific heads that align source and target tokens.
This result is also consistent with the exiting reports that pruning some attention heads in Transformers does not change its performance; on the contrary, it improves the performance~\cite{michel19,kovaleva19}.

In contrast, in the AWO setting (Figures~\ref{fig:head_aer_next} and~\ref{fig:head_job}), such a negative correlation is not observed; rather, a positive correlation is observed (Spearman's $\rho$ is $0.56$, and the Pearson's $r$ is $0.55$).
Actually, in the AWO setting, the alignments extracted from the head/layer with the highest \msafx\ is considerably worse than those from the other settings in Table~\ref{table:aer_comparison}.
Investigating the reason for these %
contrasting results would be our future work.
In Appendix~\ref{ap:nmt_wmt}, we also present the results of a model with a different number of heads.

\section{Related work}
\label{sec:relatedwork}
\subsection{Probing of Transformers} %
Transformers are used for many NLP tasks.
Many studies have probed their inner workings to understand the mechanisms underlying their success~\citep{bertology,clark19}.

There are mainly two probing perspectives to investigate these models; they differ based on whether the target of the analysis is per-token level or it considers token-to-token interactions.
The first category assesses a single word or phrase-level linguistic capabilities of BERT, such as its performance on part-of-speech tagging and word sense disambiguation performance~\cite{tenney19,jawahar19,coenen19,lin19,wallace19_number}.

The latter category explores the ability of Transformers to capture token-to-token interactions, such as syntactic relations and word alignment in the translation~\citep{clark19,kovaleva19,Htut2019,coenen19,lin19,goldberg19,ding19-saliency,zenkel_aer_giza,li19-word_align,raganato2018}.
The present study is closely related to the latter group; we have provided insights into the token-to-token attention in Transformer-based systems.

\subsection{Analyzing the token-to-token interaction}
Two types of methods are mainly considered to analyze the token-to-token interactions in Transformers.
One is to track the attention weights, and the other is to check the gradient
of the output with respect to the input of attention mechanisms.

\paragraph{Weight-based analysis:}
Many studies have analyzed the linguistic capabilities of Transformers by tracking attention weights.
This type of analysis has covered a wide range of subjects, including syntactic and semantic relationships~\cite{tang_transformer_attention_analysis,raganato2018,clark19,coenen19,jawahar19,Htut2019,kovaleva19,marecek19balustrades}.
However, as outlined in Section~\ref{subsec:problem_previous_analysis}, these studies have ignored the effect of \mfx{}.
It has been actively discussed so far whether the attention weights can be interpreted to explain the models~\citep{jain18,serrano19,wiegreffe19,pruthi19,Vashishth2019}.

\citet{brunner19} have introduced ``effective attention,'' which has upgraded the weight-based analysis.
Their proposal is similar to ours;
they exclude attention weights that do not affect the output owing to the application of transformation $f$ and input $\VEC{x}$ in the analysis.
However, our proposal differs from theirs in some aspects.
Specifically, we aim to analyze the behavior of the whole attention mechanism more accurately, whereas they aim to make the attention weights more accurate.
Furthermore, the effectiveness of their approach depends on the length of an input sequence; however, ours approach does not have such a limitation (see Appendix~\ref{ap:comp_eff_attn}).
Additionally, we incorporate the scaling effects of $f$ and $\VEC{x}$, whereas~\citet{brunner19} have considered only the binary effect---either the weight is canceled or not.

\paragraph{Gradient-based analysis:}
In the gradient analysis, the contribution of the input with respect to the output of the attention mechanism is calculated using the norm of a gradient matrix between the input and the output vector~\cite{pascual20}.
Intuitively, such gradient-based methods measure the change in the output vector with respect to the perturbations in the input vector.
Estimating the contribution of $\VEC a$ to $\VEC b = \sum k\VEC a$ by computing the gradient $\partial \VEC b/\partial \VEC a$ ($=k$) is analogous to estimating the contribution of $\VEC x$ to $\VEC y = \sum \alpha f(\VEC x)$ by observing only an attention weight $\alpha$.%
\footnote{
    For simplicity, we consider a linear example: $\VEC b = \sum k\VEC a$.
    We are aware that there is a gap between the two examples in terms of linearity. 
    Further exploration of the connection to the gradient-based method is needed.
}
The two approaches have the same kind of problems; that is, both ignore the magnitude of the input, $\VEC a$ or $f(\VEC x)$.
\section{Conclusions and future work}
This paper showed that attention weights alone are only one of two factors that determine the output of attention.
We proposed the incorporation of another factor, the transformed input vectors.
Using our norm-based method, we provided a more detailed interpretation of the inner workings of Transformers, compared to the studies using the weight-based analysis.
We hope that this paper will inspire researchers to have a broader view of the possible methodological choices for analyzing the behavior of Transformer-based models.

We believe that these findings can provide insights not only into the interpretation of the behaviors of Blackbox NLP systems but also into developing a more sophisticated Transformer-based system.
One possible direction is to design an attention mechanism that can collect almost no information from an input sequence as the current systems achieve it by exploiting the \texttt{[SEP]} token.

In future work, we plan to apply our norm-based analysis to attention in other models, such as fine-tuned BERT, RoBERTa~\cite{liu19}, and ALBERT~\cite{lan2020albert}.
Furthermore, we expect to extend the scope of analysis from the attention to an entire Transformer architecture to better understand the inner workings and linguistic capabilities of the current powerful systems in NLP.

\newpage

\section*{Acknowledgments}
We would like to thank the anonymous reviewers of the EMNLP 2020 and the ACL 2020 Student Research Workshop (SRW), and the SRW mentor Junjie Hu for their insightful comments.
We also thank the members of Tohoku NLP Laboratory for helpful comments.
This work was supported by JSPS KAKENHI Grant Number JP19H04162.
This work was also partially supported by a Bilateral Joint Research Program between RIKEN AIP Center and Tohoku University.

\bibliographystyle{acl_natbib}
\bibliography{references}

\begin{thebibliography}{36}
\expandafter\ifx\csname natexlab\endcsname\relax\def\natexlab#1{#1}\fi

\bibitem[{Ba et~al.(2016)Ba, Kiros, and Hinton}]{Ba2016_Layernormalization}
Jimmy~Lei Ba, Jamie~Ryan Kiros, and Geoffrey~E. Hinton. 2016.
\newblock \href {http://arxiv.org/abs/1607.06450} {{Layer Normalization}}.
\newblock \emph{arXiv preprint arXiv:1607.06450}.

\bibitem[{Brunner et~al.(2020)Brunner, Liu, Pascual, Richter, Ciaramita, and
  Wattenhofer}]{brunner19}
Gino Brunner, Yang Liu, Dami{\'{a}}n Pascual, Oliver Richter, Massimiliano
  Ciaramita, and Roger Wattenhofer. 2020.
\newblock \href {https://openreview.net/forum?id=BJg1f6EFDB} {{On
  Identifiability in Transformers}}.
\newblock In \emph{8th International Conference on Learning Representations
  (ICLR)}.

\bibitem[{Clark et~al.(2019)Clark, Khandelwal, Levy, and Manning}]{clark19}
Kevin Clark, Urvashi Khandelwal, Omer Levy, and Christopher~D Manning. 2019.
\newblock \href {https://doi.org/10.18653/v1/W19-4828} {{What Does BERT Look
  At? An Analysis of BERT's Attention}}.
\newblock In \emph{Proceedings of the 2019 ACL Workshop BlackboxNLP: Analyzing
  and Interpreting Neural Networks for NLP}, pages 276--286.

\bibitem[{Devlin et~al.(2019)Devlin, Chang, Lee, and
  Toutanova}]{devlin2018bert}
Jacob Devlin, Ming-Wei Chang, Kenton Lee, and Kristina Toutanova. 2019.
\newblock \href {https://doi.org/10.18653/v1/N19-1423} {{BERT: Pre-training of
  Deep Bidirectional Transformers for Language Understanding}}.
\newblock In \emph{Proceedings of the 2019 Conference of the North American
  Chapter of the Association for Computational Linguistics: Human Language
  Technologies (NAACL-HLT)}, pages 4171--4186.

\bibitem[{Ding et~al.(2019)Ding, Xu, and Koehn}]{ding19-saliency}
Shuoyang Ding, Hainan Xu, and Philipp Koehn. 2019.
\newblock \href {https://doi.org/10.18653/v1/W19-5201} {{Saliency-driven Word
  Alignment Interpretation for Neural Machine Translation}}.
\newblock In \emph{Proceedings of the 4th Conference on Machine Translation
  (WMT)}, pages 1--12.

\bibitem[{Dyer et~al.(2013)Dyer, Chahuneau, and Smith}]{dyer13-fastalign}
Chris Dyer, Victor Chahuneau, and Noah~A Smith. 2013.
\newblock \href {https://www.aclweb.org/anthology/N13-1073} {{A Simple, Fast,
  and Effective Reparameterization of IBM Model 2}}.
\newblock In \emph{Proceedings of the 2013 Conference of the North American
  Chapter of the Association for Computational Linguistics: Human Language
  Technologies (NACCL-HLT)}, pages 644--648.

\bibitem[{Goldberg(2019)}]{goldberg19}
Yoav Goldberg. 2019.
\newblock \href {http://arxiv.org/abs/1901.05287} {{Assessing BERT's Syntactic
  Abilities}}.
\newblock \emph{arXiv preprint arXiv:1901.05287}.

\bibitem[{Htut et~al.(2019)Htut, Phang, Bordia, and Bowman}]{Htut2019}
Phu~Mon Htut, Jason Phang, Shikha Bordia, and Samuel~R. Bowman. 2019.
\newblock \href {http://arxiv.org/abs/1911.12246} {{Do Attention Heads in BERT
  Track Syntactic Dependencies?}}
\newblock \emph{arXiv preprint arXiv:1911.12246}.

\bibitem[{Jain and Wallace(2019)}]{jain18}
Sarthak Jain and Byron~C Wallace. 2019.
\newblock \href {https://doi.org/10.18653/v1/N19-1357} {{Attention is not
  Explanation}}.
\newblock In \emph{Proceedings of the 2019 Conference of the North American
  Chapter of the Association for Computational Linguistics: Human Language
  Technologies (NAACL-HLT)}, pages 3543--3556.

\bibitem[{Jawahar et~al.(2019)Jawahar, Sagot, and Seddah}]{jawahar19}
Ganesh Jawahar, Beno{\^{i}}t Sagot, and Djam{\'{e}} Seddah. 2019.
\newblock \href {https://doi.org/10.18653/v1/P19-1356} {{What Does BERT Learn
  about the Structure of Language?}}
\newblock In \emph{Proceedings of the 57th Annual Meeting of the Association
  for Computational Linguistics (ACL)}, pages 3651--3657.

\bibitem[{Kovaleva et~al.(2019)Kovaleva, Romanov, Rogers, and
  Rumshisky}]{kovaleva19}
Olga Kovaleva, Alexey Romanov, Anna Rogers, and Anna Rumshisky. 2019.
\newblock \href {https://doi.org/10.18653/v1/D19-1445} {{Revealing the Dark
  Secrets of BERT}}.
\newblock In \emph{Proceedings of the 2019 Conference on Empirical Methods in
  Natural Language Processing and the 9th International Joint Conference on
  Natural Language Processing (EMNLP-IJCNLP)}, pages 4364--4373.

\bibitem[{Lan et~al.(2020)Lan, Chen, Goodman, Gimpel, Sharma, and
  Soricut}]{lan2020albert}
Zhenzhong Lan, Mingda Chen, Sebastian Goodman, Kevin Gimpel, Piyush Sharma, and
  Radu Soricut. 2020.
\newblock \href {https://openreview.net/forum?id=H1eA7AEtvS} {{ALBERT: A Lite
  BERT for Self-supervised Learning of Language Representations}}.
\newblock In \emph{8th International Conference on Learning Representations
  (ICLR)}.

\bibitem[{Li et~al.(2019)Li, Li, Liu, Meng, and Shi}]{li19-word_align}
Xintong Li, Guanlin Li, Lemao Liu, Max Meng, and Shuming Shi. 2019.
\newblock \href {https://doi.org/10.18653/v1/P19-1124} {{On the Word Alignment
  from Neural Machine Translation}}.
\newblock In \emph{Proceedings of the 57th Annual Meeting of the Association
  for Computational Linguistics (ACL)}, pages 1293--1303.

\bibitem[{Lin et~al.(2019)Lin, Tan, and Frank}]{lin19}
Yongjie Lin, Yi~Chern Tan, and Robert Frank. 2019.
\newblock \href {https://doi.org/10.18653/v1/W19-4825} {{Open Sesame: Getting
  Inside BERT's Linguistic Knowledge}}.
\newblock \emph{Proceedings of the 2019 ACL Workshop BlackboxNLP: Analyzing and
  Interpreting Neural Networks for NLP}, pages 241--253.

\bibitem[{Liu et~al.(2019)Liu, Ott, Goyal, Du, Joshi, Chen, Levy, Lewis,
  Zettlemoyer, and Stoyanov}]{liu19}
Yinhan Liu, Myle Ott, Naman Goyal, Jingfei Du, Mandar Joshi, Danqi Chen, Omer
  Levy, Mike Lewis, Luke Zettlemoyer, and Veselin Stoyanov. 2019.
\newblock \href {http://arxiv.org/abs/1907.11692} {{RoBERTa: A Robustly
  Optimized BERT Pretraining Approach}}.
\newblock \emph{arXiv preprint arXiv:1907.11692}.

\bibitem[{Mare{\v{c}}ek and Rosa(2019)}]{marecek19balustrades}
David Mare{\v{c}}ek and Rudolf Rosa. 2019.
\newblock \href {https://doi.org/10.18653/v1/W19-4827} {{From Balustrades to
  Pierre Vinken: Looking for Syntax in Transformer Self-Attentions}}.
\newblock In \emph{Proceedings of the 2019 ACL Workshop BlackboxNLP: Analyzing
  and Interpreting Neural Networks for NLP}, pages 263--275.

\bibitem[{Michel et~al.(2019)Michel, Levy, and Neubig}]{michel19}
Paul Michel, Omer Levy, and Graham Neubig. 2019.
\newblock \href
  {http://papers.nips.cc/paper/9551-are-sixteen-heads-really-better-than-one.pdf}
  {{Are Sixteen Heads Really Better than One?}}
\newblock In \emph{Advances in Neural Information Processing Systems 32
  (NIPS)}, pages 14014--14024.

\bibitem[{Och and Ney(2000)}]{och&ney00-aer}
Franz~Josef Och and Hermann Ney. 2000.
\newblock \href {https://doi.org/10.3115/1075218.1075274} {{Improved
  Statistical Alignment Models}}.
\newblock In \emph{Proceedings of the 38th Annual Meeting of the Association
  for Computational Linguistics (ACL)}, pages 440--447.

\bibitem[{Och and Ney(2003)}]{och&ney_giza}
Franz~Josef Och and Hermann Ney. 2003.
\newblock \href {https://doi.org/10.1162/089120103321337421} {{A Systematic
  Comparison of Various Statistical Alignment Models}}.
\newblock \emph{Computational Linguistics}, 29(1):19--51.

\bibitem[{Ott et~al.(2019)Ott, Edunov, Baevski, Fan, Gross, Ng, Grangier, and
  Auli}]{ott2019fairseq}
Myle Ott, Sergey Edunov, Alexei Baevski, Angela Fan, Sam Gross, Nathan Ng,
  David Grangier, and Michael Auli. 2019.
\newblock \href {https://doi.org/10.18653/v1/N19-4009} {{fairseq: A Fast,
  Extensible Toolkit for Sequence Modeling}}.
\newblock In \emph{Proceedings of the 2019 Conference of the North American
  Chapter of the Association for Computational Linguistics (Demonstrations)},
  pages 48--53.

\bibitem[{Pascual et~al.(2020)Pascual, Brunner, and Wattenhofer}]{pascual20}
Damian Pascual, Gino Brunner, and Roger Wattenhofer. 2020.
\newblock \href {http://arxiv.org/abs/2004.05916} {{Telling BERT's full story:
  from Local Attention to Global Aggregation}}.
\newblock \emph{arXiv preprint arXiv:2004.05916}.

\bibitem[{Pruthi et~al.(2020)Pruthi, Gupta, Dhingra, Neubig, and
  Lipton}]{pruthi19}
Danish Pruthi, Mansi Gupta, Bhuwan Dhingra, Graham Neubig, and Zachary~C
  Lipton. 2020.
\newblock \href {http://arxiv.org/abs/1909.07913} {{Learning to Deceive with
  Attention-Based Explanations}}.
\newblock In \emph{Proceedings of the 58th Annual Meeting of the Association
  for Computational Linguistics (ACL)}.

\bibitem[{Raganato and Tiedemann(2018)}]{raganato2018}
Alessandro Raganato and J{\"{o}}rg Tiedemann. 2018.
\newblock \href {https://doi.org/10.18653/v1/W18-5431} {{An Analysis of Encoder
  Representations in Transformer-Based Machine Translation}}.
\newblock In \emph{Proceedings of the 2018 EMNLP Workshop BlackboxNLP:
  Analyzing and Interpreting Neural Networks for NLP}, pages 287--297.

\bibitem[{Reif et~al.(2019)Reif, Yuan, Wattenberg, Viegas, Coenen, Pearce, and
  Kim}]{coenen19}
Emily Reif, Ann Yuan, Martin Wattenberg, Fernanda~B Viegas, Andy Coenen, Adam
  Pearce, and Been Kim. 2019.
\newblock \href
  {http://papers.nips.cc/paper/9065-visualizing-and-measuring-the-geometry-of-bert}
  {{Visualizing and Measuring the Geometry of BERT}}.
\newblock \emph{Advances in Neural Information Processing Systems 32 (NIPS)},
  pages 8594--8603.

\bibitem[{Rogers et~al.(2020)Rogers, Kovaleva, and Rumshisky}]{bertology}
Anna Rogers, Olga Kovaleva, and Anna Rumshisky. 2020.
\newblock \href {http://arxiv.org/abs/2002.12327} {{A Primer in BERTology: What
  we know about how BERT works}}.
\newblock \emph{arXiv preprint arXiv:2002.12327}.

\bibitem[{Sennrich et~al.(2016)Sennrich, Haddow, and Birch}]{sennrich16_bpe}
Rico Sennrich, Barry Haddow, and Alexandra Birch. 2016.
\newblock \href {https://doi.org/10.18653/v1/P16-1162} {{Neural Machine
  Translation of Rare Words with Subword Units}}.
\newblock In \emph{Proceedings of the 54th Annual Meeting of the Association
  for Computational Linguistics (ACL)}, pages 1715--1725.

\bibitem[{Serrano and Smith(2019)}]{serrano19}
Sofia Serrano and Noah~A Smith. 2019.
\newblock \href {https://doi.org/10.18653/v1/P19-1282} {{Is Attention
  Interpretable?}}
\newblock In \emph{Proceedings of the 57th Annual Meeting of the Association
  for Computational Linguistics (ACL)}, pages 2931--2951.

\bibitem[{Tang et~al.(2018)Tang, Sennrich, and
  Nivre}]{tang_transformer_attention_analysis}
Gongbo Tang, Rico Sennrich, and Joakim Nivre. 2018.
\newblock \href {https://doi.org/10.18653/v1/W18-6304} {{An Analysis of
  Attention Mechanisms: The Case of Word Sense Disambiguation in Neural Machine
  Translation}}.
\newblock In \emph{Proceedings of the 3rd Conference on Machine Translation
  (WMT): Research Papers}, pages 26--35.

\bibitem[{Tenney et~al.(2019)Tenney, Xia, Chen, Wang, Poliak, McCoy, Kim,
  Durme, Bowman, Das, and Pavlick}]{tenney19}
Ian Tenney, Patrick Xia, Berlin Chen, Alex Wang, Adam Poliak, R~Thomas McCoy,
  Najoung Kim, Benjamin~Van Durme, Samuel~R Bowman, Dipanjan Das, and Ellie
  Pavlick. 2019.
\newblock \href {https://openreview.net/forum?id=SJzSgnRcKX} {{What do you
  learn from context? Probing for sentence structure in contextualized word
  representations}}.
\newblock In \emph{7th International Conference on Learning Representations
  (ICLR)}.

\bibitem[{Vashishth et~al.(2019)Vashishth, Upadhyay, Tomar, and
  Faruqui}]{Vashishth2019}
Shikhar Vashishth, Shyam Upadhyay, Gaurav~Singh Tomar, and Manaal Faruqui.
  2019.
\newblock \href {http://arxiv.org/abs/1909.11218} {{Attention Interpretability
  Across NLP Tasks}}.
\newblock \emph{arXiv preprint arXiv:1909.11218}.

\bibitem[{Vaswani et~al.(2017)Vaswani, Shazeer, Parmar, Uszkoreit, Jones,
  Gomez, Kaiser, and Polosukhin}]{vaswani17}
Ashish Vaswani, Noam Shazeer, Niki Parmar, Jakob Uszkoreit, Llion Jones,
  Aidan~N Gomez, Lukasz Kaiser, and Illia Polosukhin. 2017.
\newblock \href {http://papers.nips.cc/paper/7181-attention-is-all-you-need}
  {{Attention is All you Need}}.
\newblock In \emph{Advances in Neural Information Processing Systems 30
  (NIPS)}, pages 5998--6008.

\bibitem[{Vilar et~al.(2006)Vilar, Popovi{\'{c}}, and Ney}]{vilar2006_rwth}
David Vilar, Maja Popovi{\'{c}}, and Hermann Ney. 2006.
\newblock \href
  {https://www.isca-speech.org/archive/iwslt{\_}06/slt6{\_}205.html} {{AER: Do
  we need to “improve” our alignments?}}
\newblock In \emph{International Workshop on Spoken Language Translation
  (IWSLT) 2006}, pages 205--212.

\bibitem[{Wallace et~al.(2019)Wallace, Wang, Li, Singh, and
  Gardner}]{wallace19_number}
Eric Wallace, Yizhong Wang, Sujian Li, Sameer Singh, and Matt Gardner. 2019.
\newblock \href {https://doi.org/10.18653/v1/D19-1534} {{Do NLP Models Know
  Numbers? Probing Numeracy in Embeddings}}.
\newblock In \emph{Proceedings of the 2019 Conference on Empirical Methods in
  Natural Language Processing and the 9th International Joint Conference on
  Natural Language Processing (EMNLP-IJCNLP)}, pages 5307--5315.

\bibitem[{Wiegreffe and Pinter(2019)}]{wiegreffe19}
Sarah Wiegreffe and Yuval Pinter. 2019.
\newblock \href {https://doi.org/10.18653/v1/D19-1002} {{Attention is not not
  Explanation}}.
\newblock In \emph{Proceedings of the 2019 Conference on Empirical Methods in
  Natural Language Processing and the 9th International Joint Conference on
  Natural Language Processing (EMNLP-IJCNLP)}, pages 11--20.

\bibitem[{Yang et~al.(2019)Yang, Dai, Yang, Carbonell, Salakhutdinov, and
  Le}]{yang19xlnet}
Zhilin Yang, Zihang Dai, Yiming Yang, Jaime Carbonell, Ruslan Salakhutdinov,
  and Quoc~V. Le. 2019.
\newblock \href
  {http://papers.nips.cc/paper/8812-xlnet-generalized-autoregressive-pretraining-for-language-understanding}
  {{XLNet: Generalized Autoregressive Pretraining for Language Understanding}}.
\newblock In \emph{Advances in Neural Information Processing Systems 32
  (NIPS)}, pages 1--18.

\bibitem[{Zenkel et~al.(2019)Zenkel, Wuebker, and DeNero}]{zenkel_aer_giza}
Thomas Zenkel, Joern Wuebker, and John DeNero. 2019.
\newblock \href {http://arxiv.org/abs/1901.11359} {{Adding Interpretable
  Attention to Neural Translation Models Improves Word Alignment}}.
\newblock \emph{arXiv preprint arXiv:1901.11359}.

\end{thebibliography}

\clearpage
\appendix
\section{Multi-head attention and the norm-based analysis}
\label{ap:re-multi}

Our norm-based analysis is applicable to the analysis of the multi-head attention mechanism 
implemented in Transformers.
The $i$-th output of the multi-head attention mechanism $\VEC{y}^{\text{integrated}}_i$ is calculated as follows:

\begin{align}
    \VEC{y}^{\text{integrated}}_i &= \sum_h{\VEC{y}^{h}_i}
    \label{eq:sum_head} \\
    \VEC{y}^{h}_i &=
    \sum^n_{j=1}{\alpha^h_{i,j}f^h(\VEC{x}_j)} %
    \label{eq:transformed_self_attention_head} \\
    f^h(\VEC{x}) &:= \left(\VEC{x}\VEC{W}^{V,h} + \VEC{b}^{V,h}\right)\VEC{W}^{O,h}
    \text{,}
    \label{eq:transform_T_head}
\end{align}

\noindent
where $\alpha^h_{i,j}$, $\VEC{W}^{V,h}$, $\VEC{b}^{V,h}$, and $\VEC{W}^{O,h}$ are the same as $\alpha_{i,j}$, $\VEC{W}^{V}$, $\VEC{b}^{V}$,  and $\VEC{W}^{O}$ in Equations~\ref{eq7:transformed_self_attention} and \ref{eq8:transform_T} for each head $h$, respectively.
$n$ is the number of tokens in the input vectors.
Equation~\ref{eq:sum_head} can be rewritten as follows:

\begin{align}
    \VEC{y}^\text{integrated}_i &=
    \sum^n_{j=1}{\tikz[baseline=(X.base)]{\node(X)[rectangle, fill=purple!9, rounded corners, text height=2.2ex,text depth=1ex]{$\sum_h$ \bgf{\textcolor{blue}{$\alpha^h_{i,j}$}}\bgfg{\textcolor{black}{
    $f^h(\VEC{x}_j)$
    }}}}} %
    \label{eq:transformed_self_attention_decomp}
\end{align}

\noindent
As shown in Equation~\ref{eq:transformed_self_attention_decomp}, the multi-head attention mechanism is also linearly decomposable, and one can analyze the flaw of the information from the $j$-th vector to the $i$-th vector by measuring $\lVert \sum_h\alpha^h_{i,j} f^h(\VEC{x}_j)\rVert$.
In Section~\ref{sec:nmt}, we actually used $\lVert \sum_h\alpha^h_{i,j} f^h(\VEC{x}_j)\rVert$ to extract the alignment from each layer's multi-head attention.

The output of the multi-head attention mechanism is calculated via the sum of the outputs of all the heads and a bias $\VEC{b}^O\in\mathbb{R}^d$.
Because adding a fixed vector is irrelevant to the token-to-token interaction that we aim to investigate, we omitted $\VEC{b}^O$ in our analysis.

\section{The source of the dispersion of $\lVert f(\VEC{x})\rVert$}
\label{ap:x_and_f}
As described in Section \ref{subsec:ignored_effect}, \msfx{} exhibits dispersion; however, it remains unclear whether this dispersion is attributed to $\lVert \VEC{x}\rVert$ or $f$.
Hence, we checked the dispersion of $\lVert \VEC{x}\rVert$ and the scaling effects of the transformation $f$.

\paragraph{Dispersion of $\lVert \VEC{x}\rVert$:}
First, we checked the coefficient of variation (CV) of $\lVert\VEC{x}\rVert$.
Table~\ref{table:x_variance} shows that the average CV is 0.12, which is less than that of \msfx{} (0.22).
The value of $\lVert \VEC{x} \rVert$ typically varies between 0.88 and 1.12 times the average value of $\lVert \VEC{x} \rVert$.
The layer normalization~\citep{Ba2016_Layernormalization} that applied at the end of the previous layer should have a large impact on the variance of $\lVert \VEC{x} \rVert$.

\begin{figure}[t]
    \center
    \includegraphics[width=.85\hsize]{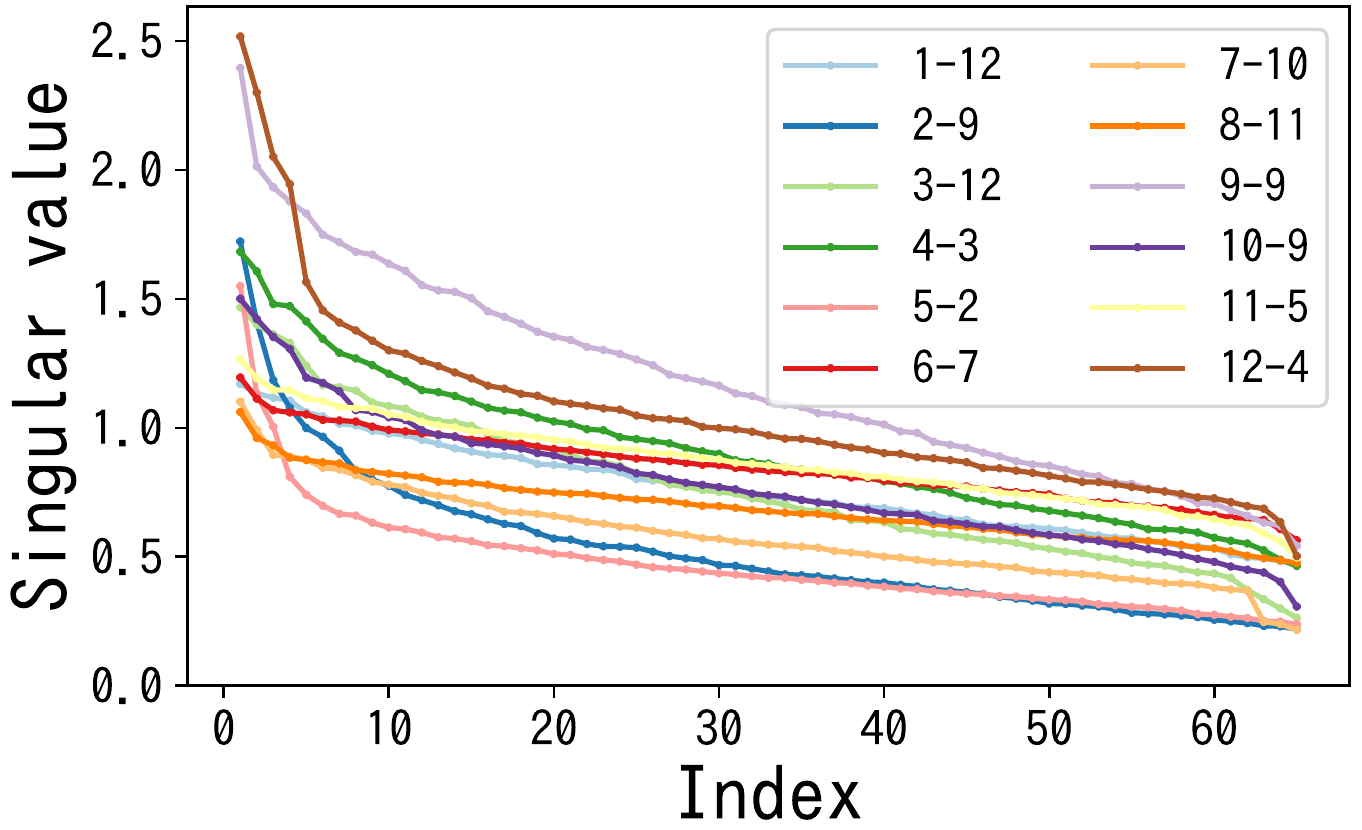}
    \caption{
    Singular values of $f$ at randomly selected heads in each layer.
    We use $\langle$layer$\rangle$-$\langle$head number$\rangle$ to denote a particular attention head.
    The singular values are 
    }
    \label{fig:singular_value}
\end{figure}

\paragraph{Scaling effects of $f$:}
Second, we investigated the scaling effect of the transformation $f$ on the norm of the input.
Because the affine transformation $f\colon \mathbb{R}^d \to \mathbb{R}^d$ can be considered a linear transformation $\mathbb{R}^{d+1} \to \mathbb{R}^{d+1}$ (Appendix \ref{affine_is_linear}), the singular values of $f$ can be regarded as its scaling effect.
Figure~\ref{fig:singular_value} shows the singular values of $f$ in randomly selected heads in BERT.
The singular values are displayed in descending order from left to right.
In each head, there is a difference of at least 1.8 times between the maximum and minimum singular values.
This difference is larger than that of $\lVert \VEC{x} \rVert$, where $\lVert \VEC{x} \rVert$ typically varies between 0.88 and 1.12 times the average value.
These results suggest that the dispersion of \msfx\ is primarily attributed to the scaling effect of $f$.

\section{Affine transformation as linear transformation}
\label{affine_is_linear}
The affine transformation $f\colon \mathbb{R}^d \to \mathbb{R}^d$ in Equation~\ref{eq8:transform_T} can be viewed as a linear transformation $\widetilde{f}\colon \mathbb{R}^{d+1} \to \mathbb{R}^{d+1}$.
Given $\widetilde{\VEC{x}} := \left[ \begin{array}{cccc} & \VEC{x} & & 1 \end{array} \right]\in\mathbb{R}^{d+1}$, where 1 is concatenated to the end of each input vector $\VEC{x}\in \mathbb{R}^d$, the affine transformation $f$ can be viewed as:
\begin{align}
&\widetilde{f}(\widetilde{\VEC{x}}) = \widetilde{\VEC{x}}\widetilde{\VEC{W}}^V\widetilde{\VEC{W}}^O \\
&\widetilde{\VEC{W}}^V := \left[
\begin{array}{cccc}
  & & & 0 \\
  & \VEC{W}^V & & \vdots \\
  & & & 0 \\
  & \VEC{b}^V & & 1
\end{array}
\right]\in\mathbb{R}^{(d+1)\times(d'+1)} \\
&\widetilde{\VEC{W}}^O := \left[
\begin{array}{cccc}
  & & & 0 \\
  & \VEC{W}^O & & \vdots \\
  & & & 0 \\
  0 & \ldots & 0 & 1
\end{array}
\right]\in\mathbb{R}^{(d'+1)\times(d+1)}
\text{.}
\end{align}
\begin{table}[t]
\centering
\setlength{\tabcolsep}{4pt}  %
\renewcommand{\arraystretch}{0.9}
    {\small
\begin{tabular}{cccccc}
\toprule
Layer & $\mu$ & $\sigma$ & CV & Max & Min \\
\cmidrule(r){1-1} \cmidrule(lr){2-2} \cmidrule(lr){3-3} \cmidrule(lr){4-4} \cmidrule(lr){5-5} \cmidrule(l){6-6} 
12 (max CV) & 20.49 & 4.62 & \textbf{0.23} & 32.84 & 4.13 \\
7 (min CV) & 21.64 & 1.40 & \textbf{0.06} & 23.03 & 11.87 \\
Average & 19.93 & 2.39 & \textbf{0.12} & - & - \\ \bottomrule
\end{tabular}
}
\caption{Mean (μ), standard deviation (σ), coefficient of variance (CV), and maximum and minimum values of $\lVert\mathbf{x}\rVert$; the former three are averaged on all the layers.
}
\label{table:x_variance}
\end{table}

\section{Details on Sections~\ref{subsec:re-examin} and~\ref{sec:freq_and_fx}}
\label{ap:re-examin}
We describe the detailed experimental setup presented in Sections~\ref{subsec:re-examin} and~\ref{sec:freq_and_fx}.

\subsection{Notations}
The dataset consists of several sequences; $\text{Data} = (s_1,\cdots,s_{|Data|})$. Each sequence consists of several tokens, $s_p = (t^{p}_1,\cdots,t^p_{\lvert s_p\rvert})$, where $t^p_q$ is the $q$-th token in the $p$-th sequence.
For simplicity, we define the following functions:

\begin{align}
    \nonumber
    \text{Weight}(p, q, \ell, h) &= \frac{1}{\lvert s_p\rvert} \sum_{i=1}^{\lvert s_p\rvert} \alpha_{p,i,q}^{\ell,h} \\
    \nonumber
    \text{Norm}(p, q, \ell, h) &= \lVert f^{\ell,h}(\VEC{x}_{p,q}^\ell) \rVert\ \\
    \nonumber
    \text{WNorm}(p,q,\ell,h) &= \frac{1}{\lvert s_p\rvert} \sum_{i=1}^{\lvert s_p\rvert} \lVert \alpha_{p,i,q}^{\ell,h}f^{\ell,h}(\VEC{x}_{p,q}^\ell) \rVert 
    \text{,}
\end{align}

\noindent
where $\alpha_{p,i,q}^{\ell,h}$ is the attention weight assigned from the $i$-th pre-update vector to the $q$-th input vector in the $p$-th sequence.
$h$ and $\ell$ denote that the score is obtained from the $h$-th head of the $\ell$-th layer.
$\VEC{x}_{p,q}^\ell$ denotes the input vector for token $t^p_q$ in the $\ell$-th layer.
$f^{\ell,h}(\VEC{x}_{p,q}^\ell)$ is the transformed vector for $\VEC{x}_{p,q}^\ell$ in the $h$-th head of the $\ell$-th layer.

Next, the vocabulary $\mathcal V$ of BERT is divided into the following four categories: 
\begin{align}
    \nonumber
    A &= \{\text{\texttt{[CLS]}}\} \\
    \nonumber
    B &= \{\text{\texttt{[SEP]}}\} \\
    \nonumber
    C &= \{\text{``.''}, \text{``,''}\} \\
    D &= \mathcal V \setminus (A \cup B \cup C)
    \label{eq:category}
    \text{.}
\end{align}

\noindent
Let $T(p,Z)$ be a function that returns all tokens $t^p_q$ belonging to the category $Z$ in the $p$-th sequence.
To formally describe our experiments, several functions are defined as follows.
Note that we analyzed a model with 12 heads in each layer.

\small
\begin{align}
    \nonumber
    \text{MeanN}(Z,\ell,h,p) &= \frac{1}{\lvert T(Z,p) \rvert} \sum_{t^p_q \in T(Z,p)} \text{Norm}(p,q,\ell,h) \\
    \nonumber
    \text{SumW}(Z,\ell,h,p) &= \sum_{t^p_q \in T(Z,p)} \text{Weight}(p,q,\ell,h) \\
    \nonumber
    \text{SumWN}(Z,\ell,h,p) &= \sum_{t^p_q \in T(Z,p)} \text{WNorm}(p,q,\ell,h) \\
    \nonumber
    \text{HeadN}(Z,\ell,h) &= \frac{1}{\lvert \text{Data}\rvert} \sum_{s_p \in \text{Data}} \text{MeanN}(Z,\ell,h,p) \\
    \nonumber
    \text{HeadW}(Z,\ell,h) &= \frac{1}{\lvert \text{Data}\rvert} \sum_{s_p \in \text{Data}} \text{SumW}(Z,\ell,h,p) \\
    \nonumber
    \text{HeadWN}(Z,\ell,h) &= \frac{1}{\lvert \text{Data}\rvert} \sum_{s_p \in \text{Data}} \text{SumWN}(Z,\ell,h,p) \\
    \nonumber
    \text{LayerW}(Z,\ell) &= \frac{1}{12} \sum_{h = 1}^{12} \text{HeadW}(Z,\ell,h) \\
    \nonumber
    \text{LayerWN}(Z,\ell) &= \frac{1}{12} \sum_{h = 1}^{12} \text{HeadWN}(Z,\ell,h)
    \text{.}
 \end{align}
\normalsize
\noindent
The LayerW$(\cdot)$ and LayerWN$(\cdot)$ functions are used to analyze the average behavior of the heads in a layer.

\subsection{Experimental setup for Section~\ref{subsec:re-examin}}
\label{ap:setting:a-fx}
In Figure~\ref{fig:alpha_at_comparison}, the results of each layer are reported for each category.
In Figures~\ref{fig:clark_a} and~\ref{fig:clark_afx}, the values for each category $Z$ were calculated using LayerW$(Z,\ell)$ and LayerWN$(Z,\ell)$, respectively.

In Figure~\ref{fig:sep_detail_analysis}, \malpha\ and \msfx\ in the $h$-th head of the $\ell$-th layer were calculated using HeadW$(Z, \ell, h)$ and HeadN$(Z, \ell, h)$, respectively.
The scores reported in Table~\ref{table:rank_corr} are the Spearman rank correlation coefficient $r$ between Weight$(p,q,\ell,h)$ and WNorm$(p,q,\ell,h)$.
We calculated the $r$ using all the pairs of Weight$(p,q,\ell,h)$ and WNorm$(p,q,\ell,h)$ for the possible values of $p$, $q$, $\ell$, and $h$.
In Figure~\ref{fig:relation_alpha-fx}, each plot corresponds to the pair of Weight$(p,q,\ell,h)$ and WNorm$(p,q,\ell,h)$, where the combination of $(p, q, \ell, h)$ was randomly determined.

\subsection{Visualizations of \malpha{} and \msfx{} for each word category}
\label{ap:detail:a-fx}
\begin{figure}[t]
    \centering
    \begin{minipage}[t]{.47\hsize}
        \centering
        \includegraphics[width=\hsize]{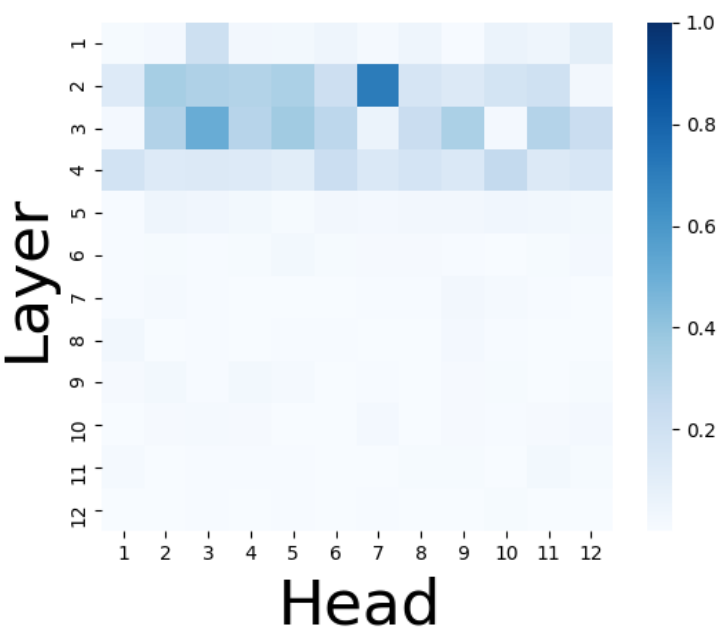}
        \subcaption{
        \malpha{}.
        }
        \label{fig:cls_weight}
    \end{minipage}
    \;
    \begin{minipage}[t]{.47\hsize}
        \centering
        \includegraphics[width=\hsize]{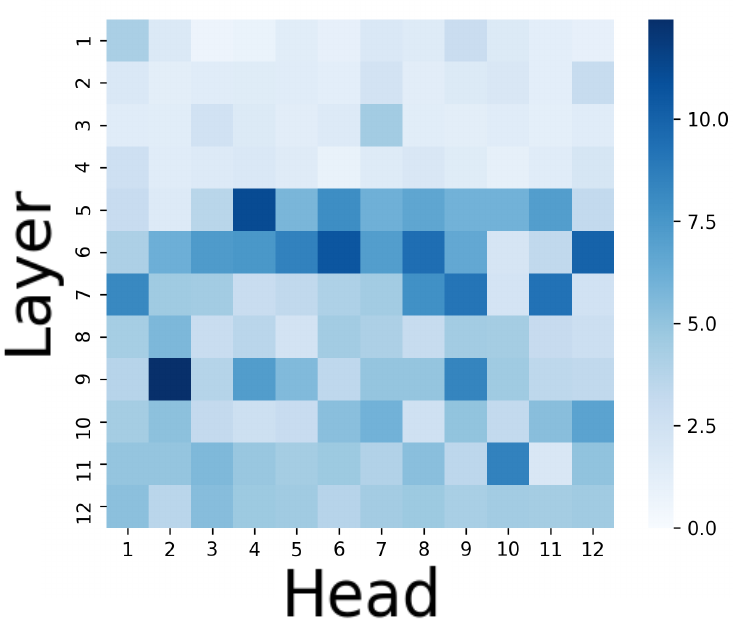}
        \subcaption{
        \msfx{}.
        }
        \label{fig:cls_xt}
    \end{minipage}
    \caption{
    $\alpha$ and \msfx{} corresponding to \texttt{[CLS]} token, averaged on all the input text.}
    \label{fig:cls_detail_analysis}
\end{figure}

\begin{figure}[t]
    \centering
    \begin{minipage}[t]{.47\hsize}
        \centering
        \includegraphics[width=\hsize]{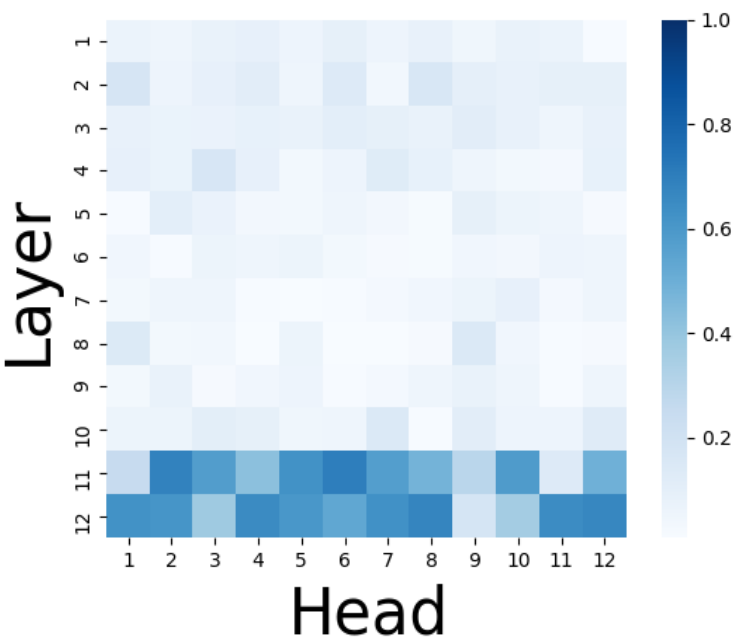}
        \subcaption{
        \malpha{}.
        }
        \label{fig:per_com_weight}
    \end{minipage}
    \;
    \begin{minipage}[t]{.47\hsize}
        \centering
        \includegraphics[width=\hsize]{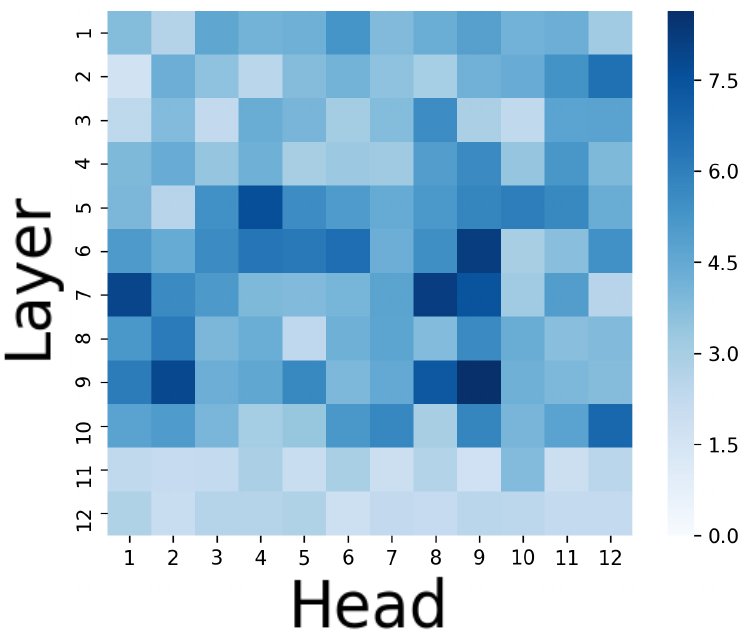}
        \subcaption{
        \msfx{}.
        }
        \label{fig:per_com_xt}
    \end{minipage}
    \caption{
    $\alpha$ and \msfx{} corresponding to periods and commas, averaged on all the input text.}
    \label{fig:per_com_detail_analysis}
\end{figure}

\begin{figure}[t]
    \centering
    \begin{minipage}[t]{.47\hsize}
        \centering
        \includegraphics[width=\hsize]{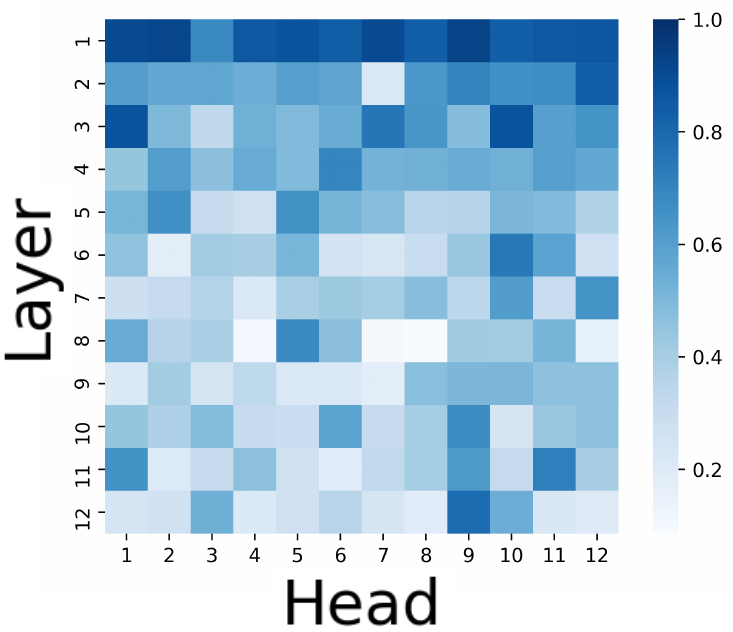}
        \subcaption{
        \malpha{}.
        }
        \label{fig:other_weight}
    \end{minipage}
    \;
    \begin{minipage}[t]{.47\hsize}
        \centering
        \includegraphics[width=\hsize]{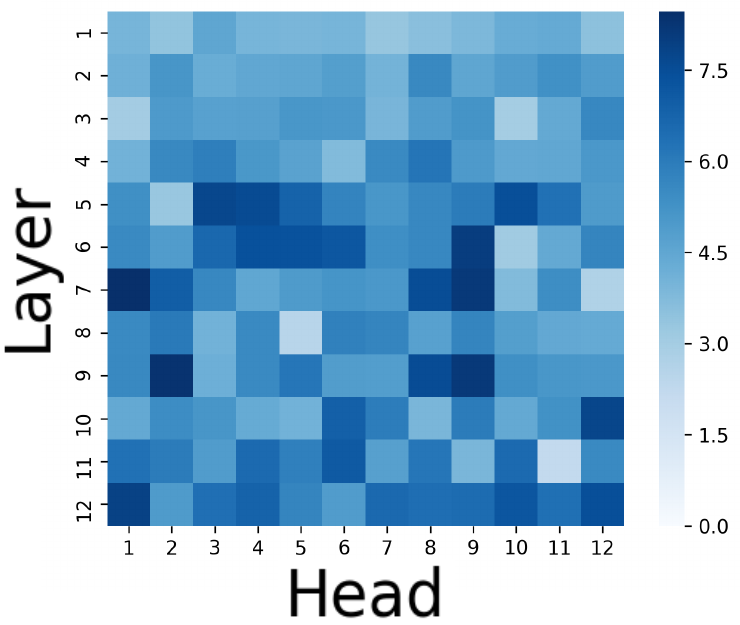}
        \subcaption{
        \msfx{}.
        }
        \label{fig:other_xt}
    \end{minipage}
    \caption{
    $\alpha$ and \msfx{} corresponding to other tokens, averaged on all the input text.
    }
    \label{fig:other_detail_analysis}
\end{figure}

As described in Section~\ref{subsec:re-examin}, \malpha{} and \msfx{} for the \texttt{[SEP]} token were canceled out in almost all heads (Figure~\ref{fig:sep_detail_analysis}).
Here, we show the trends for the other categories---$B$, $C$, and $D$ in Equation~\ref{eq:category}.
Figures~\ref{fig:cls_detail_analysis},~\ref{fig:per_com_detail_analysis}, and~\ref{fig:other_detail_analysis} show the trends of \malpha{} and \msfx{} for category $B$ (the \texttt{[CLS]} token), $C$ (periods and commas), and $D$ (other tokens), respectively.
The values in these figures were calculated as described in Appendix~\ref{ap:setting:a-fx}.
Figures~\ref{fig:cls_detail_analysis} and~\ref{fig:per_com_detail_analysis} show that the trends for categories $B$ and $C$ were analogous to those for the \texttt{[SEP]} token; the large \malpha\ was canceled by the small  \msfx.
However, the trends for category $D$ do not exhibit the trends of the negative correlation between \malpha\ and \msfx.
In each heatmap of \msfx{}, the color scale is determined by the maximum value of \msfx{} in each category.

\begin{figure}[t]
    \centering
    \begin{minipage}[t]{.47\hsize}
        \centering
        \includegraphics[width=\hsize]{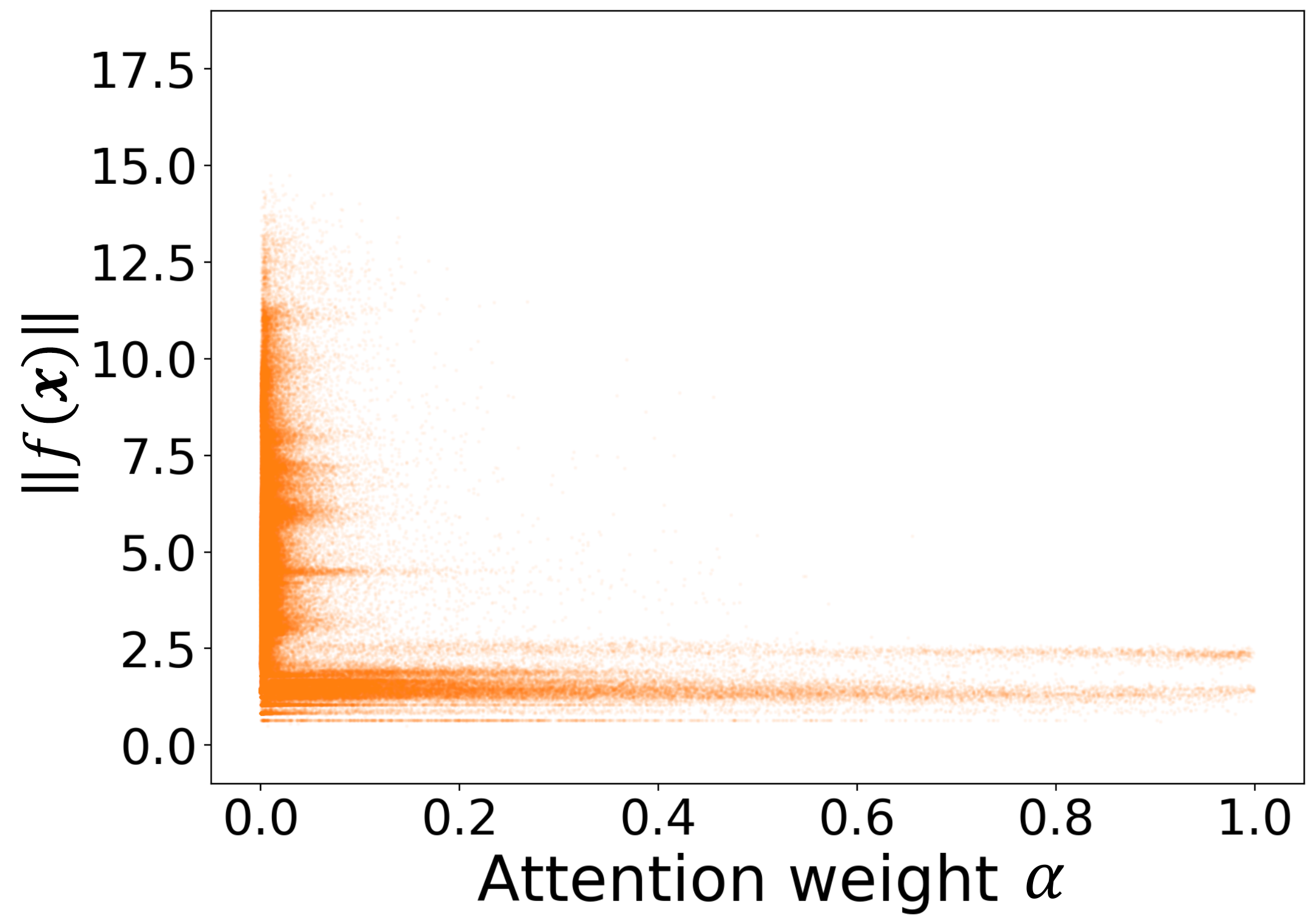}
        \subcaption{
        \texttt{[CLS]}.
        }
    \end{minipage}
    \;
    \begin{minipage}[t]{.47\hsize}
        \centering
        \includegraphics[width=\hsize]{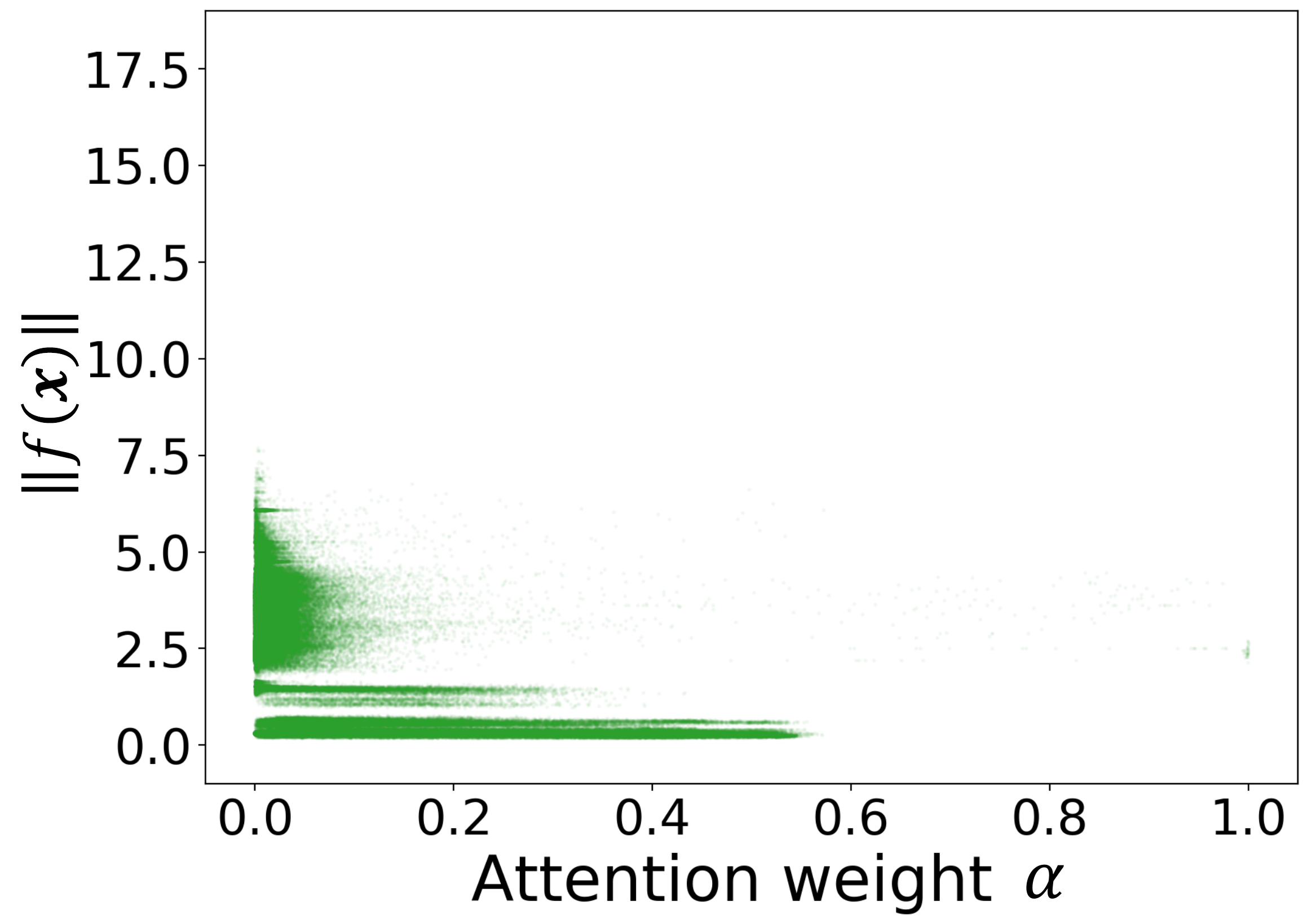}
        \subcaption{
        \texttt{[SEP]}. 
        }
    \end{minipage}
    \begin{minipage}[t]{.47\hsize}
        \centering
        \includegraphics[width=\hsize]{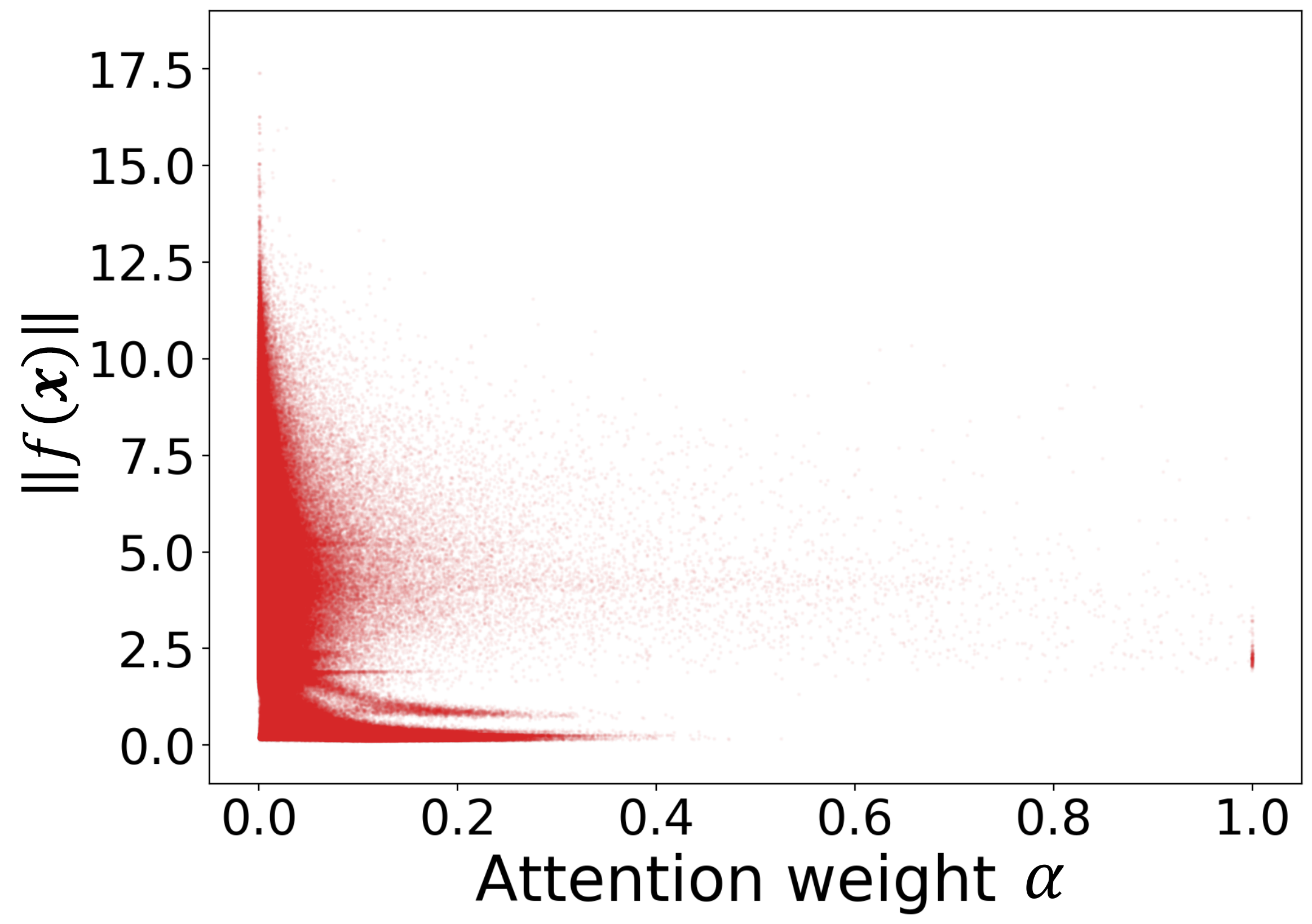}
        \subcaption{
        Periods and commas.
        }
    \end{minipage}
    \;
    \begin{minipage}[t]{.47\hsize}
        \centering
        \includegraphics[width=\hsize]{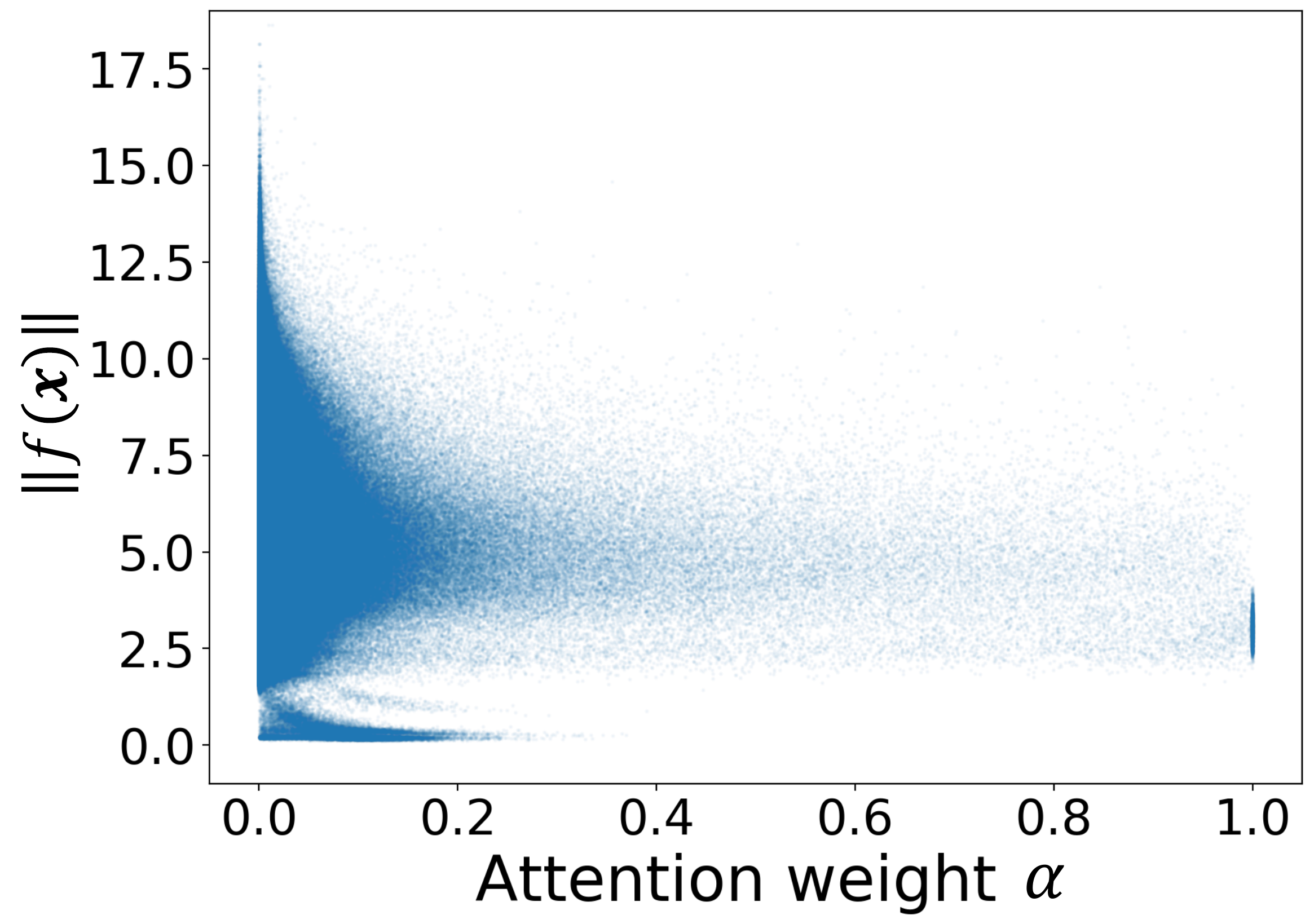}
        \subcaption{
        Other tokens. 
        }
    \end{minipage}
    \caption{
    Relationship between \malpha\ and \msfx\ for each category. 
    }
    \label{fig:alpha-fx_each_category}
\end{figure}

We also reported the relationship between \malpha{} and \msfx{} in Section~\ref{subsec:re-examin} (Figure~\ref{fig:relation_alpha-fx}). 
Figure~\ref{fig:alpha-fx_each_category} shows the results for each word category to provide a clearer display of the results.

\subsection{Experimental setup and visualizations for Section~\ref{sec:freq_and_fx}}
\label{ap:freq}

\begin{figure}[t]
    \centering
    \begin{minipage}[t]{\hsize}
        \centering
        \includegraphics[height=4.0cm]{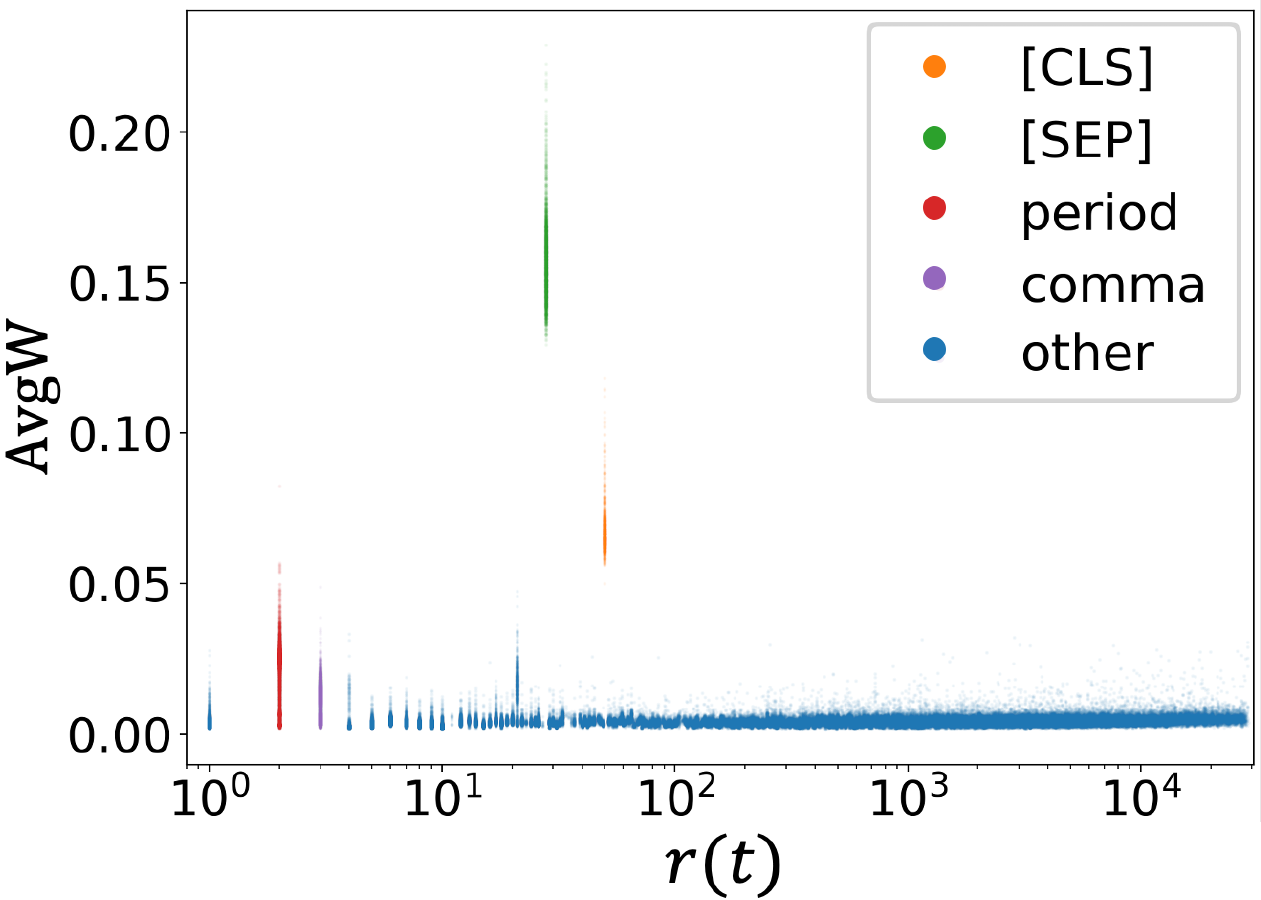}
        \subcaption{
        Relationship between $r(t)$ and AvgW.
        }
        \label{fig:relation_ids-alpha}
    \end{minipage}
    \;
    \begin{minipage}[t]{\hsize}
        \centering
        \includegraphics[height=4.0cm]{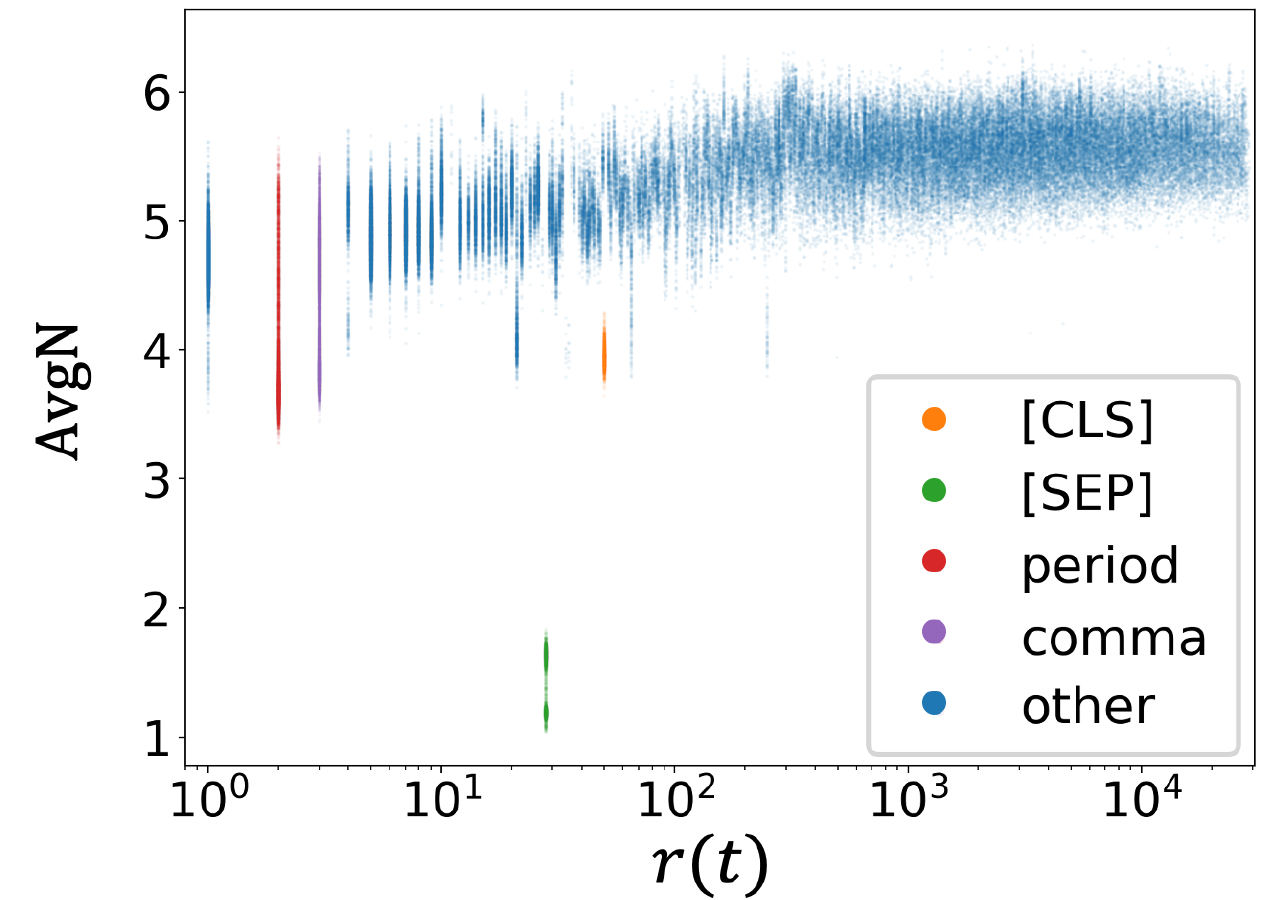}
        \subcaption{
        Relationship between $r(t)$ and AvgN.
        }
        \label{fig:relation_ids-fx}
    \end{minipage}
    \caption{Relationship between frequency rank $r(t^p_q)$ and
    AvgW$(p,q)$, and that between $r(t^p_q)$ and AvgN$(p,q)$.
    }
    \label{fig:relation_ids}
\end{figure}

\begin{figure}[t]
    \centering
    \begin{minipage}[t]{\hsize}
        \centering
        \includegraphics[height=4.0cm]{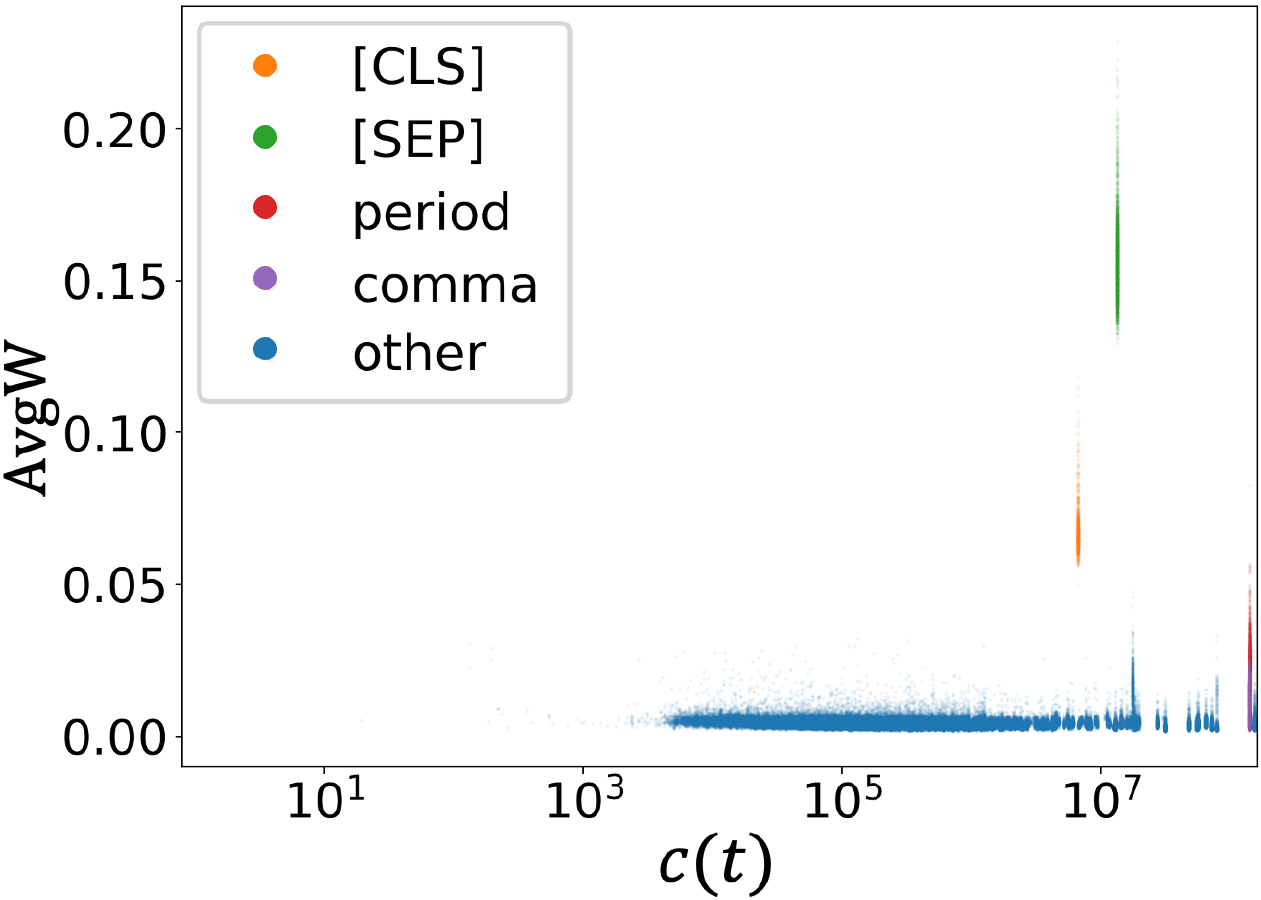}
        \subcaption{
        Relationship between $c(t)$ and AvgW.
        }
        \label{fig:relation_count-alpha}
    \end{minipage}
    \;
    \begin{minipage}[t]{\hsize}
        \centering
        \includegraphics[height=4.0cm]{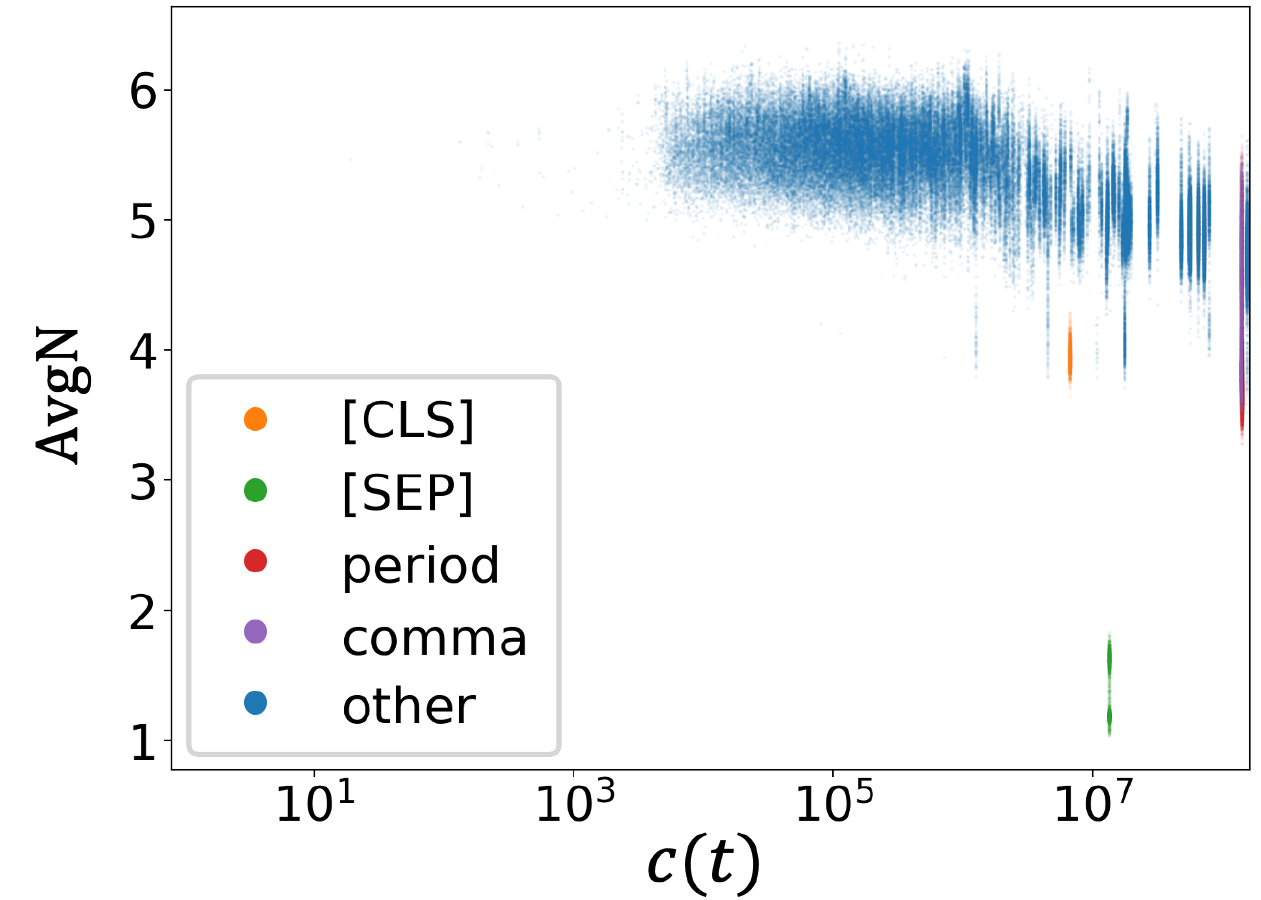}
        \subcaption{
        Relationship between $c(t)$ and AvgN.
        }
        \label{fig:relation_count-fx}
    \end{minipage}
    \caption{
    Relationship between frequency count $c(t^p_q)$ and
    AvgW$(p,q)$, and that between $c(t^p_q)$ and AvgN$(p,q)$.
    }
    \label{fig:relation_count}
\end{figure}

In Section~\ref{sec:freq_and_fx}, we analyzed the relationship between the word frequency and \msfx{}.
To formally describe our experiments, we further define the functions as follows:

\small
\begin{align}
    \nonumber
    \text{AvgW}(p,q) &= \frac{1}{12 \cdot 12} \sum_{\ell = 1}^{12} \sum_{h = 1}^{12} \text{Weight}(p,q,\ell,h) \\
    \nonumber
    \text{AvgN}(p,q) &= \frac{1}{12 \cdot 12} \sum_{\ell = 1}^{12} \sum_{h = 1}^{12} \text{Norm}(p,q,\ell,h) 
    \text{.}
\end{align}
\normalsize

\noindent
Note that we analyzed a model comprising 12 layers; each layer has 12 attention heads.
Let $r(\cdot)$ be a function that returns the frequency rank of a given word.
We first calculated the Spearman rank correlation coefficient between $r(t^p_q)$ and $\text{AvgW}(p,q)$.
The score was 0.06, which suggests that there is no relationship between \malpha{} and the frequency rank of the word.
Then, we calculated the Spearman rank correlation coefficient between $r(t^p_q)$ and $\text{AvgN}(p,q)$.
The score was 0.75, which suggests a strong correlation between \msfx\ and the frequency rank of the word;
Figure~\ref{fig:relation_ids} shows these results. 

In addition, the results for the word frequency, instead of the frequency rank, are shown in Figure~\ref{fig:relation_count}.
$c(\cdot)$ denotes a function that returns the frequency of a given word in the training dataset of BERT.
We reproduced the dataset because it is not released.
\section{Details on Section~\ref{sec:nmt}}
\label{ap:nmt}

\subsection{Hyperparameters and training settings}
\label{ap:nmt_hyp}
We used the Transformer~\cite{vaswani17} NMT model implemented in fairseq~\citep{ott2019fairseq} for the experiments.
Table~\ref{tbl:hyperparam} shows the hyperparameters of the model, which were the same as those used by \citet{ding19-saliency}.
We used the model with the highest BLEU score in the development set for our experiments.

We conducted the data  preprocessing\footnote{\url{https://github.com/lilt/alignment-scripts}} following the method by~\citet{zenkel_aer_giza} and~\citet{ding19-saliency}.
All the words in the training data of the NMT systems were split into subword units using byte-pair encoding (BPE, \citet{sennrich16_bpe}) with 10k merge operations.
Following~\citet{ding19-saliency}, the last 1000 instances of the training data were used as the development data.

\begin{table*}[t]
    \centering
    {\small   
    \begin{tabular}{llc} \toprule
    \multirow{9}{*}{Fairseq model} & architecture & transformer\_iwslt\_de\_en \\
     & encoder embed dim. & 512 \\
     & decoder embed dim. & 512 \\
     & encoder ffn embed dim. & 1024 \\
     & decoder ffn embed dim. & 1024 \\
     & encoder attention heads & 4 \\
     & decoder attention heads & 4 \\
     & encoder layers & 6 \\
     & decoder layers & 6 \\
    \cmidrule(lr){1-1} \cmidrule(lr){2-2} \cmidrule(lr){3-3}
     Activation & function &  Relu \\
    \cmidrule(lr){1-1} \cmidrule(lr){2-2} \cmidrule(lr){3-3}
    \multirow{2}{*}{Loss} & type & label smoothed cross entropy \\
    & label smoothing & 0.1 \\
    \cmidrule(lr){1-1} \cmidrule(lr){2-2} \cmidrule(lr){3-3}
    \multirow{5}{*}{Optimizer} & algorithm & Adam \\
    & learning rates & 0.001 \\
    & $\beta_1$ & 0.9 \\
    & $\beta_2$ & 0.98 \\
    & weight decay & 0.0 \\
    & clip norm & 0.0 \\
    \cmidrule(lr){1-1} \cmidrule(lr){2-2} \cmidrule(lr){3-3}
    \multirow{3}{*}{Learning rate scheduler} & type & inverse\_sqrt \\
    & warmup updates & 4,000 \\
    & warmup init lrarning rate & 1e-07 \\
    \cmidrule(lr){1-1} \cmidrule(lr){2-2} \cmidrule(lr){3-3}
    \multirow{6}{*}{Training} & batch size & 80 \\ %
    & max tokens & 4000 \\
    & max epoch & 100 \\
    & update freq & 8 \\
    & drop out & 0.1 \\
    & seed & 2 \\
    & number of GPUs used & 2 \\
        \bottomrule
        \end{tabular}
        }
        \caption{Hyperparameters of the NMT model.}
        \label{tbl:hyperparam}
\end{table*}

\subsection{Settings of the word alignment extraction}
\label{ap:nmt_ext}
First, we applied BPE, which was used to split the training data of the NMT systems to create the evaluation data used for calculating the AER scores.
Next, we extracted the scores of \malpha\ and \msafx\ for each subword in the evaluation data for the force decoding setup.
The gold alignments are annotated at the word-level, not the subword-level.
To calculate the word-level alignment scores, \malpha\ and \msafx\ for the subwords were merged along with the target token in the gold data by averaging, then merged along with the source tokens in the gold data by summation.
These operations were the same as~\citet{li19-word_align}.

In existing studies, \texttt{$\langle$/s$\rangle$}, the special token for the end of the sentence, was probably removed in calculating word alignments.
We included \texttt{$\langle$/s$\rangle$} as the alignment targets and we considered the alignments to \texttt{$\langle$/s$\rangle$} as no alignment.
In other words, if the model aligns a certain word with \texttt{$\langle$/s$\rangle$}, we assume that the model decides that the word is not aligned to any word.
\subsection{Layer-wise analysis}
\label{subsec:nmt:layer-wise}

\begin{figure}[t]
    \centering
    \begin{minipage}[t]{\hsize}
        \centering
        \includegraphics[width=\hsize]{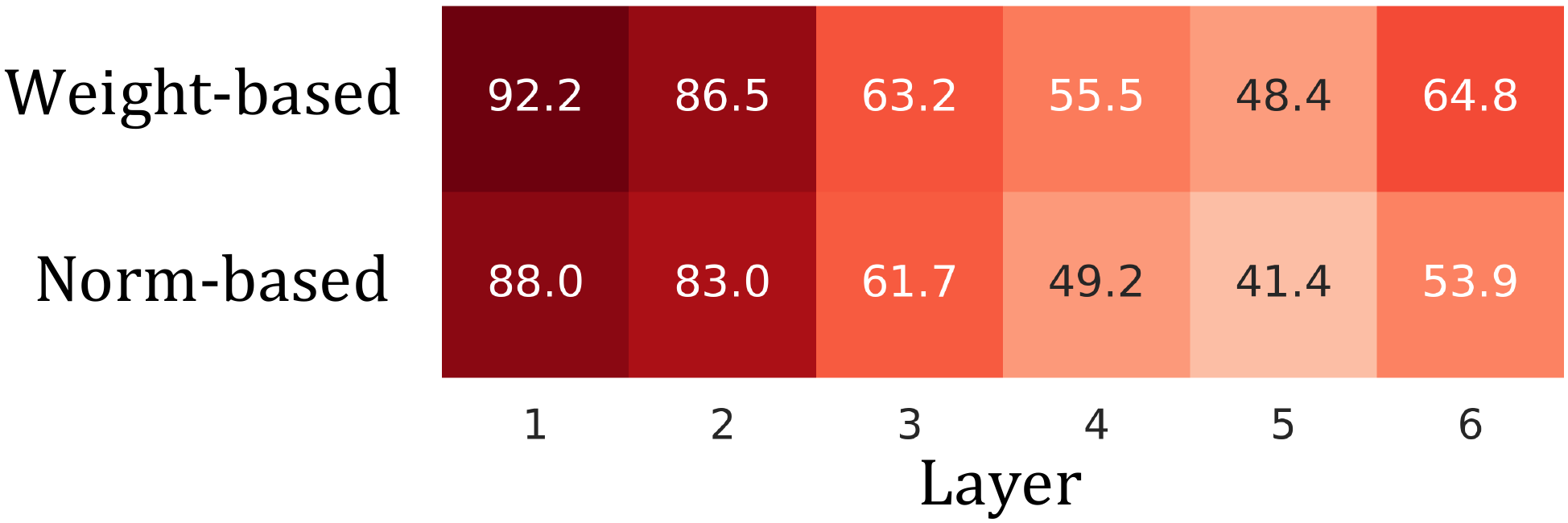}
        \subcaption{
        AWO setting.
        }
        \label{fig:layer-wise_aer_next}
    \end{minipage}
    \begin{minipage}[t]{\hsize}
        \centering
        \includegraphics[width=\hsize]{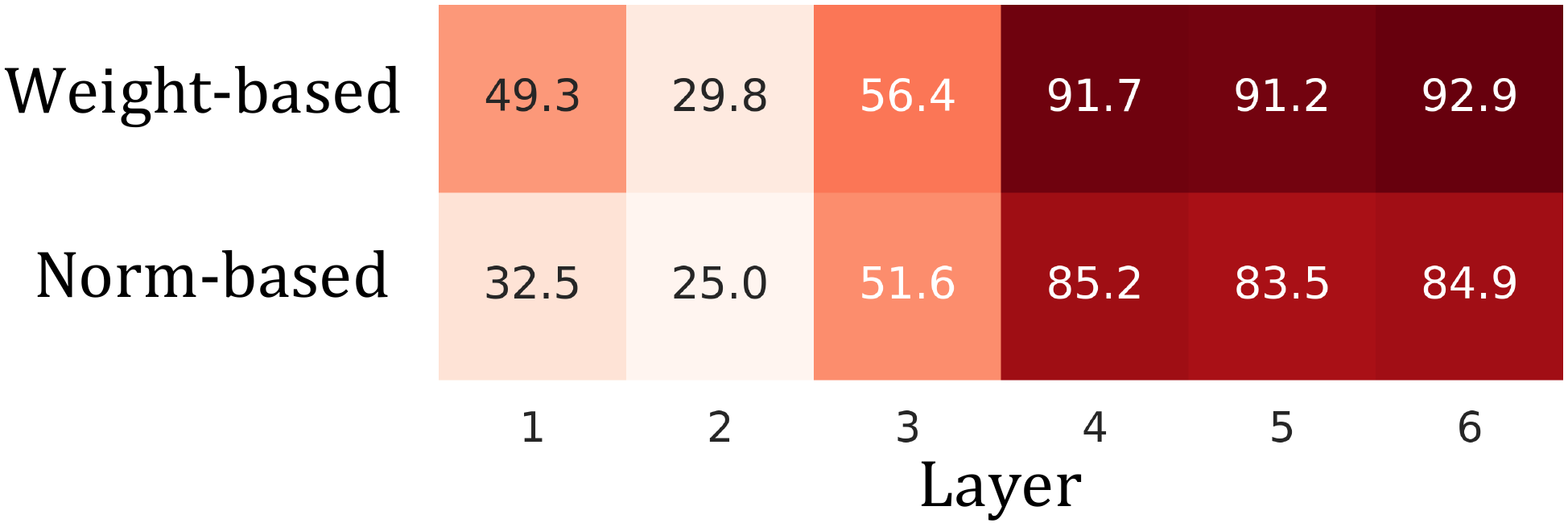}
        \subcaption{
        AWI setting.
        }
        \label{fig:layer-wise_aer_cuurrent}
    \end{minipage}
    \caption{
    Layer-wise AER scores.
    Each value is the average of five random seeds.
    The closer the extracted word alignment is to the reference, the lower the AER score---the lighter the color.
}
    \label{fig:layer-wise_aer}
\end{figure}

We preliminarily investigated how the source-target attentions in a Transformer-based NMT system behave depending on the layer.
\citet{tang_transformer_attention_analysis} have reported that they behave differently depending on the layer.
The AER scores in the AWI and AWO settings were calculated for each layer (Figure~\ref{fig:layer-wise_aer}).
In the AWO setting, AER scores tend to be better in the latter layers than in the earlier layers (Figure~\ref{fig:layer-wise_aer_next}).
In contrast, the AER scores tend to be better in the earlier layers than in the latter layers in the AWI setting (Figure~\ref{fig:layer-wise_aer_cuurrent}).

These results suggest that the earlier and latter layers focus on the source word that is aligned with the input and output target word, respectively (as shown in Figure~\ref{fig:enc-dec}).
Furthermore, we believe that it is a convincing result to extract cleaner word alignments from the AWI setting than the AWO setting (Figure~\ref{fig:layer-wise_aer}), because the AWI setting is more advantageous. 
The main advantage is that while the decoder may fail to predict the correct output words, the input words are perfectly accurate owing to the teacher forcing.
\subsection{Alignments in different layers}
\label{ap:nmt_other_lyaer}
\begin{figure*}[t]
    \centering
    \begin{minipage}[t]{.3\hsize}
        \centering
        \includegraphics[height=4cm]{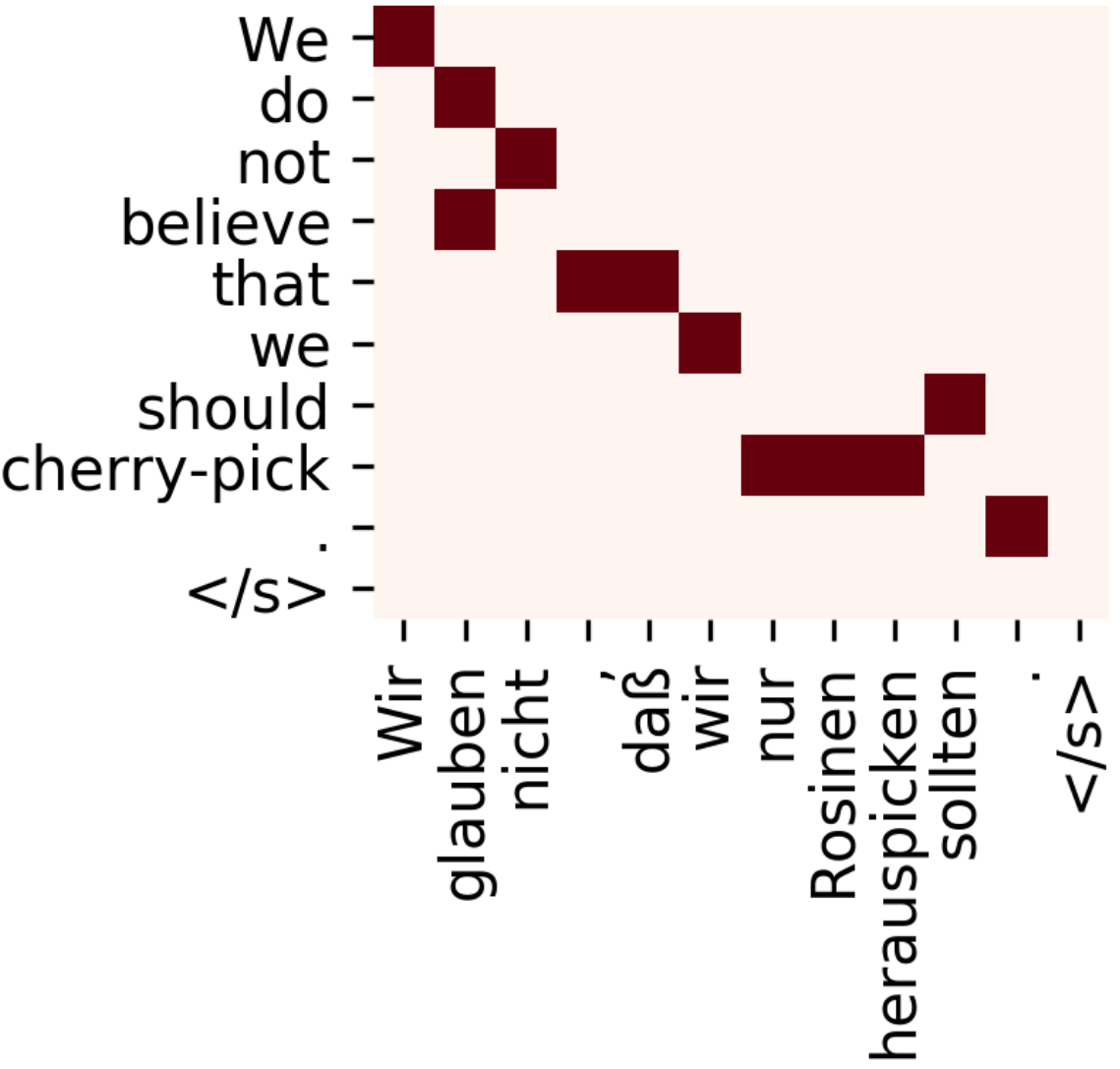}
        \subcaption{
        Reference.
        }
    \end{minipage}
    \;
    \begin{minipage}[t]{.3\hsize}
        \centering
        \includegraphics[height=4cm]{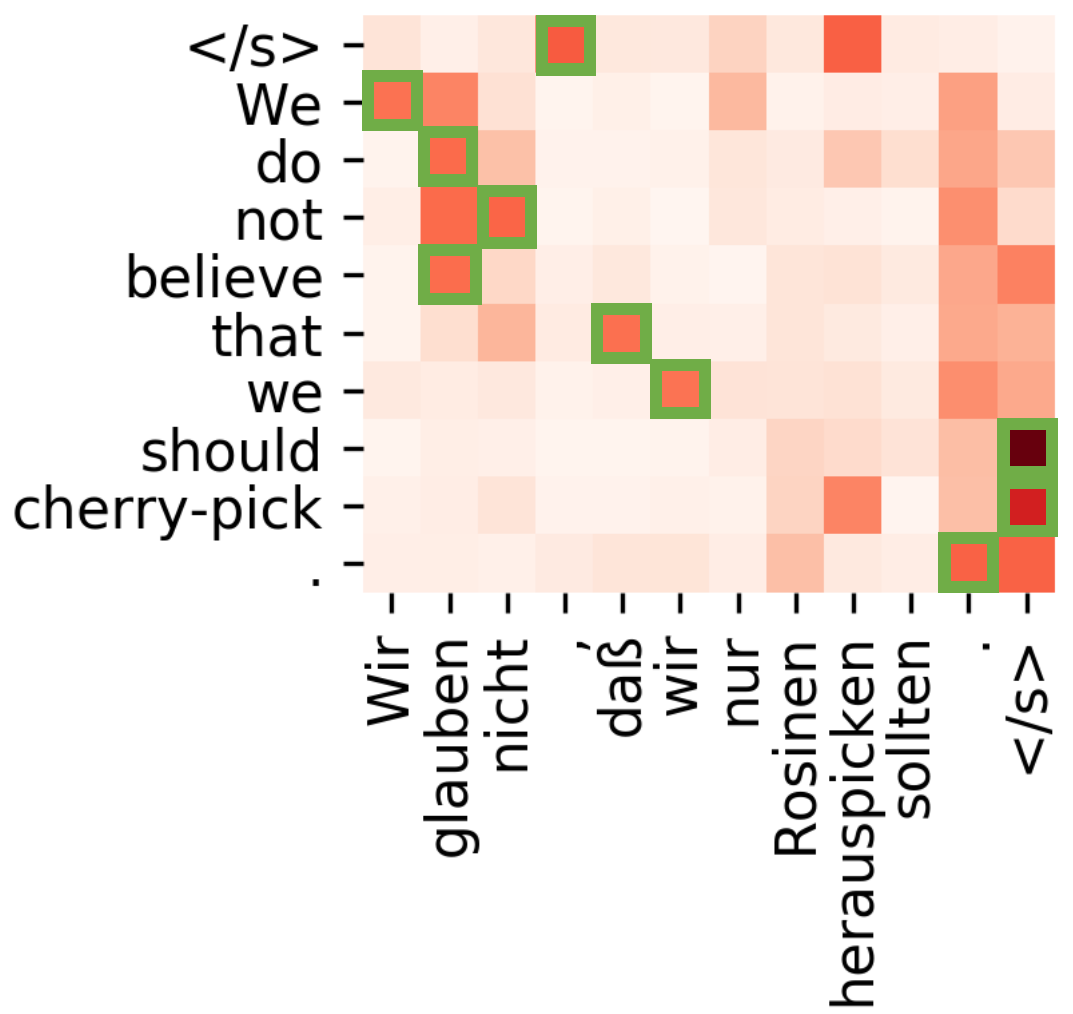}
        \subcaption{
        Attention-weights. %
        }
    \end{minipage}
    \;
    \begin{minipage}[t]{.3\hsize}
        \centering
        \includegraphics[height=4cm]{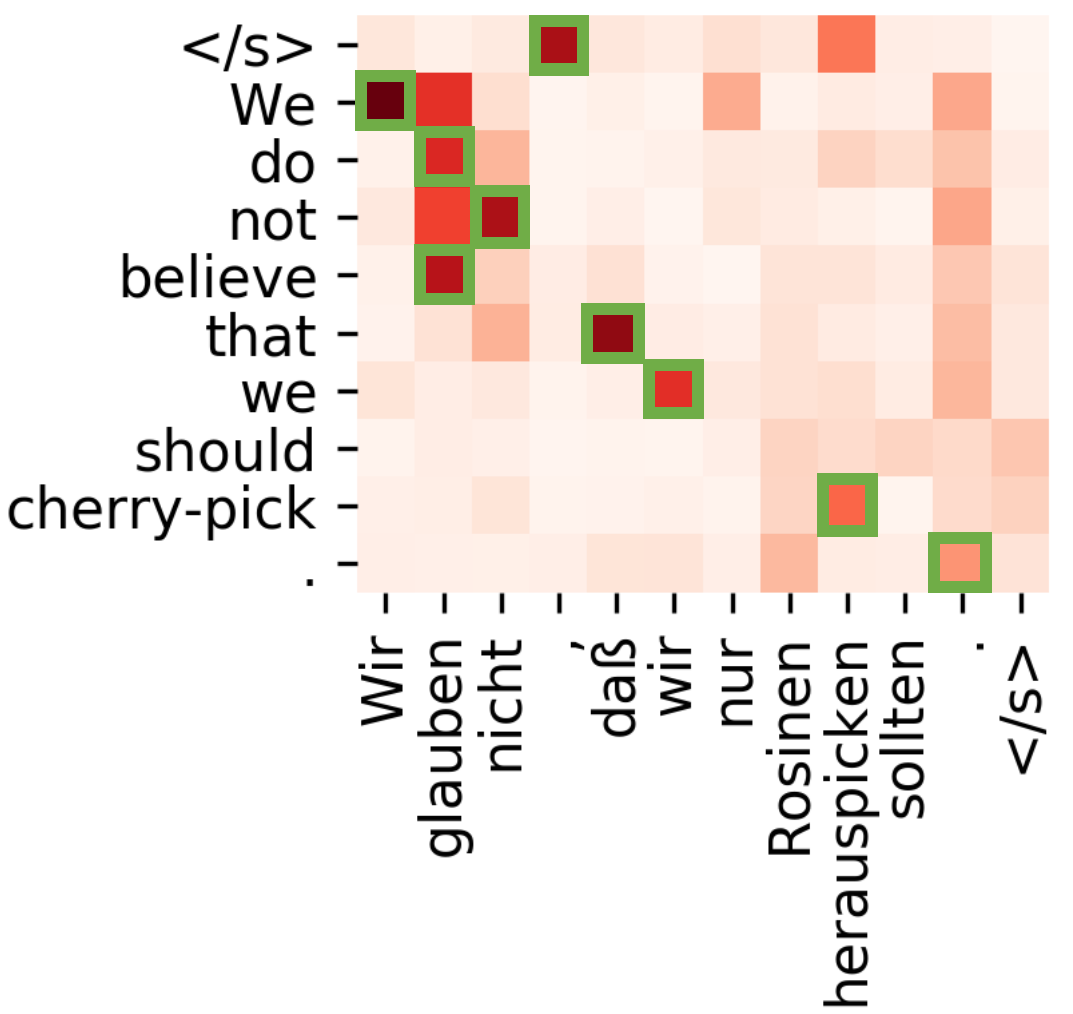}
        \subcaption{
        Vector-norms (ours).
        }
    \end{minipage}
    \caption{
    Examples of the reference alignment and the extracted patterns by each method in layer 1.
    Word pairs with a green frame shows the word with the highest weight or norm.
    The vertical axis represents the input source word in the decoder, and the pairs with a green frame are extracted as alignments in the AWI setting.
    Note that pairs that contain \texttt{$\langle$/s$\rangle$} not extracted.
    }
    \label{fig:aer_ex_layer1}
\end{figure*}

\begin{figure}[t]
    \centering
    \begin{minipage}[t]{.57\hsize}
        \centering
        \includegraphics[height=4cm]{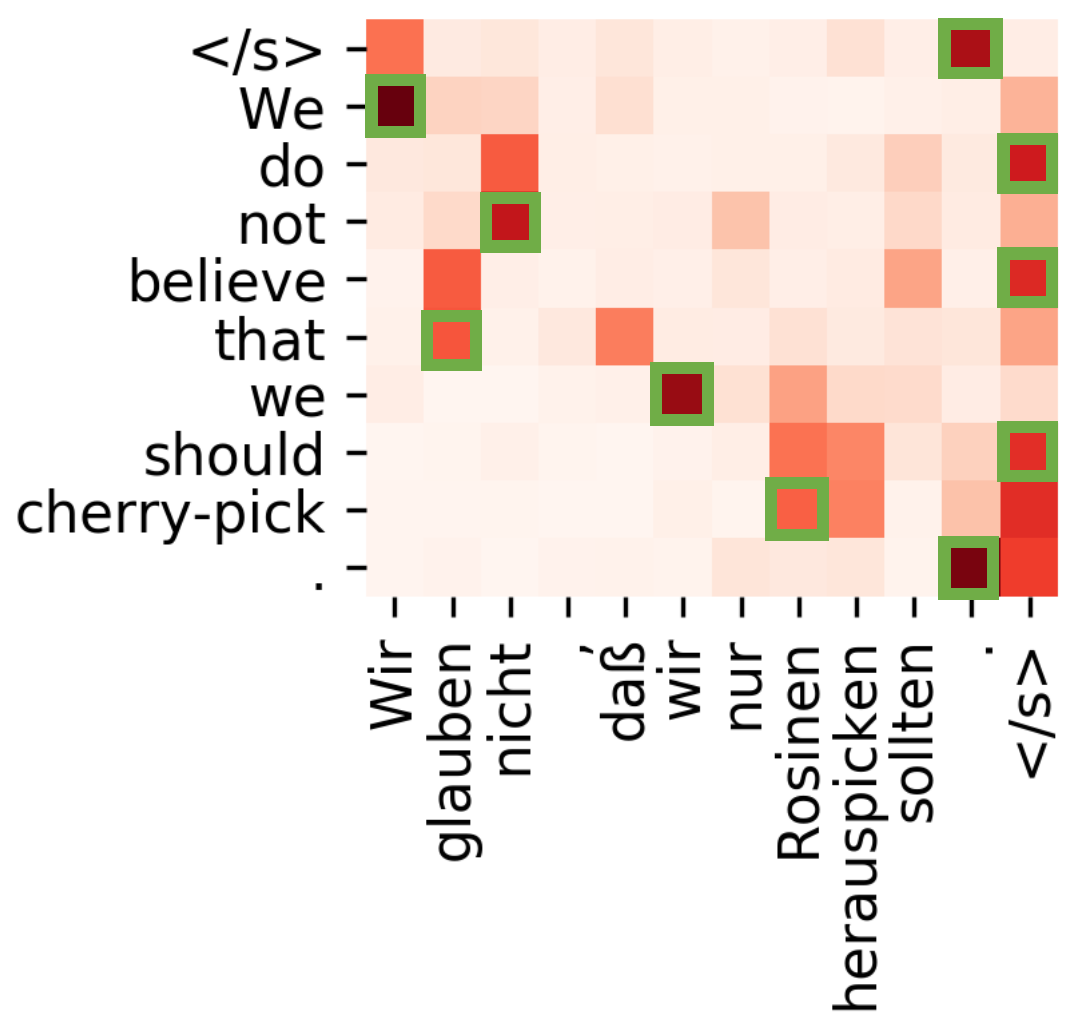}
        \subcaption{
        Attention-weights. %
        }
    \end{minipage}
    \begin{minipage}[t]{.37\hsize}
        \centering
        \includegraphics[height=4cm]{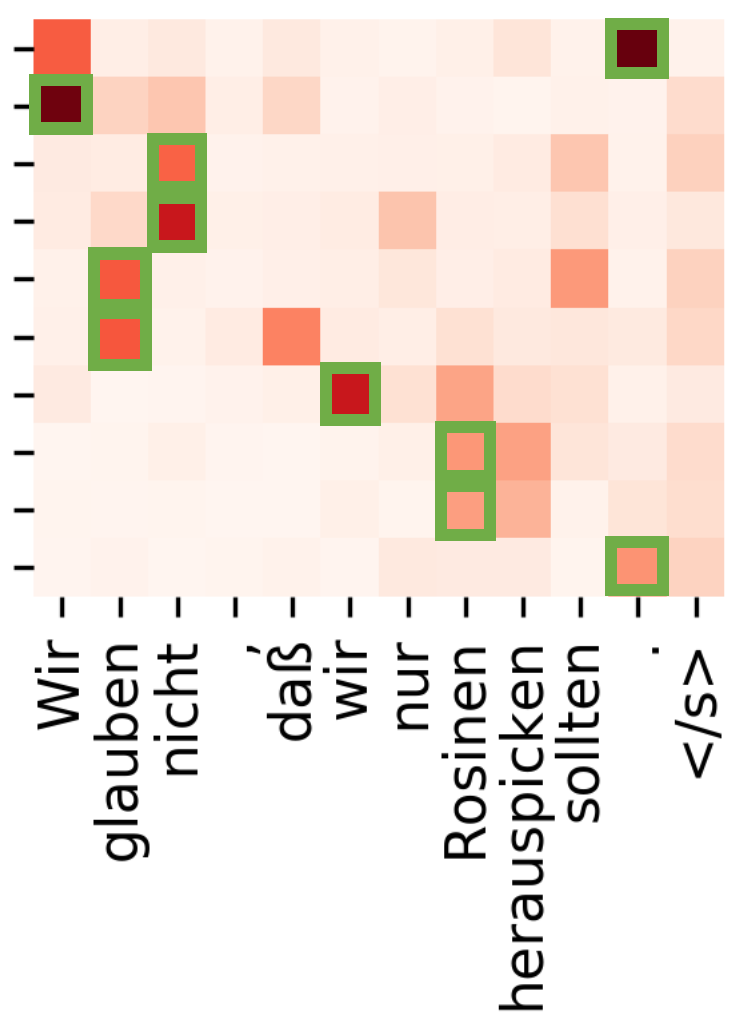}
        \subcaption{
        Vector-norms.%
        }
    \end{minipage}
    \caption{
    Examples of the reference alignment and the extracted patterns by each method in layer 2.
    }
    \label{fig:aer_ex_layer2}
\end{figure}

\begin{figure}[t]
    \centering
    \begin{minipage}[t]{.57\hsize}
        \centering
        \includegraphics[height=4cm]{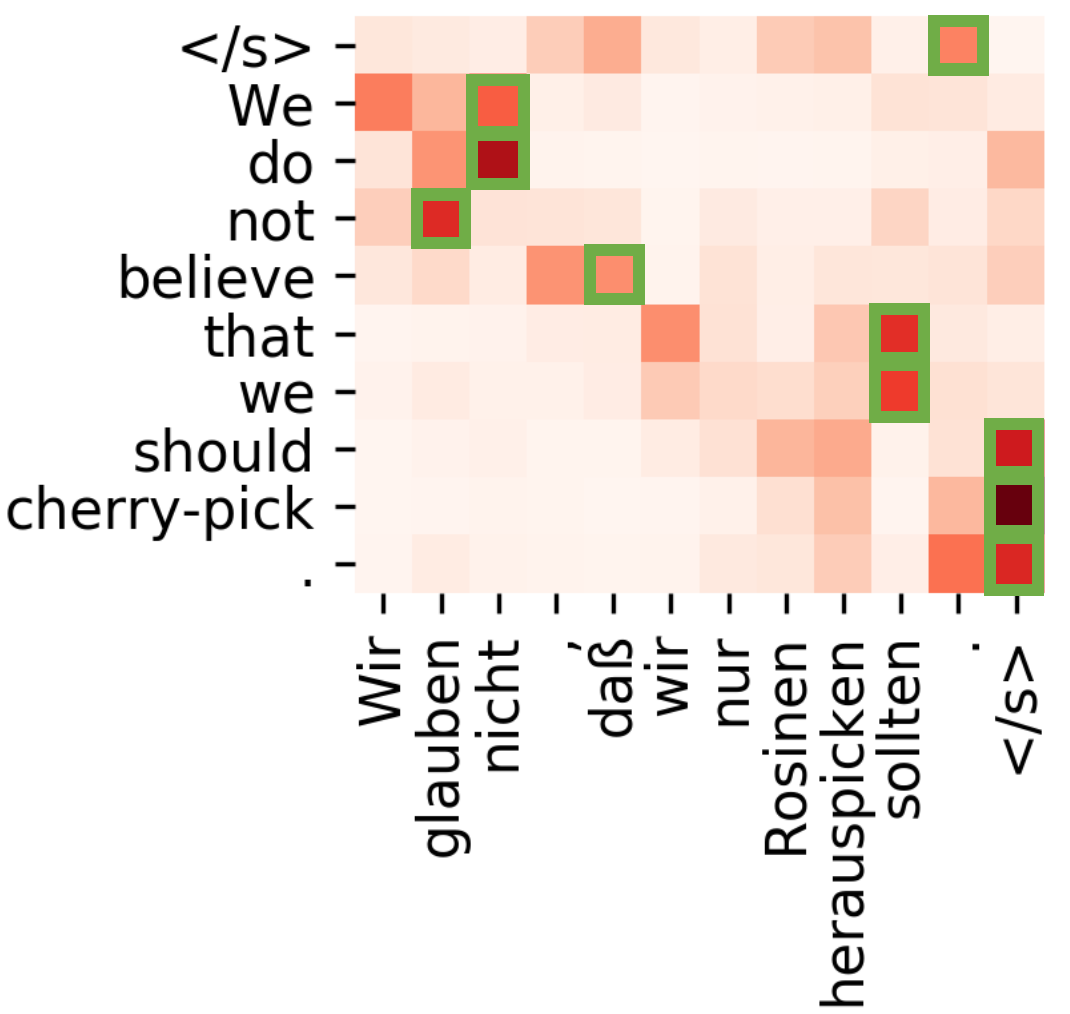}
        \subcaption{
        Attention-weights. %
        }
    \end{minipage}
    \begin{minipage}[t]{.37\hsize}
        \centering
        \includegraphics[height=4cm]{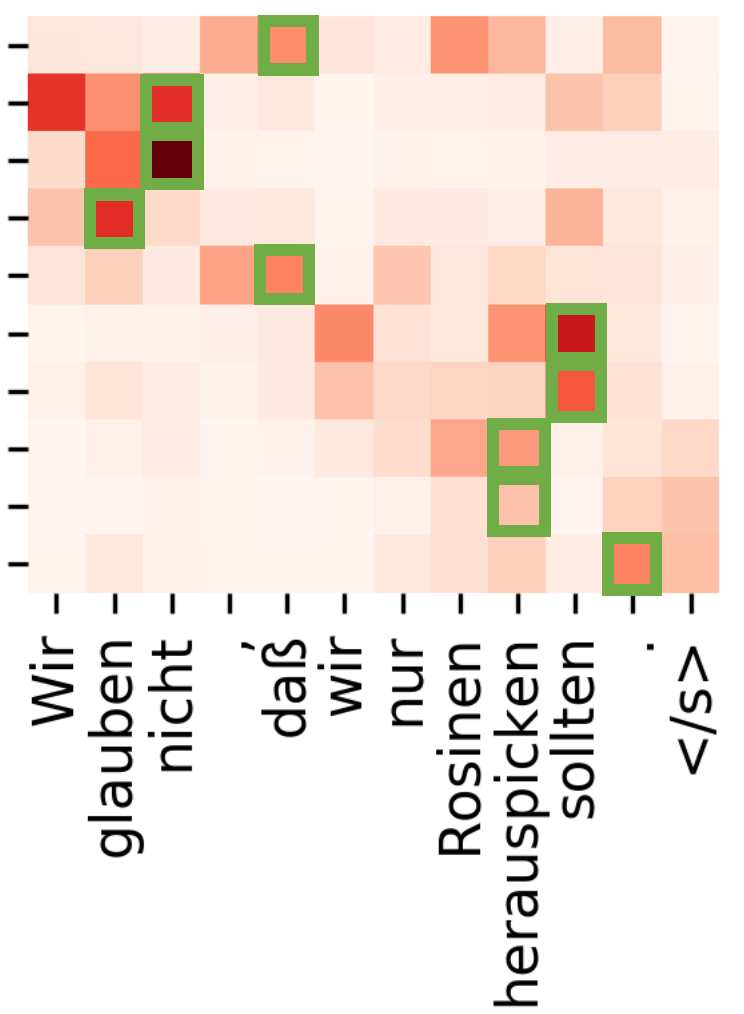}
        \subcaption{
        Vector-norms.%
        }
    \end{minipage}
    \caption{
    Examples of the reference alignment and the extracted patterns by each method in layer 3.
    }
    \label{fig:aer_ex_layer3}
\end{figure}

\begin{figure}[t]
    \centering
    \begin{minipage}[t]{.57\hsize}
        \centering
        \includegraphics[height=4cm]{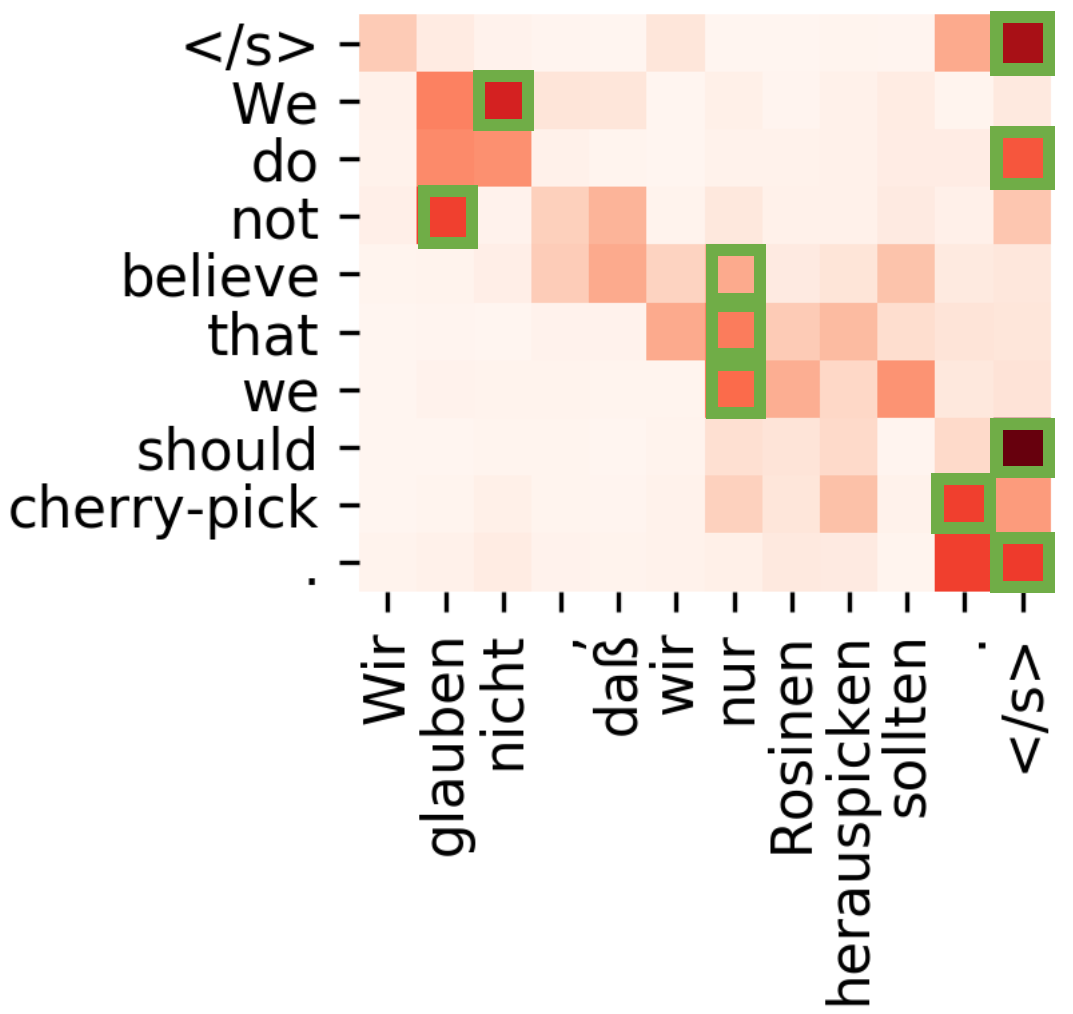}
        \subcaption{
        Attention-weights. %
        }
    \end{minipage}
    \begin{minipage}[t]{.37\hsize}
        \centering
        \includegraphics[height=4cm]{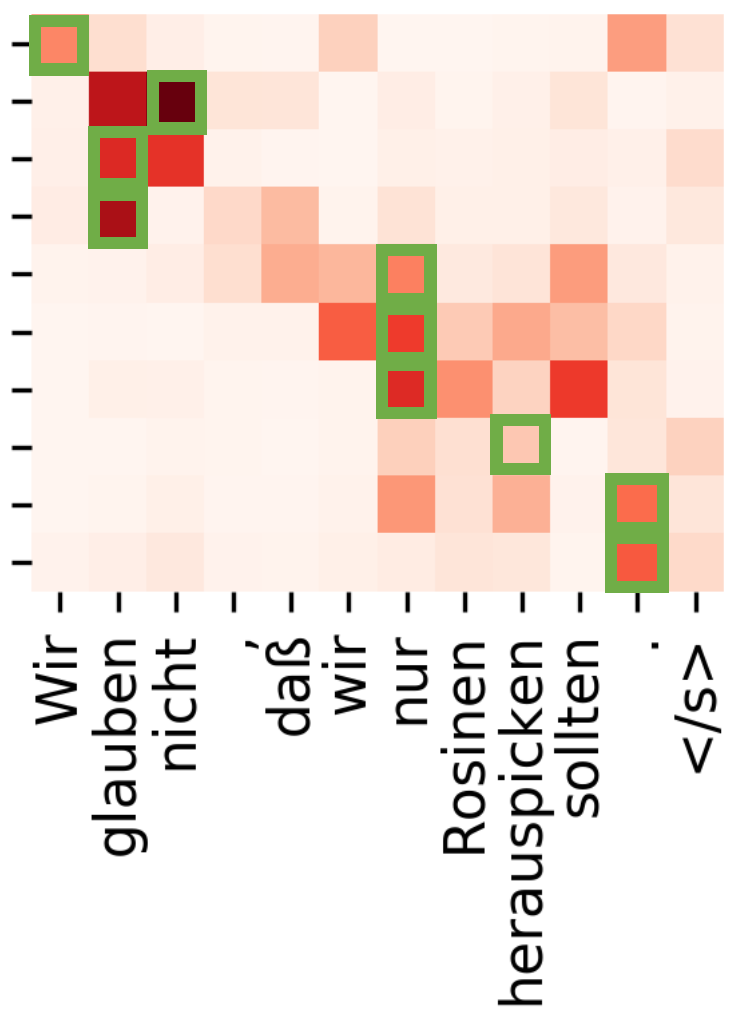}
        \subcaption{
        Vector-norms.%
        }
    \end{minipage}
    \caption{
    Examples of the reference alignment and the extracted patterns by each method in layer 4.
    }
    \label{fig:aer_ex_layer4}
\end{figure}

\begin{figure}[t]
    \centering
    \begin{minipage}[t]{.57\hsize}
        \centering
        \includegraphics[height=4cm]{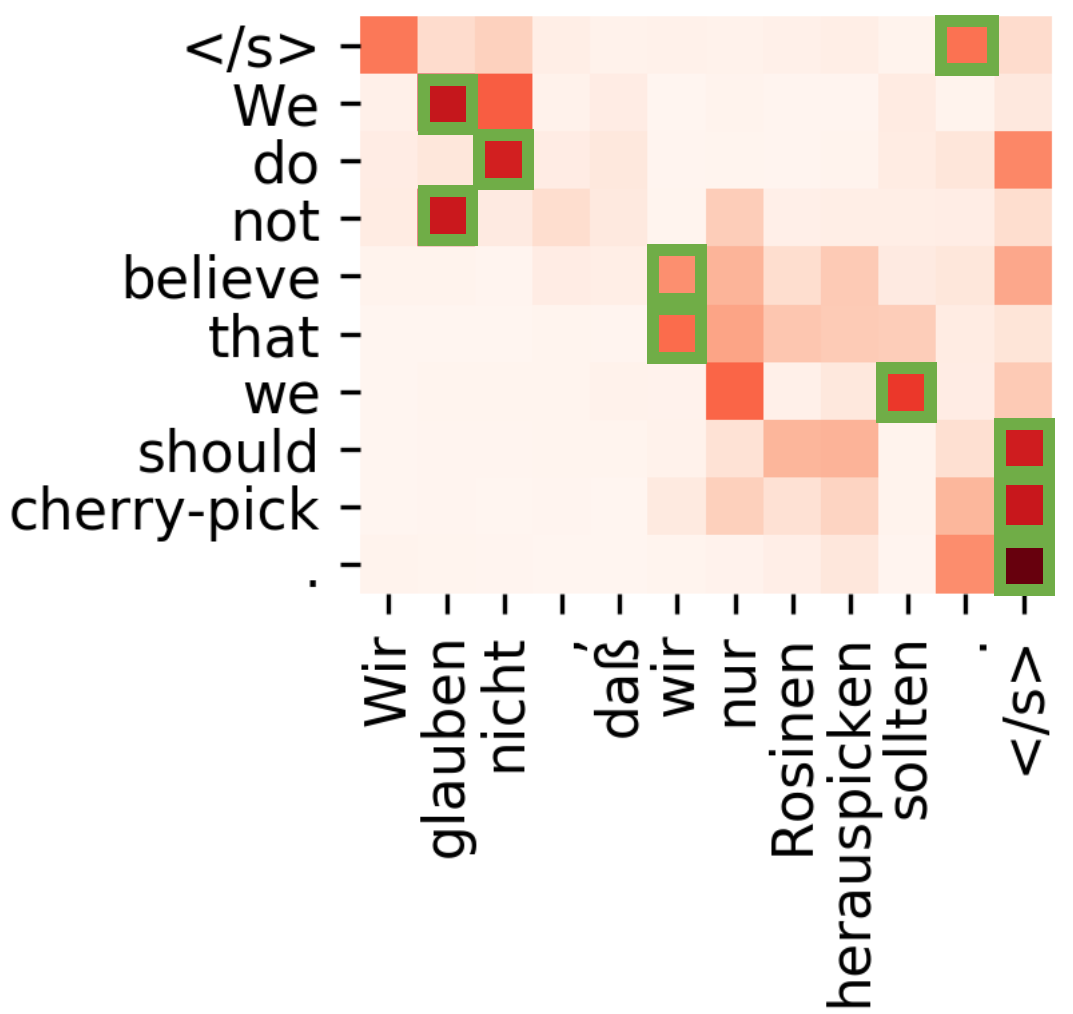}
        \subcaption{
        Attention-weights. %
        }
    \end{minipage}
    \begin{minipage}[t]{.37\hsize}
        \centering
        \includegraphics[height=4cm]{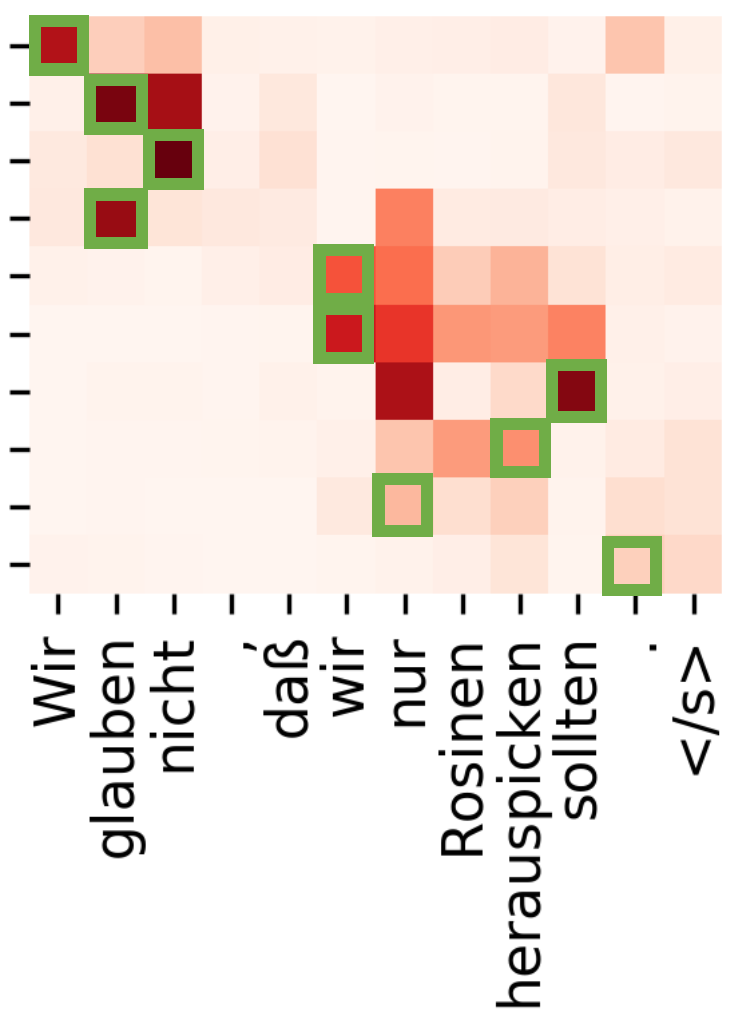}
        \subcaption{
        Vector-norms.%
        }
    \end{minipage}
    \caption{
    Examples of the reference alignment and the extracted patterns by each method in layer 5.
    }
    \label{fig:aer_ex_layer5}
\end{figure}

\begin{figure}[t]
    \centering
    \begin{minipage}[t]{.57\hsize}
        \centering
        \includegraphics[height=4cm]{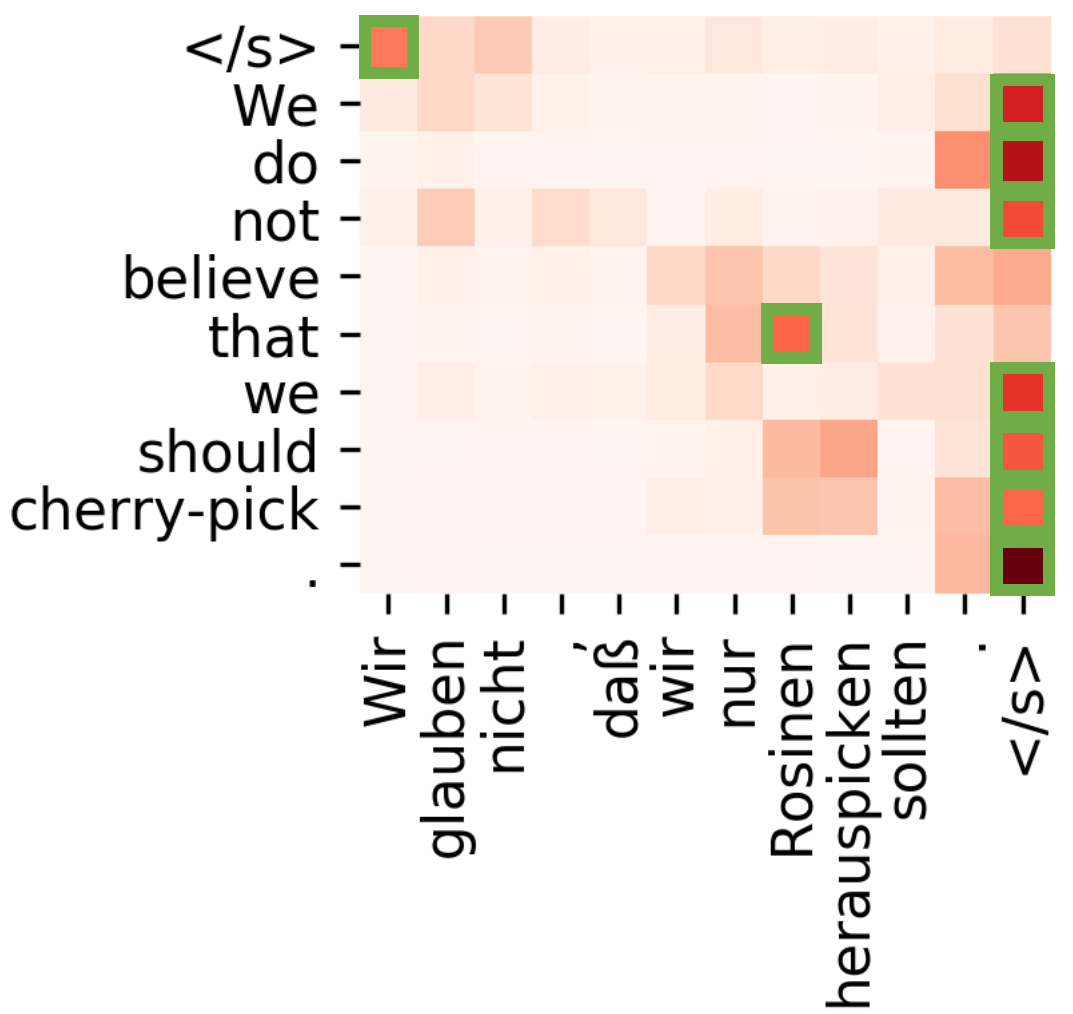}
        \subcaption{
        Attention-weights. %
        }
    \end{minipage}
    \begin{minipage}[t]{.37\hsize}
        \centering
        \includegraphics[height=4cm]{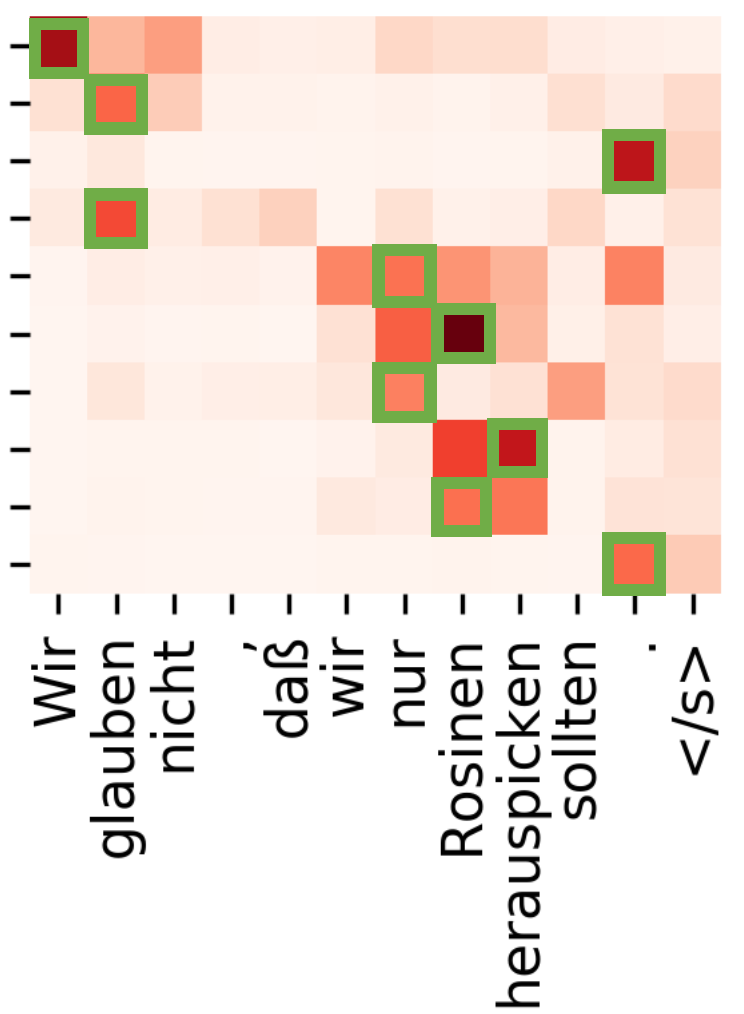}
        \subcaption{
        Vector-norms. %
        }
    \end{minipage}
    \caption{
    Examples of the reference alignment and the extracted patterns by each method in layer 6.
    }
    \label{fig:aer_ex_layer6}
\end{figure}

Figures~\ref{fig:aer_ex_layer1} to \ref{fig:aer_ex_layer6} show additional examples of the extracted alignments from the different layers of the NMT system. 
Note that the color scale in each heatmap %
is determined by the maximum value in each figure.
One can observe that while the attention weights \malpha\ are biased towards \texttt{$\langle$/s$\rangle$}, the norms \msafx\ corresponding to the token are small.

\section{Word alignment experiments on different settings}
\label{ap:nmt_wmt}
To verify whether the results obtained in the Section~\ref{sec:nmt} are reproducible in different settings, we conducted an additional experiment using the model with a different number of attention heads.
Specifically, we used a model with eight attention heads in both the encoder and decoder.
Table~\ref{table:8head_aer} shows the AER scores of the 8-head model.
As with the results obtained by the 4-head model, word alignments extracted using the proposed norm-based approach were more reasonable than those extracted using the weight-based approach, and better word alignments are extracted in the AWI setting than in the AWO setting.
Furthermore, the alignments extracted using the head or the layer with the highest average \msafx\ in the AWI setting are competitive with one of the existing word aligners---fast\_align.
With respect to the weight-based extraction, the scores obtained using the 8-head model were worse than those obtained using the 4-head model.
This may be owing to the increase in the number of heads that do not capture reasonable alignments. %

\begin{figure}[t]
    \centering
    \begin{minipage}[t]{\hsize}
        \centering
        \includegraphics[width=.8\hsize]{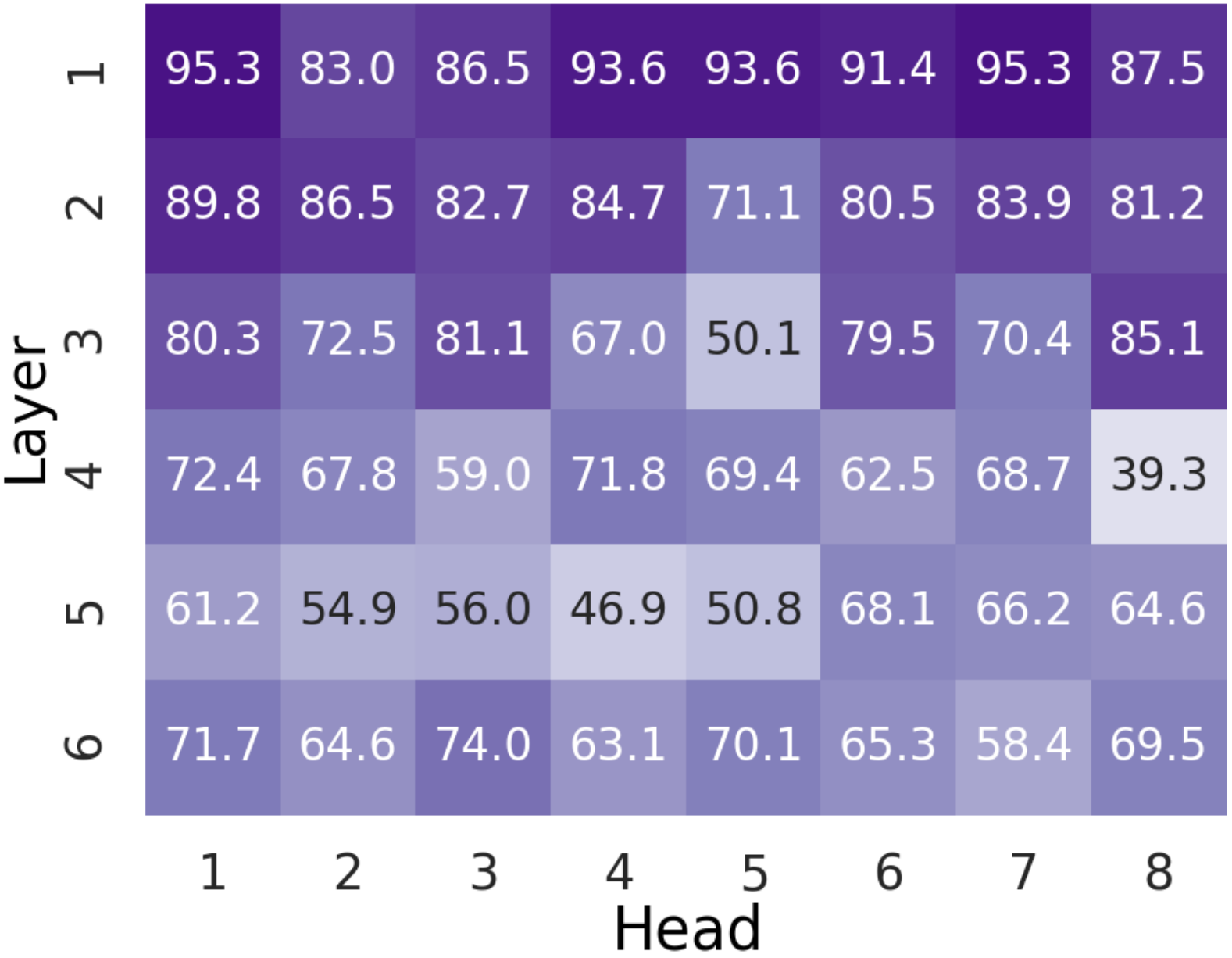}
        \subcaption{
        AER in the AWO setting.
        }
        \label{fig:8head_aer_next}
    \end{minipage}
    \begin{minipage}[t]{\hsize}
        \centering
        \includegraphics[width=.8\hsize]{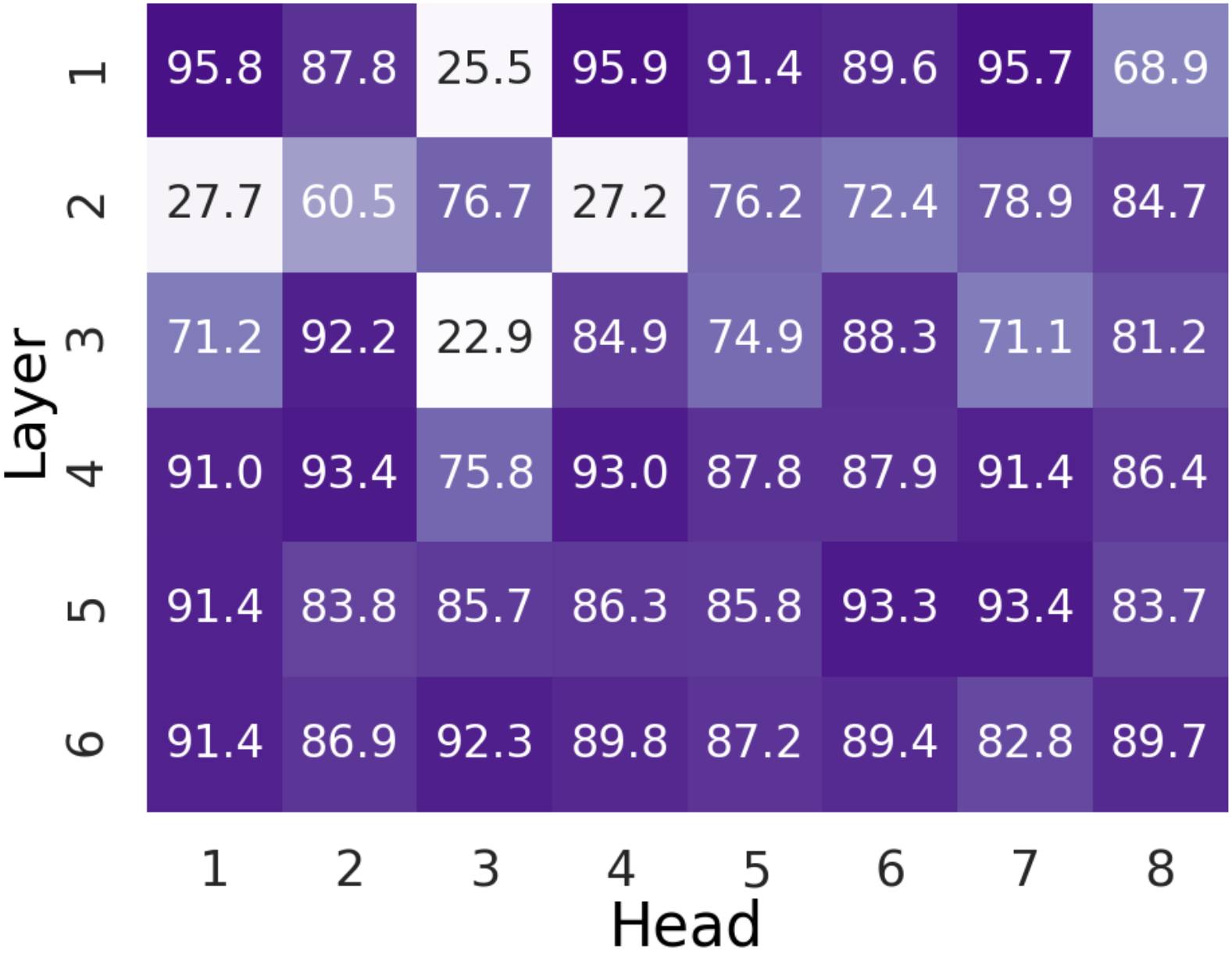}
        \subcaption{
        AER in the AWI setting.
        }
        \label{fig:8head_aer_current}
    \end{minipage}
    \begin{minipage}[t]{\hsize}
        \centering
        \includegraphics[width=.8\hsize]{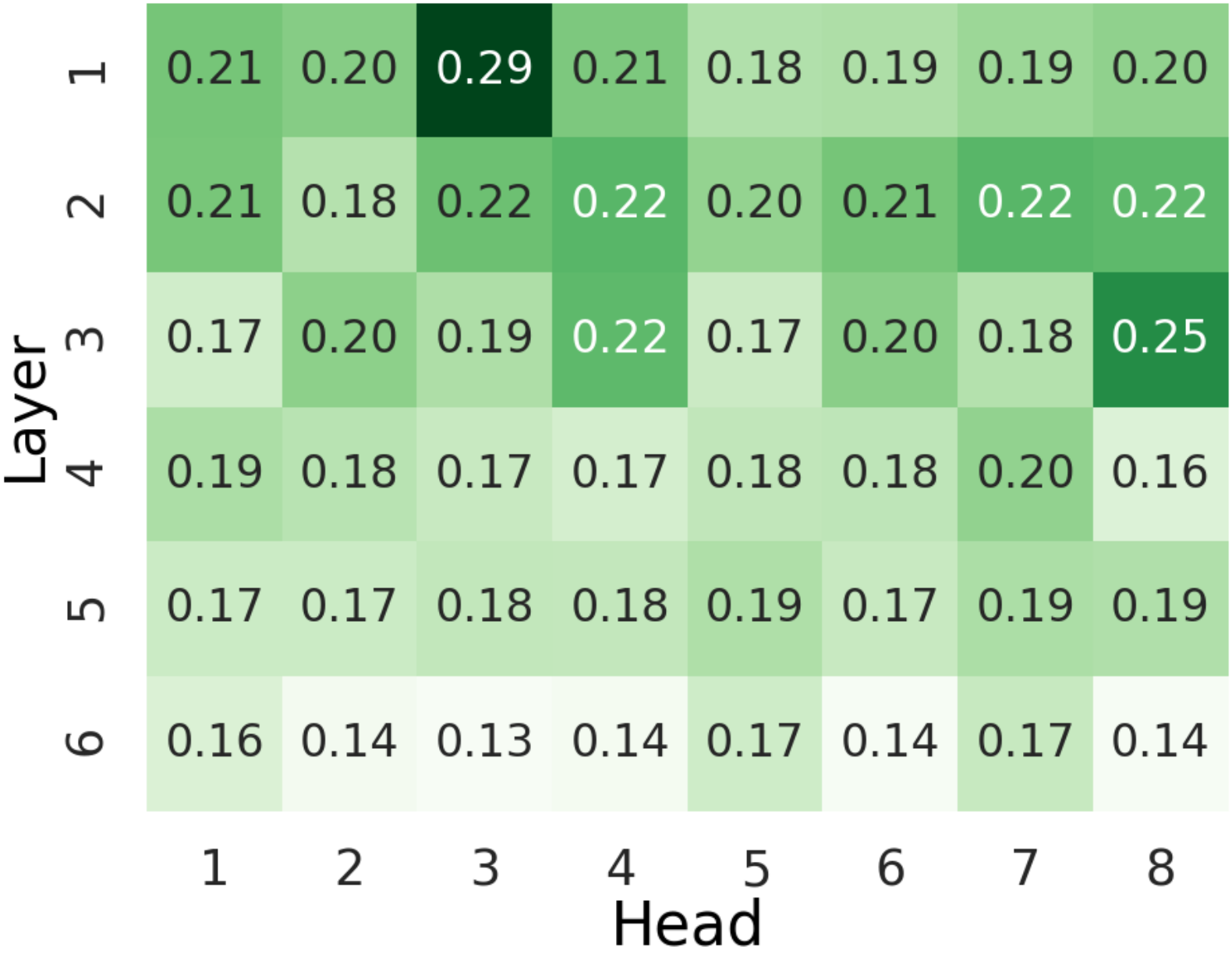}
        \subcaption{
        Averaged \msafx.
        }
        \label{fig:8head_job}
    \end{minipage}
    \caption{AER scores and averaged \msafx{} for each head in a model with 8 heads.
    }
    \label{fig:8head_aer_job}
\end{figure}

Figures~\ref{fig:8head_aer_next} and \ref{fig:8head_aer_current} show the AER scores of the alignments obtained by the norm-based extraction at each head on one out of five seeds.
Figure~\ref{fig:8head_job} shows the average of \msafx{} at each head.
As with the results obtained by the 4-head model, the heads with the low (i.e., better) AER score in the AWI setting tended to have the high \msafx{}
(the Spearman rank and Pearson correlation coefficients between the AER scores and averaged \msafx{} among the 6$\times$8 heads are $-0.26$ and $-0.50$). %
In contrast, in the AWO setting, such a negative correlation is not observed; rather, a positive correlation is observed (the Spearman's $\rho$ is $0.40$ %
and the Pearson's $r$ is $0.40$). %

\begin{figure}[t]
    \centering
    \begin{minipage}[t]{\hsize}
        \centering
        \includegraphics[width=\hsize]{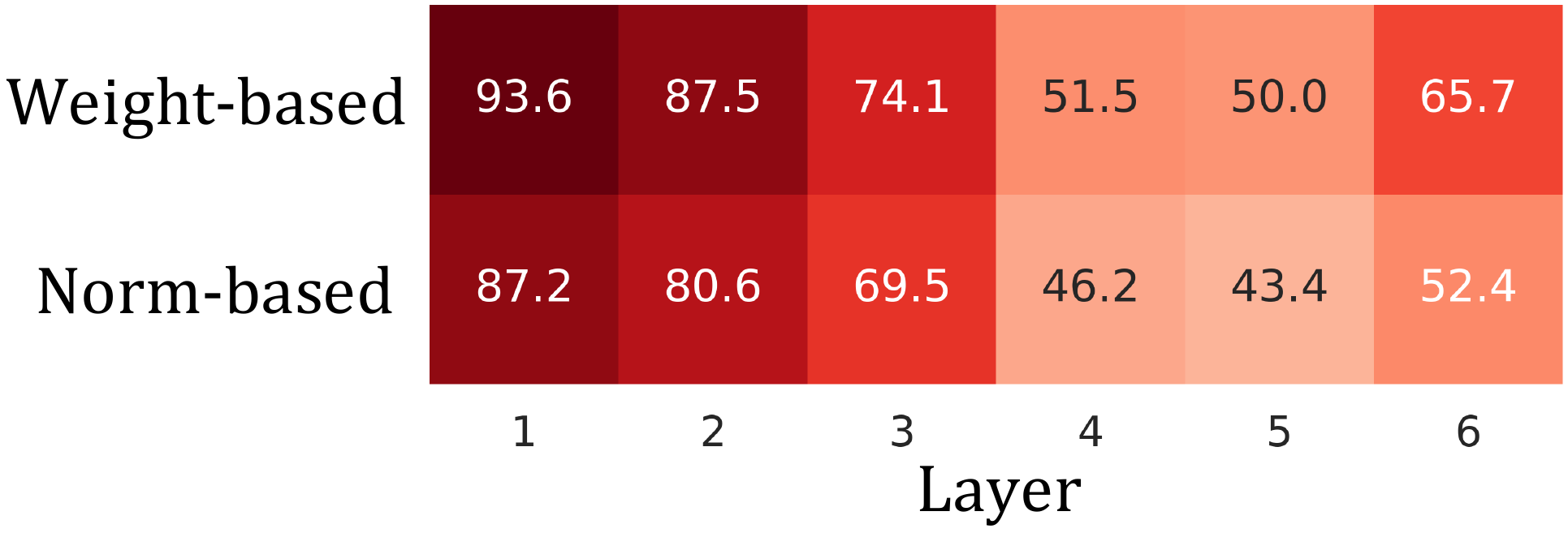}
        \subcaption{
        AWO setting.
        }
        \label{fig:layer-wise_aer_next_8head}
    \end{minipage}
    \begin{minipage}[t]{\hsize}
        \centering
        \includegraphics[width=\hsize]{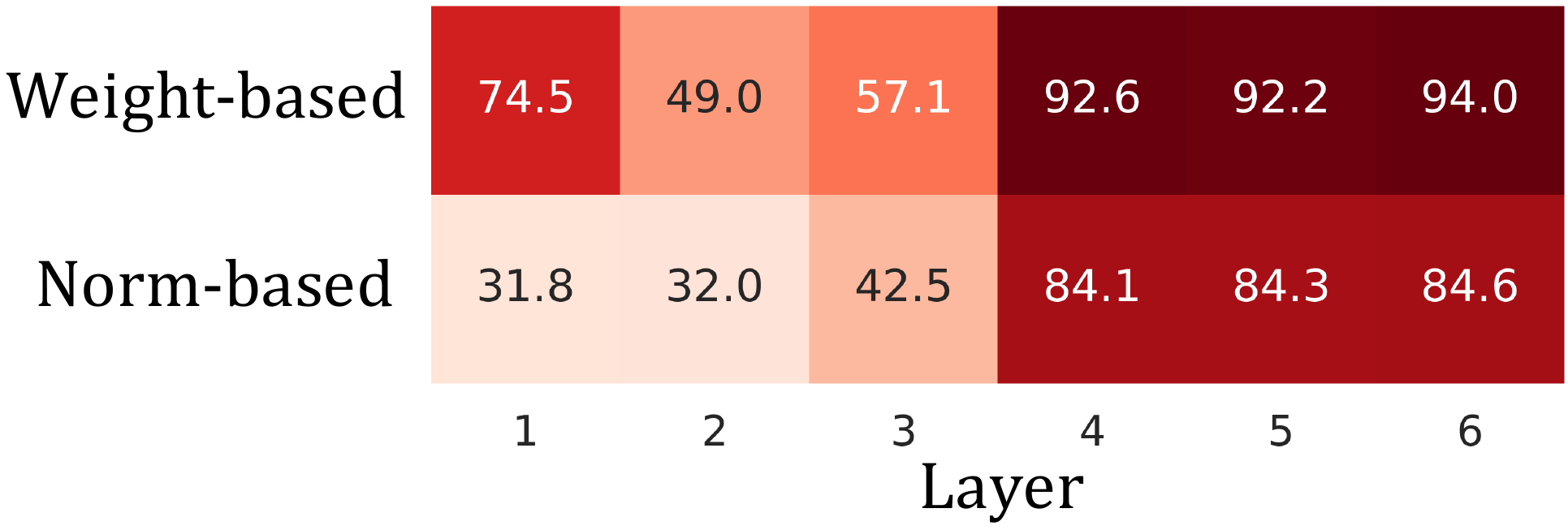}
        \subcaption{
        AWI setting.
        }
        \label{fig:layer-wise_aer_cuurrent_8head}
    \end{minipage}
    \caption{
    Layer-wise AER scores.
    Each value is the average of five random seeds.
    The closer the extracted word alignment is to the reference, the lower the AER score---the lighter the color.
}
    \label{fig:layer-wise_aer_8head}
\end{figure}

Additionally, following Appendix~\ref{subsec:nmt:layer-wise}, the AER scores for both the AWI and AWO settings for each layer were calculated (Figure~\ref{fig:layer-wise_aer_8head}).
As with the 4-head model (Appendix~\ref{subsec:nmt:layer-wise}), the latter layers correspond to the AWO setting and the earlier layers corresponds to the AWI setting in the 8-head model.

\begin{table}[t]
\centering
\setlength{\tabcolsep}{4pt}  %
\renewcommand{\arraystretch}{0.85}
{\small
\begin{tabular}{lcc}
\toprule
\multicolumn{1}{c}{\textbf{Methods}} & AER & ±SD \\
\cmidrule(r){1-1} \cmidrule{2-2} \cmidrule(l){3-3}
\textbf{Transformer -- Attention-based Approach} \\
\multicolumn{1}{c}{--- \textit{Alignment with output} setting ---} \\
Weight-based \\
\hspace{15pt} layer mean & 70.4 & 0.6 \\ 
\hspace{15pt} best layer (layer 4 or 5) & 49.3 & 1.2 \\ 
Norm-based (ours) \\
\hspace{15pt} layer mean & 63.2 & 0,7 \\ 
\hspace{15pt} best layer (layer 5) & 43.4 & 0.8 \\
\hspace{15pt} head with the highest average \msafx\  & 87.2 & 0.6 \\
\hspace{15pt} layer with the highest average \msafx\ & 83.7 & 2.2 \\
\\
\multicolumn{1}{c}{--- \textit{Alignment with input} setting ---} \\
Weight-based \\
\hspace{15pt} layer mean & 76.6 & 1.7 \\ 
\hspace{15pt} best layer (layer 2 or 3) & 38.7 & 8.9 \\ 
Norm-based (ours) \\
\hspace{15pt} layer mean & 59.9 & 1.0 \\ 
\hspace{15pt} best layer (layer 2 or 3) & 26.3 & 1.9 \\
\hspace{15pt} head with the highest average \msafx\  & 24.9 & 1.7 \\
\hspace{15pt} layer with the highest average \msafx\ & 26.5 & 1.9 \\
\cmidrule(r){1-1} \cmidrule{2-2} \cmidrule(l){3-3}
\textbf{Word Aligner} \\
fast\_align from \citet{zenkel_aer_giza} & 28.4 & - \\
GIZA++ from \citet{zenkel_aer_giza} & 21.0 & - \\ \bottomrule
\end{tabular}
}
\caption{
Results on a model trained with the same settings as described in Appendix~\ref{ap:nmt_hyp} except that the number of attention heads in the encoder and decoder is 8. Each value is the average of five random seeds.
}
\label{table:8head_aer}
\end{table}

\section{Comparison with effective attention \citep{brunner19}}
\label{ap:comp_eff_attn}
\begin{figure}[t]
    \centering
    \begin{minipage}[t]{\hsize}
        \centering
        \includegraphics[height=4.5cm]{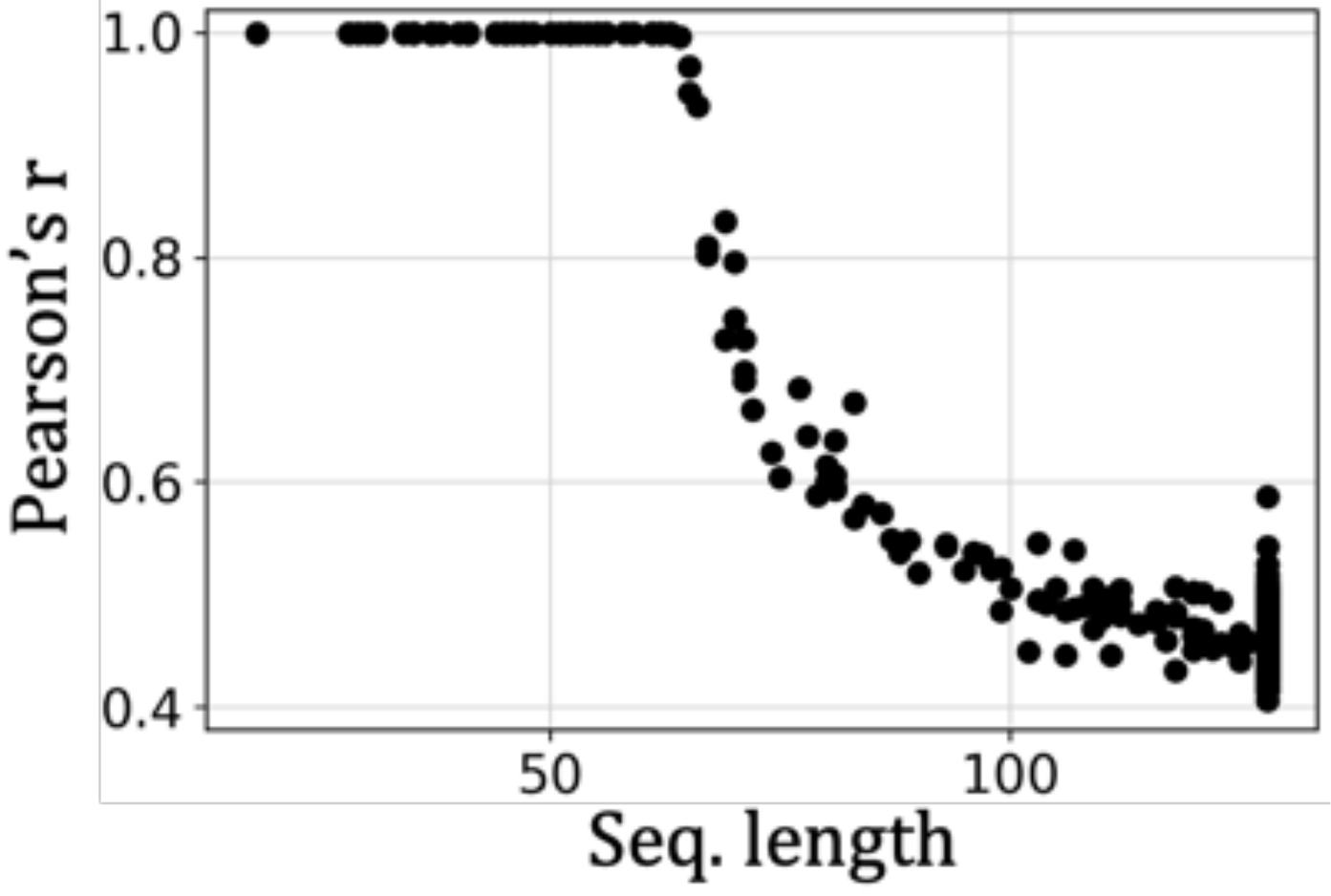}
        \subcaption{
        Effective attention.
        }
        \label{fig:eff_attn_pearson}
    \end{minipage}
    \;
    \begin{minipage}[t]{\hsize}
        \centering
        \includegraphics[height=4.5cm]{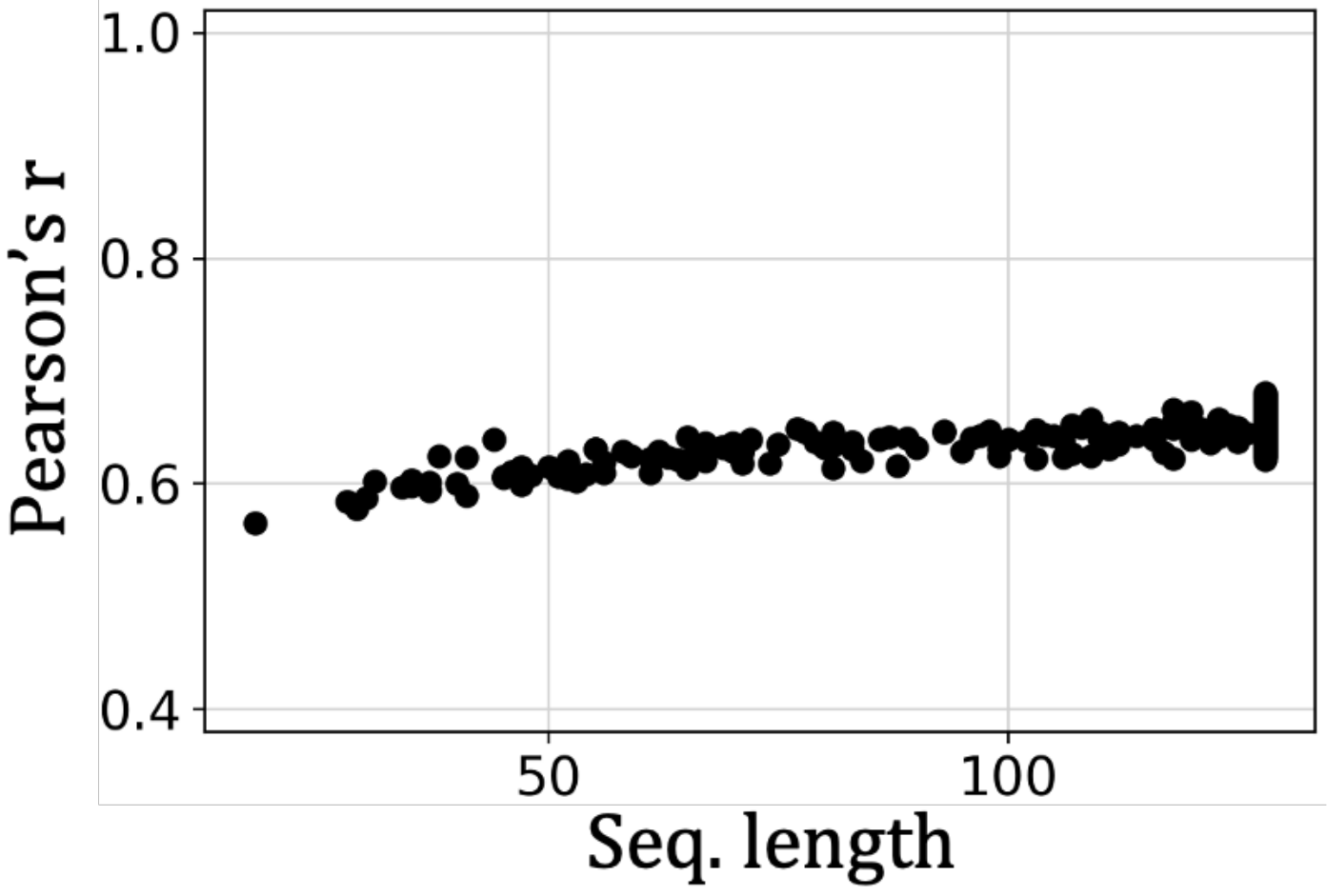}
        \subcaption{
        \msafx.
        }
        \label{fig:afx_pearson}
    \end{minipage}
    \caption{
    Each point represents the Pearson correlation coefﬁcient of raw attention and each method toward token length.
    }
    \label{fig:attn_pearson}
\end{figure}

In this section, we discuss the difference between our approach and ``effective attention''~\citep{brunner19}, which is an enhanced version of the weight-based analysis.
The effective attention exclude the components that do not affect the output owing to the application of transformation $f$ and input $\VEC{x}$ from the attention weight matrix $\VEC{A}$.
The output-irrelevant components are derived from the null space of the matrix $\VEC{T}$, which is the stack of \mfx.
Figure~\ref{fig:eff_attn_pearson} shows the Pearson correlation coefficient between the raw attention weight and the effective attention.
Since the dimension of the null space of the matrix $\VEC{T}$ depends on the length of the input sequence, as shown in Figure~\ref{fig:eff_attn_pearson}, the effective attention and raw attention weight are identical for short input sequences.
Figure~\ref{fig:afx_pearson} shows the Pearson correlation coefficient between the raw attention weight and our norm-based method.
Since we incorporate the scaling effects of $f$ and $\VEC{x}$, which contain canceling, our proposed method \msafx{} differs from the raw attention weight, whether the input sequence is long or short.

\end{document}